\documentclass[11pt, twoside]{report}
\usepackage[utf8]{inputenc}
\usepackage{graphicx}
\usepackage[hidelinks]{hyperref}
\usepackage{caption}
\usepackage{subcaption}
\usepackage{amsmath,amssymb}
\usepackage{fixltx2e}
\usepackage{algorithm, algcompatible}
\usepackage[a4paper,width=150mm,top=25mm,bottom=25mm,bindingoffset=6mm]{geometry}
\usepackage{titlesec}
\usepackage{setspace}  

\usepackage{biblatex}
\begin{document}
    \pagenumbering{roman}
    \pagestyle{empty}
    
\begin{titlepage}
\begin{center}

\newcommand{\HorizontalLine}{\rule{\linewidth}{0.3mm}}

\HorizontalLine \\[0.4cm]
\begin{spacing}{3.0}
    {\huge \bfseries Neural architecture impact } \\
    {\huge \bfseries on identifying temporally extended} \\
    {\huge \bfseries Reinforcement Learning tasks}\\
\end{spacing}
\HorizontalLine \\[1.25cm]

{\Large Master Thesis} \\[1.25cm]

{\normalsize \textbf{Written by:}} \\[0.25cm]
{\Large Victor Vadakechirayath George} \\[1.25cm]

{\normalsize \textbf{Writing period:}} \\[0.25cm]
{\Large 15.04.2021-9.9.2021} \\[1.25cm]

{\normalsize \textbf{Examiners:}} \\[0.25cm]
{\Large Prof. Dr. Frank Hutter} \\[0.25cm]
{\Large Prof. Dr. Thomas Brox} \\[1.25cm]

{\normalsize \textbf{Advisor:}} \\[0.25cm]
{\Large Raghu Rajan} \\[1.25cm]

{\normalsize \textbf{Published in:}} \\[0.25cm]
\Large {
    Albert-Ludwigs-University, Freiburg \\
    Faculty of Engineering \\
    Department of Computer Science \\
    Chair for Machine Learning \\
}

\end{center}
\end{titlepage}

    \begin{flushleft}
    
    \chapter*{Declaration}
    I hereby declare, that I am the sole author and composer of my thesis and that no other sources or learning aids, other than those listed, have been used. Furthermore, I declare that I have acknowledged the work of others by providing detailed references of said work.  \newline
I hereby also declare, that my Thesis has not been prepared for another examination or assignment, either wholly or excerpts thereof.

\vspace*{8em}\noindent
\begin{tabular}{ll}
\makebox[2.5in]{}       & \makebox[2.5in]{}\\
Freiburg, 9-Sept-2021   &                  \\
\rule{0.3\textwidth}{0.4pt} & \rule{0.35\textwidth}{0.4pt} \\[0.1in]
Place, Date             & Victor Vadakechirayath George\\
\end{tabular}
    
    \chapter*{Acknowledgements}
    I would like to thank my parents for their love and support, GPU Cluster support group uni-freiburg for seamless GPU access and Raghu for his insights and supervision.
    
    \chapter*{Abstract}
    \thispagestyle{plain}
\begin{center}
    \Large
    \textbf{Neural architecture impact on identifying temporally extended RL tasks}
    
    \vspace{0.4cm}
    
    Victor Vadakechirayath George
    \vspace{0.4cm}
\end{center}
Inspired by recent developments in attention models for image classification and natural language processing, we present various Attention based architectures in reinforcement learning (RL) domain, capable of performing well on OpenAI Gym Atari-2600 game suite. In spite of the recent success of Deep Reinforcement learning techniques in various fields like robotics, gaming and healthcare, they suffer from a major drawback that neural networks are difficult to interpret. We try to get around this problem with the help of Attention based models. In Attention based models, extracting and overlaying of attention map onto images allows for direct observation of information used by agent to select actions and easier interpretation of logic behind the chosen actions. Our models in addition to playing well on gym-Atari environments, also provide insights on how agent perceives its environment. In addition, motivated by recent developments in attention based video-classification models using Vision Transformer, we come up with an architecture based on Vision Transformer, for image-based RL domain too. Compared to previous works in Vision Transformer, our model is faster to train and requires fewer computational resources. 
    
    \tableofcontents
    
    \listoffigures
    \listoftables
    \listofalgorithms
    
    \pagestyle{plain}
    
    \titleformat{\chapter}%
    {\normalfont\bfseries\Huge}{\thechapter.}{10pt}{}
    
    \pagenumbering{arabic}
    \chapter{Introduction}
    
Reinforcement learning involves mapping situations to actions by an agent, in order to maximize a numerical signal called reward. The agent is rewarded for optimal actions and punished for nonoptimal ones via the scalar reward signal. It eventually learns to act optimally on every situation, by trying all actions and discovering which one yields the most reward \cite{Sutton1998}. Deep reinforcement learning brings Deep learning also into the solution, allowing agents to make decisions from unstructured input data like images, without manual state space engineering. For instance, Deep RL algorithms are now able to consume substantial amount of image pixels rendered to screen in a video game and decide on optimum actions \cite{wiki:deep_reinforcement_learning} based on just the pixel data \cite{Mnih2015}. Deep RL algorithms have been used for a diverse set of applications which includes but not limited to robotics \cite{rl_in_robotics}, video-games(\cite{Mnih2015}, \cite{DBLP:journals/corr/abs-1802-01561}, \cite{DBLP:journals/corr/MnihBMGLHSK16}) and healthcare\cite{DBLP:journals/corr/abs-1908-08796}. \\[0.1in]

\textbf{Black-box nature of Neural networks}\\
Unlike more interpretable models like regression equations and Decision Trees \cite{Quinlan1986}, Neural networks are considered difficult-to-interpret ``black-box" models(Figure \ref{fig:interpretability_models}) (\cite{article:interpretability_models}, \cite{DBLP:journals/corr/MontavonSM17}, \cite{gilpin2019explaining}). Even though the Neural networks can successfully approximate complex functions, examining it's construction does not provide any insights on structure of function being approximated. Consequently, deep reinforcement learning techniques, despite their success in mastering canonical video games like Atari(\cite{Mnih2015}, \cite{DBLP:journals/corr/abs-1802-01561}, \cite{DBLP:journals/corr/MnihBMGLHSK16}), do not provide much insights on how it is able to play the game at a Super-Human level. To an extent, this inhibits the application of reinforcement learning techniques in safety critical real-world applications where trust and reliability are important\cite{DBLP:journals/corr/abs-1711-00138}.\\[0.1in]

\begin{figure}[hbt!]
    \centering
    \includegraphics[width=0.5\textwidth]{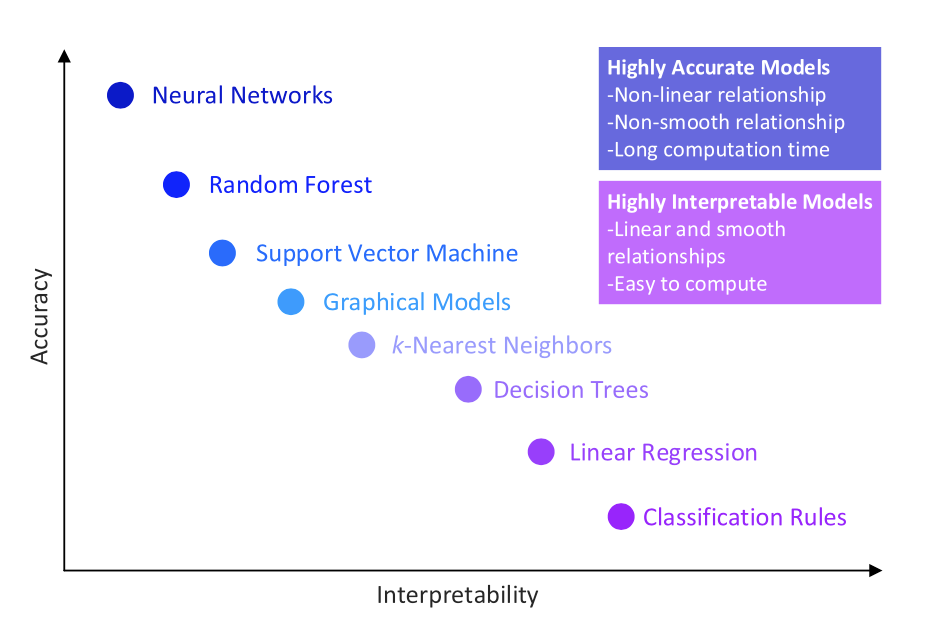}
    \caption{Comparison of interpretability of various machine learning models.(Image source: \cite{article:interpretability_models})}
    \label{fig:interpretability_models}
\end{figure}

\textbf{Enhancing interpretability using Attention mechanisms}\\
Attention mechanisms, which have been proven to work remarkably well with Natural Language Processing \cite{NIPS2017_3f5ee243} and Computer Vision models \cite{dosovitskiy2021an} are changing the way we work with neural networks. Attention is the cognitive process of selectively concentrating on relevant artifacts while ignoring other superficial details. Attention mechanism also attempts to do the same via deep neural networks. In NLP domain, Bahdanau et al.\cite{bahdanau2014neural} proposed to do it by considering all previous input words for building the current context vector, with relative importance being assigned to each one of them. The same procedure could also be extrapolated to the image domain too. Attention probability matrices or attention maps generated during attention calculations, could be extracted and overlaid on the input images to  visualize how every region of image influenced the corresponding output. Analysing output attention vectors allows for direct observation of information used by an agent to select actions and easier interpretation of logic behind the agent's actions. \\[0.1in]

\textbf{Challenges faced during development cycle}\\
In our research, we worked with various attention based architectures ranging from LSTMs\cite{HochSchm97} to the ones using Transformers\cite{NIPS2017_3f5ee243}, to solve gym Atari\cite{brockman2016openai} tasks and model long-term temporal dependency between actions taken and rewards received by RL agent. We kicked off our development with LSTM based model proposed by Mott et al.\cite{DBLP:journals/corr/abs-1906-02500}. Although the model performed reasonably well for relatively easier Atari Enduro environment and toy-environment MinAtar\cite{DBLP:journals/corr/abs-1903-03176}, the LSTM based policy-core was unable to learn long-term temporal dependencies seen in Breakout or Pacman environments. Also, since LSTMs process data sequentially, it is difficult to parallelize the training job which resulted in slower training. To overcome these defects, we adopted and modified Transformer-XL\cite{DBLP:journals/corr/abs-1901-02860} variant-based RL architecture proposed in \cite{DBLP:journals/corr/abs-2004-03761}. In addition to performing well on challenging Atari environments like Pacman and Breakout, the Transformer based models improved on training speed due to their parallel processing capabilities. However, since images were encoded as 1-D tokens during Transformer processing, the possibility of spatial segmentation was unfortunately ruled out with Transformer based model\cite{DBLP:journals/corr/abs-2004-03761}. In order to incorporate both Transformer's faster processing ability and the possibility of spatial segmentation, we came up with a new architecture, replacing LSTM policy core in Mott et al. \cite{DBLP:journals/corr/abs-1906-02500} with a Transformer core. Even though the model performed well in Atari environments, the attention visualizations generated did not explain agent's actions well enough. This lead us to come up with our final model based on Vision Transformer\cite{dosovitskiy2021an}. The ViT based model not only did learn the Atari environments but also produced better quality visualizations. \\[0.1in]

Our key contributions are as follows: \\
\begin{enumerate}
    \item Inspired by existing Attnetion based works in RL (\cite{DBLP:journals/corr/abs-1906-02500}, \cite{DBLP:journals/corr/abs-2004-03761}, \cite{DBLP:journals/corr/abs-1910-06764}), we  propose multiple Attention based temporal architectures, to solve partially-observable time-extended Atari RL environment.
    
    \item Generate good quality spatial and space-time segmentation of the agent's environment using attention maps.
    We try to unveil patterns in the agent's behaviour and help progress towards the goal of reducing ``black box'' nature of neural networks. We also perform perturbation based saliency map analysis proposed in \cite{DBLP:journals/corr/abs-1711-00138} to analyse and verify agent's behaviour.
    
    \item Application of Vision Transformers\cite{dosovitskiy2021an} (ViT) in the reinforcement learning domain. 
    ViT \cite{dosovitskiy2021an} claims performance comparable to state-of-art CNN networks without having the inductive bias seen in CNNs (\cite{DBLP:journals/corr/abs-2102-05095},\cite{dosovitskiy2021an}). To extent of our knowledge, our work is one of the early works using ViT in RL domain.
    
    \item Customize convolution-free video classification work \cite{DBLP:journals/corr/abs-2102-05095} using ViT \cite{dosovitskiy2021an} to function with RL image-based environments. In doing so, we try to establish spatio-temporal similarities between video classification tasks and temporally extended Atari tasks.
    
    \item Finally, compared to existing ViT based models (\cite{dosovitskiy2021an}, \cite{DBLP:journals/corr/abs-2102-05095}), our ViT based model is way less sample-intensive.
    We combined our ViT model with the caching technique proposed in Transformer-XL variant\cite{DBLP:journals/corr/abs-1901-02860} to achieve high sample efficiency. Moreover, we consolidate key network-design decisions and preprocessing techniques from previous works (\cite{Mnih2015}, \cite{DBLP:journals/corr/abs-2004-03761}, \cite{DBLP:journals/corr/abs-1910-06764}) that improved training speed.
\end{enumerate}

    \chapter{Background}
    In this chapter, we present the relevant background information and terminology used in the chapters following. We have restricted the discussion to include only the most essential topics referenced in our architecture. \\[0.1in]

\section{Off-policy Reinforcement Learning}
Our models, similar to the decoupled distributed learning setup seen with (\cite{DBLP:journals/corr/abs-1802-01561}, \cite{DBLP:journals/corr/abs-1906-02500},\cite{DBLP:journals/corr/MnihBMGLHSK16}), use off-policy learning because of the lag between agent's actions and parameter updates.\\[0.1in]

The state-value function of a state \textit{s} under a policy $\pi$, denoted by $v_{\pi}(s)$, is the expected return when starting from \textit{s} and following $\pi$ thereafter \cite{Sutton1998}. Similarly, action-value function for policy $\pi$ denoted by $q_{\pi}(s,a)$, is the expected return when starting from state \textit{s}, taking action \textit{a}, and following policy $\pi$ thereafter \cite{Sutton1998}. RL Agents seek to learn optimum policy by maximising the value functions. But in order to come up with the optimum policy, agent needs to behave non-optimally and explore to find possible optimal actions. Off-policy RL is an approach to address this exploration-exploitation dilemma \cite{Sutton1998}. It uses two policies: \textbf{target policy}-one being learned about, denoted by $\pi$ and \textbf{behaviour policy}-one which is more exploratory in nature, denoted by $b$. Target policy $\pi$ is typically a deterministic greedy policy that eventually becomes the optimal policy, whereas behaviour policy $b$ is more stochastic, exploratory in nature and generates behaviour. In other words, our aim is to estimate $v_{\pi}$ or $q_{\pi}$ with episode-data following the behaviour policy $b$.

\subsection{Importance Sampling (IS)}
Off-policy methods utilize Importance Sampling \cite{Sutton1998} for estimating expected values under target policy $\pi$ given samples under behaviour policy $b$. Importance Sampling ratio denoted by $\rho$, is defined as the relative probability of target and behaviour policy trajectories \cite{Sutton1998}.

\begin{equation}
    \rho_{t:T-1} = \prod_{k=t}^{T-1} \frac{\pi(A_k|S_k)}{b(A_k|S_k)}
\end{equation}
where $A_t, S_{t+1}, A{t+1}, ..., S_{T}$ represents subsequent state-action trajectory for starting state $S_t$.
With $G_t$ as returns due to behaviour policy $b$, transformed returns $v_{\pi}(s)$ from target policy $\pi$ is given by
\begin{equation}
    v_{\pi}(s) = \mathop{\mathbb{E}}[\rho_{t:T-1}*G_t|S_t=s]
\end{equation}

\section{Actor-Critic algorithm}
In Actor-Critic method \cite{Sutton1998}, policy structure used to select actions is referred as \textbf{actor} and the estimated value function which criticizes action made by actor, is referred to as \textbf{critic}. Typically, the critic is a state-value function. Critic criticize current actor's policy and provides a learning signal to actor in form of TD error \cite{Sutton1998} (Figure \ref{fig:actor_crtic}).\\[0.1in]

\begin{figure}[hbt!]
    \centering
    \includegraphics[width=0.4\textwidth]{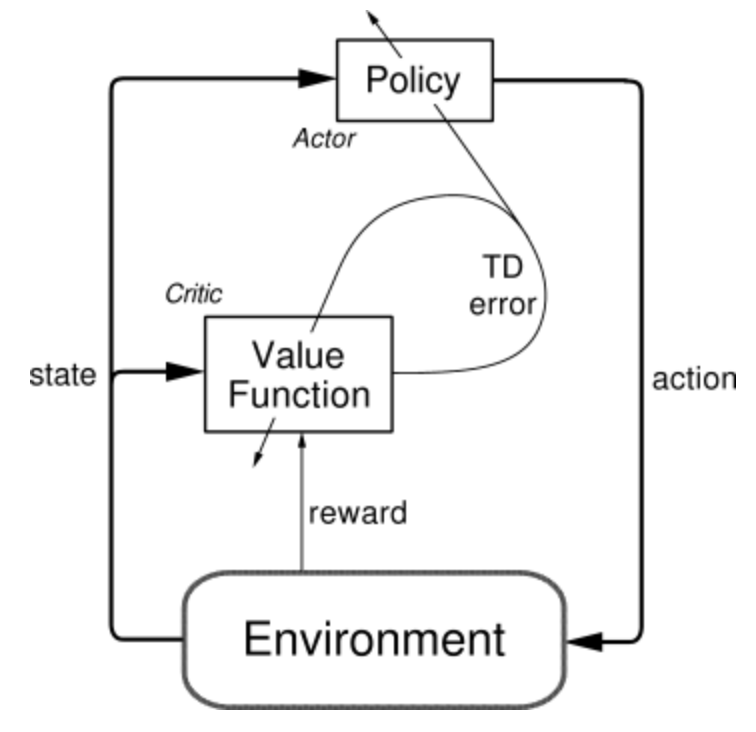}
    \caption{Actor critic architecture. (Image source: \cite{Sutton1998})}
    \label{fig:actor_crtic}
\end{figure}

After each action selection, critic evaluates the new state via TD error, to determine if the feedback is positive or negative. TD error \cite{Sutton1998} is given by

\begin{equation}
    \delta_{t} = r_{t + 1} + \gamma V(s_{t+1}) - V(s_{t})
\end{equation}
where $\gamma$ denotes discount factor and $V$ the current value function implemented by the critic. If TD error is positive, previous action $a_t$ should be promoted and taken more frequently in future. On the other hand, if TD error is negative, $a_t$ should be discouraged. \\[0.1in]

\subsection{Policy Gradient methods}
Policy gradient methods attempt to model and optimize the policy directly \cite{weng2018PG}. Policy is parametrized using $\omega$ as $\pi_{\omega}(a|s)$. Policy gradient theorem \cite{NIPS1999_464d828b} provides a simpler formulation for gradient of the corresponding objective function $J(\omega)$ as,
\begin{equation}
    {\nabla_\omega}J(\omega) \propto \sum_{k\geq0}d^{\pi}(s_k) \sum_{k}Q^{\pi}(s_k,a_k){\nabla_\omega}\pi_{\omega}(a_k|s_k)
\end{equation}

where $d^{\pi}(s_k)$ denotes the state distribution.\\[0.1in]

For on-policy case, as given in IMPALA \cite{DBLP:journals/corr/abs-1802-01561}, gradient of value function $V^{\pi}(x_0)$ with respect to parameter of policy $\pi$ is given by,
\begin{equation}
    {\nabla}V^{\pi}(x_0) = \mathop{\mathbb{E_{\pi}}} \left [\sum_{k\geq0}\gamma^{k}{\nabla}\log \pi(a_k|s_k) Q^{\pi}(s_k,a_k) \right ]
\end{equation}

where $Q^{\pi}(x_k, s_k) = \mathop{\mathbb{E_{\pi}}}\left [\sum_{t\geq k}\gamma^{t-k}r_{t}|s_k, a_k \right]$ is state-action value for policy $\pi$ at $(s_k, a_k)$. Policy parameters are updated in direction of $\mathop{\mathbb{E_{\mathit{a_k} \sim \pi(.|\mathit{s_k})}}}[{\nabla}\log\pi(a_k|s_k)q_k|s_k]$, where $q_k = r_k + {\gamma}v_{k+1}$ is an estimate of $Q^{\pi}(s_k, a_k)$, calculated from V-trace \cite{DBLP:journals/corr/abs-1802-01561} estimate $v_{k+1}$.
In an off-policy setting, the expression is modified to use an importance-sampling weight between target policy $\pi$ and behaviour policy $b$ as follows to update the policy parameters,
\begin{equation}
    \mathop{\mathbb{E_{\mathit{a_k} \sim \mathit{b}(.|\mathit{s_k})}}} \left[\frac{\pi(a_k|s_k)}{b(a_k|s_k)}{\nabla}\log{\pi}(a_k|s_k)q_k|s_k \right]
\end{equation}

In addition, to reduce the variance of the policy gradient estimate, a state-dependent baseline, $V(x_k)$ is subtracted from $q_k$.

\section{Importance Weighted Actor-Learner Architecture (IMPALA)}
IMPALA \cite{DBLP:journals/corr/abs-1802-01561} achieves stable learning at high data throughput by decoupling  acting and learning with a novel off-policy actor-critic algorithm called V-trace, for distributed actor-learner architecture. IMPALA actors ship out trajectories of experience (sequence of state-action-reward tuples) to a centralised learner (Figure \ref{fig:impala_single_learner}). Since the learner has access to the full experience trajectory, GPUs are used to aggressively parallelize mini-batch updates, thereby ensuring high throughput. \\[0.1in]
Since the actor's policy used to generate a trajectory can lag behind learner's policy by several updates, learning becomes off-policy. The harmful discrepancy is resolved by introducing \textbf{V-trace} off-policy actor-critic algorithm \cite{DBLP:journals/corr/abs-1802-01561}.

\begin{figure}[hbt!]
    \centering
    \includegraphics[width=0.4\textwidth]{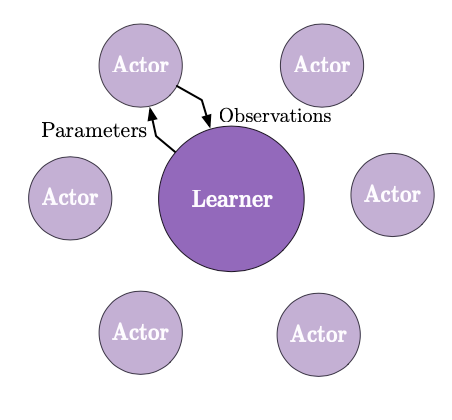}
    \caption{Individual actors generate and send trajectories via a queue to the central-learner. Actor retrieves latest policy parameters from the learner before starting next trajectory. (Image source: \cite{DBLP:journals/corr/abs-1802-01561})}
    \label{fig:impala_single_learner}
\end{figure}

\subsection{V-trace target}
Off-policy learning is necessary for distributed actor-learner architecture due to lag between actors' actions and parameter update via learner's gradient calculation. \\[0.1in]
Consider an off-policy RL setting with trajectory $(x_t, a_t, r_t)_{t=k}^{t=k+n}$ generated by actors following policy $b$, where $x_{t}$, $a_{t}$ and $r_{t}$ denotes state, action and reward at time $t$. $n$-steps V-trace target \cite{DBLP:journals/corr/abs-1802-01561} for $V(s_k)$ or value approximation at state $s_k$, denoted by $v_k$, is defined as,

\begin{equation}
    \label{eq:v_trace_target_eqn}
    \textit{v}_k = V(s_k) + \sum_{t=k}^{k+n-1}\gamma^{t-k}(\prod_{i=k}^{t-1}c_i)\delta_{t}V
\end{equation}
where $\delta_{t}V = \rho(r_t + {\gamma}V(s_{t+1}) - V(s_t))$ is temporal difference for V and $\rho_t = \texttt{min}(\overline{\rho}, \frac{\pi(a_t|s_t)}{b(a_t|s_t)})$, $c_i = \texttt{min}(\overline{c}, \frac{\pi(a_i|s_i)}{b(a_i|s_i)})$ are truncated importance sampling weights. The truncation levels are set such that $\overline{\rho} \geq \overline{c}$. \\[0.1in]
Considering the special on-policy case (when $\pi = b$) and assuming $\overline{c} \geq 1$, then all $c_i = 1$ and $\rho_t = 1$, Eq  \ref{eq:v_trace_target_eqn} becomes, 
\begin{equation}
    \textit{v}_k = V(s_k) + \sum_{t=k}^{k+n-1}\gamma^{t-k}(r_t + {\gamma}V(s_{t+1}) - V(s_t))
\end{equation}

which is same as on-policy \textit{n}-steps Bellman\cite{Sutton1998} target. Hence, on-policy \textit{n}-steps Bellman update is a special case for V-trace target. \\[0.1in]

The product $c_k ... c_{t-1}$ measures the impact of temporal difference term ${\delta_t}V$ at time $t$ on update of value function at a previous time $k$. Variance of this product term increase with how off the policies $\pi$ and $b$ are. Truncation levels $\overline{\rho}$ and $\overline{c}$ represents different features of algorithm: $\overline{\rho}$ influence nature of value function we converge to and $\overline{c}$ determines speed at which we converge to this function \cite{DBLP:journals/corr/abs-1802-01561}. \\[0.1in]
\textbf{Remark 1.} V-trace target $v_k$ is computed in the algorithmic implementation using the following recursive expression,

\begin{equation}
    v_k = V(s_k) + \delta_{k}V + {\gamma}c_{k}(v_{k+1} - V(s_{k+1}))
\end{equation}

\subsection{V-trace Actor-Critic algorithm} \label{section:monobeast_loss_function}
Consider value function function $V_{\theta}$ and current target policy $\pi_{\omega}$ parametrised by $\theta$ and $\omega$ respectively. Actors following behaviour policy $b$ generate trajectories. Value parameter $\theta$ is updated by gradient descent on $L2$ loss with respect to target $v_k$ in direction,
\begin{equation}
    (v_k - V_{\theta}(s_k)){\nabla_{\theta}}V_{\theta}(s_k)
\end{equation}
Also the policy parameters $\omega$ are updated in the direction of policy gradient,
\begin{equation}
    \rho_{k}\nabla_{\omega}\log\pi_{\omega}(a_k|s_k)(r_k + {\gamma}v_{k+1} - V_{\theta}(s_k))
\end{equation}
Similar to A3C \cite{DBLP:journals/corr/MnihBMGLHSK16}, an entropy loss term is also included in the total loss to avoid premature convergence and to encourage agent to explore more\cite{doi:10.1080/09540099108946587}. Entropy loss gradient is given as,
\begin{equation}
    -\nabla_{\omega}\sum_{a}\pi_{\omega}(a|s_k)\log\pi_{\omega}(a|s_k)
\end{equation}

The three gradients are scaled by their respective importance hyper-parameters and summed to get the total gradient for parameter update.

\section{TorchBeast: PyTorch platform for Distributed RL}
TorchBeast \cite{torchbeast2019} is a PyTorch implementation of IMPALA  \cite{DBLP:journals/corr/abs-1802-01561} for fast, asynchronous, parallel training of RL agent. The algorithm uses an off-policy method with a behaviour policy $b$ for collecting experience and a target policy $\pi_\omega$ which is being updated. As explained in previous section, target policy $\pi_{\omega}$ and value function estimate $V_{\theta}$ parameters are updated using Policy Gradient method with V-trace as off-policy correction method.

\subsection{TorchBeast data pipeline}
TorchBeast \cite{torchbeast2019} architecture consist of a single learner and multiple actors producing episode roll-outs in an indefinite loop similar to Figure \ref{fig:impala_single_learner}. One roll-out consist of \texttt{unroll{\textunderscore}length} number of environment-agent interactions. Batches of roll-outs are packed in a Python dictionary and fed to the learner. In order to ensure high throughput, the number of actors should be set large enough such that batches of experience are generated fast and the learner GPU is fully utilized. Typical learner input-dictionary is of the format:

\begin{verbatim}
    {
      "observation": tensor(T, B, *obs_shape, dtype=torch.uint8),
      "reward": tensor(T, B, dtype=torch.float),
      "done": tensor(T, B, dtype=torch.uint8),
      "policy_logits": tensor(T, B, num_actions, dtype=torch.float),
      "baseline": tensor(T, B, dtype=torch.float),
      "actions": tensor(T, B, dtype=torch.int8),
    }
\end{verbatim}
where \texttt{tensor(T, B)} denotes tensor of shape \texttt{(T, B)} with \texttt{T} as unroll length and \texttt{B} as batch size, \texttt{obs{\textunderscore}shape} represents observation shape tuple or frame dimensions \texttt{(H,W)}, and \texttt{num{\textunderscore}actions} denotes number of possible actions.

\subsection{MonoBeast algorithm}
MonoBeast is a lighter version of TorchBeast\cite{torchbeast2019} which performs actor evaluations on CPU instead of GPU and the learner runs on a single GPU. Monobeast runs on a single machine and requires a relatively large amount of constantly allocated shared memory. The algorithm is detailed out in [\ref{alg:monobeast_data_processing}].\\[0.1in]

\begin{algorithm}
    \caption{MonoBeast data processing (Source: \cite{torchbeast2019})}
    \label{alg:monobeast_data_processing}
    \begin{algorithmic}[1]
    \STATE Create  \texttt{num{\textunderscore}buffers} sets of rollout buffers, each of them a dictionary with keys being: \texttt{"observation"}, \texttt{"reward"}, \texttt{"done"}, \texttt{"policy\textunderscore logits"}, \texttt{"baseline"}, \texttt{"actions"} and values the respective shared-memory tensors.
    \STATE Create two shared queues \texttt{free{\textunderscore}queue} and \texttt{full{\textunderscore}queue} which will exchange integers amongst them using UNIX pipes. Each integer represents buffer number of either a fully filled or used (already trained) roll-out buffer. 
    \STATE Start \texttt{num{\textunderscore}actors} many actor process, each with it's own copy of environment. Actors dequeue an index \texttt{idx} from \texttt{free{\textunderscore}queue} and writes rollout data into shared rollout buffers created in step(1). Once rollout is done, index \texttt{idx} is enqueued to the \texttt{full{\textunderscore}queue} and next index is dequeued from the \texttt{free{\textunderscore}queue}.
    \STATE Main learner thread does the following: 
    \begin{enumerate}
        \item Dequeues \texttt{batch{\textunderscore}size} number of indices from \texttt{full{\textunderscore}queue}, feeds them to learner model running on GPU and puts dequeued indices back to \texttt{free{\textunderscore}queue}.
        \item Sends dequeued batch through the model, compute losses, does backward pass, and update the weights. 
    \end{enumerate}
    
    \STATE Actor-models are loaded with the latest weights from the learner-model once the weight update is done.
    \end{algorithmic}
\end{algorithm}

\section{LSTM networks}
Architecture proposed in Mott et al. \cite{DBLP:journals/corr/abs-1906-02500} on which we are basing our initial experiments, uses LSTM \cite{HochSchm97} networks as policy-core to model time dependencies. Long short-term memory (LSTM) \cite{HochSchm97} belongs to the family of recurrent neural network architectures and are capable of learning long-term dependencies. LSTM has the ability to remove or add information to cell state, regulated by gates \cite{colah:lstm_tutorial}. A typical gate consist of a sigmoid neural net layer followed by a pointwise multiplication operation. A common LSTM unit (Figure \ref{fig:lstm_structure}) has three gates: input, output and forget gates. The cell remembers previous states over arbitrary time intervals and the three gates control the flow of information in and out of the cell \cite{wiki:lstm_tutorial}. LSTMs address the vanishing gradient problem observed in RNNs to an extent, by allowing gradients to flow unchanged across input-output gates \cite{HochSchm97:lstm_vanishing_gradient_solution}. \\[0.1in]
The sigmoid layer considers previous hidden state $\textit{h}_{t-1}$, current input $\textit{x}_t$ and updates forget ratio $f_t$ as given in Eq \ref{eq:lstm_cell_state_update_1}.

\begin{equation}
    \label{eq:lstm_cell_state_update_1}
    f_t = \sigma(W_f.[\textit{h}_{t-1}, \textit{x}_t] + b_f)
\end{equation}

With $\textit{i}_t$ given by Eq \ref{eq:i_t_update} and $\tilde{C}_t$ by Eq \ref{eq:C_t_update}, old cell state $\textit{C}_{t-1}$ is updated into new cell state $\textit{C}_t$ as given by Eq \ref{eq:lstm_cell_state_update_2}, where old states are forgotten via scaling with forget ratio $f_t$ and new states $\tilde{C}_t$ included via scaling with $i_t$.

\begin{equation}
    \label{eq:lstm_cell_state_update_2}
    C_t = f_t * C_{t-1} + i_t * \tilde{C}_t
\end{equation}

\begin{equation}
    \label{eq:i_t_update}
    i_t = \sigma(W_i.[\textit{h}_{t-1}, \textit{x}_t] + b_i)
\end{equation}

\begin{equation}
    \label{eq:C_t_update}
    \tilde{C}_t = \tanh(W_c.[\textit{h}_{t-1}, \textit{x}_t] + b_c)
\end{equation}

With the output ratio $o_t$ given by sigmoid gate Eq \ref{eq:o_t_update}, new hidden state $h_t$ will be a filtered version of updated cell state $\textit{C}_t$ given by Eq \ref{eq:lstm_output_update}
\begin{equation}
    \label{eq:lstm_output_update}
    h_t = o_t * \tanh(C_t)
\end{equation}

\begin{equation}
    \label{eq:o_t_update}
    o_t = \sigma(W_o.[\textit{h}_{t-1}, \textit{x}_t] + b_o)
\end{equation}

\begin{figure}[hbt!]
    \centering
    \includegraphics[width=0.5\textwidth]{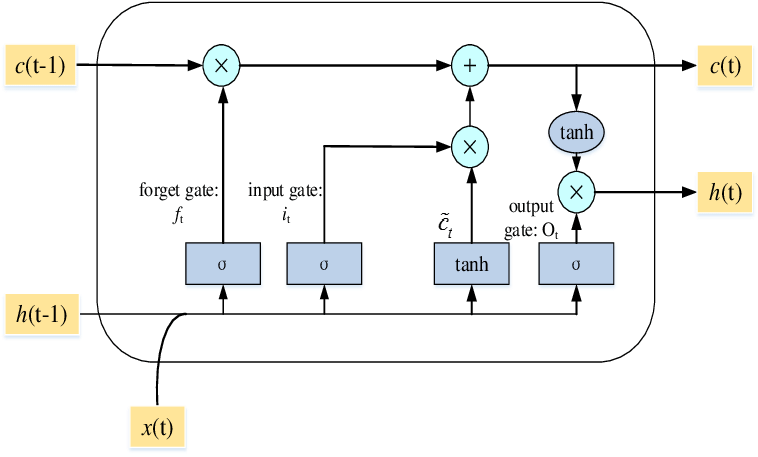}
    \caption{LSTM structure. (Image source \cite{8654687})}
    \label{fig:lstm_structure}
\end{figure}

\subsection{Convolutional LSTM}
Fully connected LSTM suffers from a major drawback that it does not encode the spatial-structure information of input data in the information flow \cite{DBLP:journals/corr/ShiCWYWW15}. In order to overcome this defect, fully connected LSTMs are extended to have a convolutional structure in Convolutional LSTM(ConvLSTM) \cite{DBLP:journals/corr/ShiCWYWW15}. \\[0.1in]
In ConvLSTM, input $X_1, ..., X_t$, cell outputs $C_1, ..., C_t$, hidden states $H_1, ..., H_t$ and gates $i_t$, $f_t$, $o_t$ are all 3D tensors whose last two dimensions are spatial dimensions (height and width). ConvLSTM captures spatio-temporal correlations better than fully connected LSTM \cite{DBLP:journals/corr/ShiCWYWW15}.

\section{Self-Attention} \label{sec:transformer_self_attention_calculation}
Self attention \cite{NIPS2017_3f5ee243} relates different position of a single sequence to compute a representation of the same sequence. Attention function maps a query $\texttt{Q}_i$ and key-value pair $(\texttt{K}_i, \texttt{V}_i)$ to an output which is the weighted sum of values $\texttt{V}_i$. Weight assigned to each value $\texttt{V}_i$ represents compatibility of query $\texttt{Q}_i$ with corresponding key $\texttt{K}_i$.

\subsection{Scaled dot product attention}
Input consists of a set of queries packed into a matrix \textbf{Q} $\in  \mathbb{R}^{N_q \times d_k}$. Keys and Values are packed respectively into matrix \textbf{K} $\in  \mathbb{R}^{N_k \times d_k}$ and \textbf{V} $\in  \mathbb{R}^{N_k \times d_v}$. Attention matrix is given by matrix dot product of \textbf{Q} and \textbf{K}, scaled by $1/\sqrt[]{d_k}$ (Eq \ref{eq:attn_q_k_product}).

\begin{equation}
    \label{eq:attn_q_k_product}
    \texttt{Attention}(\mathbf{Q}, \mathbf{K}, \mathbf{V}) = \texttt{softmax}(\frac{\mathbf{Q}\mathbf{K^T}}{\sqrt[]{d_k}}).\mathbf{V}
\end{equation}
Final output \textbf{Y} given by Eq \ref{eq:attn_output_AV}, is a weighted average of \textbf{V}
\begin{equation}
    \label{eq:attn_output_AV}
    \mathbf{Y} = \mathbf{A}\mathbf{V}
\end{equation}

where \textbf{A} $\in  \mathbb{R}^{N_q \times N_k}$ denotes attention matrix calculated in Eq \ref{eq:attn_q_k_product}, \textbf{Y} $\in  \mathbb{R}^{N_q \times d_v}$ the dot-product output of \textbf{A} and \textbf{V}. Softmax operation is applied to scaled dot-product to calculate attention scores or weights corresponding to each key. For large values of $\textit{d}_k$, dot product grows in magnitude and pushes softmax to regions of extremely small gradient which is prevented by scaling with $1/\sqrt[]{d_k}$ \cite{NIPS2017_3f5ee243}. Block diagram of the attention calculation is given in Figure \ref{fig:attention_mechanisms} left.

\begin{figure}[hbt!]
\centering
\begin{subfigure}{0.3\textwidth}
    \centering
    \includegraphics[width=1.0\linewidth]{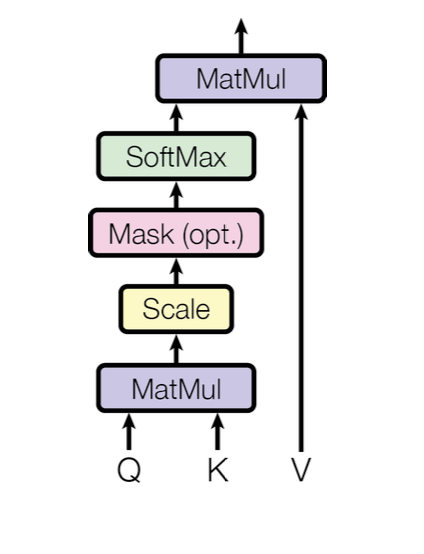}
    \label{fig:scaled_dot_product_attention}
\end{subfigure}%
\hspace{0.1\textwidth}
\begin{subfigure}{0.3\textwidth}
    \centering
    \includegraphics[width=1.0\linewidth]{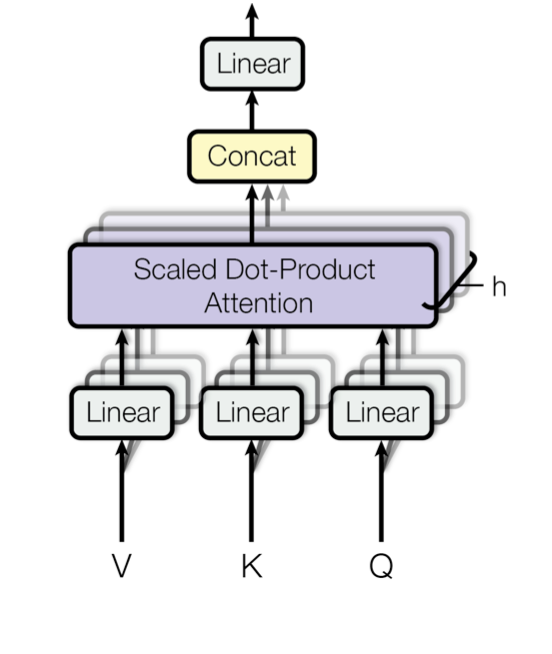}
    \label{fig:multi_head_attention}
\end{subfigure}%
\caption{Left:Scaled dot-product attention and Right:Multi-Head Attention mechanism. (Image source: Vaswani et al.
\label{fig:attention_mechanisms}
\cite{NIPS2017_3f5ee243})}
\end{figure}

\subsection{Multi-Head Attention}
Multi-Head Attention (Figure \ref{fig:attention_mechanisms}, Right) consist of multiple attention layers running in parallel. It enables model to attend to multiple representation sub-spaces at different positions simultaneously. This cannot be achieved using single attention head due to the averaging effect \cite{NIPS2017_3f5ee243}.
\begin{equation}
    \texttt{MultiHead}(\mathbf{Q}, \mathbf{K}, \mathbf{V}) = [\texttt{head}_{1};...;\texttt{head}_{h}]
\end{equation}
where $\texttt{head}_{i} = \texttt{Attention}(\mathbf{Q}, \mathbf{K}, \mathbf{V})$

\section{Transformers}
LSTMs \cite{HochSchm97} and gated \cite{DBLP:journals/corr/ChungGCB14} recurrent neural networks have been considered state-of-art approaches for modelling long-term dependencies. Recently, due to their ability to integrate information over longer time horizons and process massive amounts of data quickly, self-attention architectures, mainly Transformers, have made their way into domains like natural language processing and machine translation \cite{bahdanau2014neural}. There have been previous works (\cite{DBLP:journals/corr/abs-1910-06764} and \cite{DBLP:journals/corr/abs-2004-03761}) where Transformers were successfully applied to partially observable RL problems, where episodes extend to more than thousands of steps and critical observation associated with an action could span the entire episode. \\[0.1in]
Transformers unlike LSTMs, do not compress the entire history of the model into fixed-size hidden states. Recurrent models generate a sequence of hidden states $\textit{h}_t$ as a function of previous hidden state $\textit{h}_{t-1}$ and current input $\textit{x}_t$. This sequential nature of data processing rules out the possibility of parallelization with training data and hence maximum utilization of computational resources, which becomes critical at longer sequence lengths \cite{NIPS2017_3f5ee243}.

\subsection{Transformer-Architecture}
Vanilla transformer \cite{NIPS2017_3f5ee243} have an encoder-decoder architecture (Figure \ref{fig:transformer_architecture}). Model auto-regressively process the previous outputs along with the current input to generate the next output.\\[0.1in]

\begin{figure}[hbt!]
    \centering
    \includegraphics[width=0.6\textwidth]{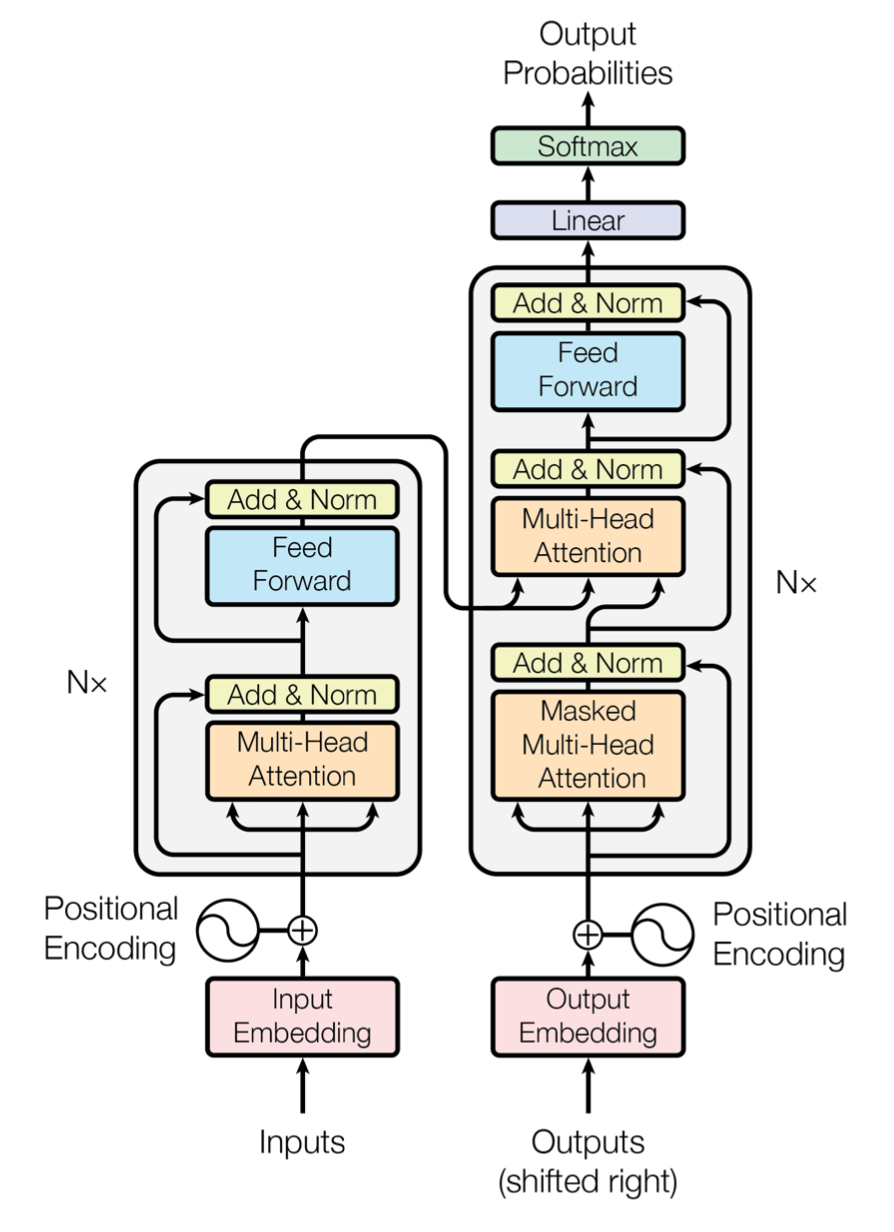}
    \caption{Vanilla-transformer full model architecture from Vaswani et al. \cite{NIPS2017_3f5ee243}}
    \label{fig:transformer_architecture}
\end{figure}

\textbf{Encoder}\\
Encoder composed of $N = 6$ identical layers, generates an attention-based representation of input. It maps an input sequence $(x_1,...,x_n)$ to continuous sequence representation \textbf{z} = $(z_1,...,z_n)$. Each layer has two sublayers: first one, a multihead self-attention (MHA) layer and the second, a fully connected feed-forward network. A residual \cite{DBLP:journals/corr/HeZRS15} connection followed by layer normalization \cite{ba2016layer} is employed around each sublayer. \\[0.1in]

\textbf{Decoder}\\
Decoder consist of $N = 6$ identical layers with similar sublayer structure similar to Encoder. Provided \textbf{z}, decoder generates an output sequence $\mathbf{y} = (y_1,...,y_m)$ one by one auto-regressively using previous time-step outputs. Compared to Encoder, there is an additional masked MHA layer with look-ahead mask present in Decoder, which attends to it's own output. Query \textbf{Q} receives output from the additional masked MHA layer(Figure \ref{fig:transformer_architecture} Right) and \textbf{K,V} receives the Encoder output to calculate attention weights. Weights so calculated represent importance given to Decoder's input based on Encoder's current output. The Decoder predicts the next token by looking at the Encoder output and self-attending to its own output \cite{tf_tranformers_tutorial}. Output embeddings are offset by one position and look ahead masking is applied such that, for predicting at position \textit{i}, only tokens till position \textit{i} are used. \\[0.1in]

\textbf{Positional Encoding} \\
Positional encoding is added to the embedded vector in order to give model some information about the relative position of the tokens in the sequence. An embedding or token represent a word or an image in a d-dimensional space such that the tokens corresponding to similar words or images would be closer to each other. Relative positions of the tokens are encoded by adding positional encoding to the tokens. Hence, after the addition, tokens will be closer to each other based on not only their content's embedding similarity, but also on their temporal correlation. Positional encodings (Figure \ref{fig:encodings_transformer_xl} Left) are initialized with following sinusoidal functions

\begin{equation}
    \texttt{PE}_{(pos, 2i)} = \sin(pos/10000^{2i/d_{model}})
\end{equation}
\begin{equation}
    \texttt{PE}_{(pos, 2i+1)} = \cos(pos/10000^{2i/d_{model}})
\end{equation}

where \textit{pos} varies from 0 to length of the sequence considered, \textit{i} varies from 0 to $d_{model}$. In Figure:\ref{fig:encodings_transformer_xl}, \textit{pos} varies along the y-axis and \textit{i} along the x-axis. Positional encoding also belongs to the total set of trainable parameters of the model. On Figure \ref{fig:encodings_transformer_xl} Right, encodings are visualized after training. Adjacent encoding tokens along y-axis look similar and the dissimilarity between them grows with their relative separation  along y-axis(Figure: \ref{fig:encodings_transformer_xl}).

\begin{figure}[hbt!]
    \centering
    \includegraphics[width=1.0\textwidth]{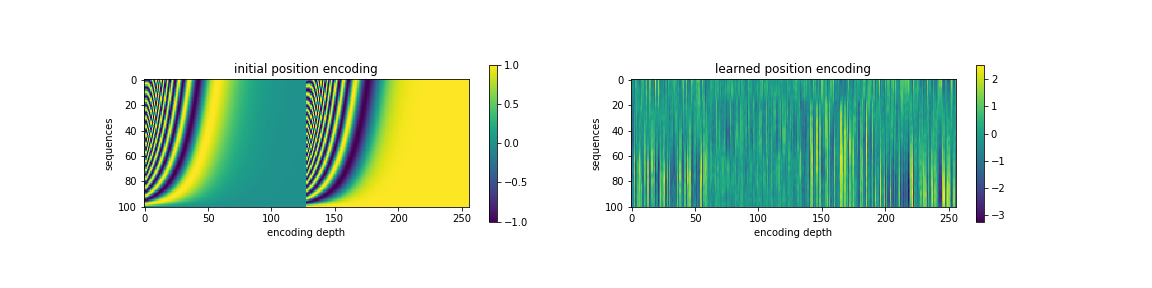}
    \caption{Position encodings corresponding to a sequence length of 110 consecutive tokens and encoding dimension 256. Left: Initial positional encoding. Right: Positional encoding after training.}
    \label{fig:encodings_transformer_xl}
\end{figure}

\subsection{Transformer-XL}
\label{sec:transformer_xl_background}
Transformer-XL \cite{DBLP:journals/corr/abs-1901-02860} belonging to general family of Transformers, allows for learning long-term dependencies beyond a fixed context length. In NLP domain, Transformer-XL outperforms both RNN and vanilla Transformer \cite{NIPS2017_3f5ee243} by a big margin \cite{DBLP:journals/corr/abs-1901-02860}. \\[0.1in]

\begin{figure}[hbt!]
    \centering
    \includegraphics[width=0.8\linewidth]{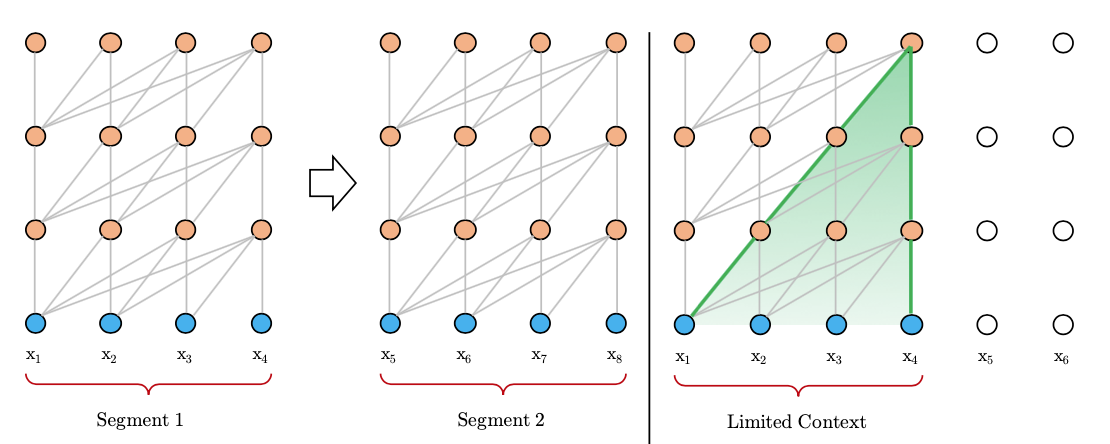}
    \caption{Data processing as proposed in vanilla transformer \cite{NIPS2017_3f5ee243}. Information never flows across segments in either forward or backward pass, thereby upper bounding largest possible dependency length to segment length.(Image source: \cite{DBLP:journals/corr/abs-1901-02860})}
    \label{fig:vanilla_transformer_processing}
\end{figure}

\textbf{Issues with vanilla Transformers} One major drawback with vanilla Transformers \cite{NIPS2017_3f5ee243} is that self-attention is performed only over separate fixed-length segments, without any information flowing across the segments. Hence, the model is unable to capture long-term dependencies beyond the predefined context length (Figure \ref{fig:vanilla_transformer_processing}). \\[0.1in]
\textbf{Method: Segment-level recurrence with state reuse}  In order to resolve this issue, Transformer-XL, during training, fixes and caches hidden state sequences from previous segment to be reused for the next cycle (Figure \ref{fig:transformer_xl_processing}). The cached hidden states serve as extended context for the current segment and hence introduce a recurrent connection between segments \cite{DBLP:journals/corr/abs-1901-02860}. \\[0.1in]

\begin{figure}[hbt!]
    \centering
    \includegraphics[width=1.0\linewidth]{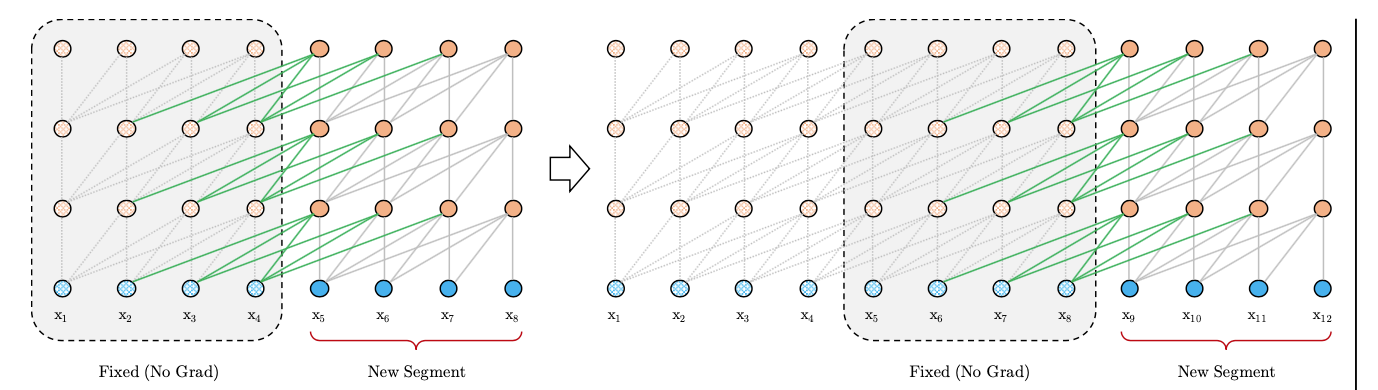}
    \caption{Data processing scheme with segment-level recurrence as proposed in Transformer-XL \cite{DBLP:journals/corr/abs-1901-02860}.(Image source \cite{DBLP:journals/corr/abs-1901-02860})}
    \label{fig:transformer_xl_processing}
\end{figure}

Consider two consecutive segments $s_{\tau} = [x_{\tau, 0}, ..., x_{\tau, L-1}]$ and $s_{\tau+1} = [x_{\tau+1, 0}, ..., x_{\tau+1, L-1}]$. Let $\mathbf{h}_{\tau}^n$ $\in  \mathbb{R} ^{L\times d}$ be Transformer's \textit{n}-th layer
hidden state sequence produced by $\tau$-th segment $s_{\tau}$. Then, \textit{n}-th hidden state for segment $s_{\tau+1}$ is generated as follows,
\begin{equation}
    \tilde{\mathbf{h}}_{\tau+1}^{n-1} = \left[ \texttt{SG}(\mathbf{h}_{\tau}^{n-1}) \circ \mathbf{h}_{\tau+1}^{n-1} \right]
\end{equation}
where $\tilde{h}$ represents the extended context, \texttt{SG}(.) denotes stop-gradient on the cached previous segment $h_{\tau}^{n-1}$, $h_{\tau+1}^{n-1}$ the current segment and $[h_u \circ h_v]$ denotes concatenation of two hidden state sequence along sequence-length axis. Query, key, values are computed using model parameters \textbf{W} as follows: \\[0.1in]
\begin{align*}
    \mathbf{q}_{\tau+1}^{n} = \tilde{h}_{\tau+1}^{n-1} \mathbf{W}_{q}^{T} \\
    \mathbf{k}_{\tau+1}^{n} = \tilde{h}_{\tau+1}^{n-1} \mathbf{W}_{k}^{T} \\
    \mathbf{v}_{\tau+1}^{n} = \tilde{h}_{\tau+1}^{n-1} \mathbf{W}_{v}^{T}
\end{align*}
Unlike in vanilla transformers, key $\mathbf{k_{\tau+1}^n}$ and value $\mathbf{v_{\tau+1}^n}$ are conditioned on extended context $\tilde{h}_{\tau+1}^{n-1}$ which in turn depends on cached previous segment $h_{\tau}^{n-1}$. This recurrence relation is denoted by green lines in figure \ref{fig:transformer_xl_processing}. Applying the recurrence mechanism to every two adjacent segments introduce a segment-level recurrence resulting in effective context extending beyond just the considered two segments. Finally, hidden state for next layer is given by,
\begin{equation}
    {h}_{\tau+1}^n = \texttt{TransformerLayer}(\mathbf{q}_{\tau+1}^{n}, \mathbf{k}_{\tau+1}^{n}, \mathbf{v}_{\tau+1}^{n})
\end{equation}

\section{Vision Transformers}
\label{sec:vision_transformer_background}
Vision Transformer(ViT) introduced by Dosovitskiy et al. \cite{dosovitskiy2021an} presents an alternative, pure transformer based approach for image classification compared to traditional methods using CNNs. Naive application of self-attention to images would result that each pixel attends to every other pixel. For reasonable resolution images, this approach would not be scalable due to $\mathcal{O}(N^2)$ cost in the number of pixels. In ViT \cite{dosovitskiy2021an}, a feasible alternative is proposed to get around this issue. Even though the model required pre-training with large amount of data, ViT achieved state-of-art convolutional networks performance using fewer training resources.\\[0.1in]
\textbf{Method} \\
In ViT \cite{dosovitskiy2021an}, an image is split into equal-sized patches (Figure \ref{fig:vit_architecture}). Each of the patches are linearly embedded into a higher dimensional space and summed with parametrised position encoding to inject position information into the token sequence. The resulting sequence of patch embeddings is fed into a standard Transformer \cite{NIPS2017_3f5ee243}. Patch embeddings are analogous to word tokens in an NLP application where the input is a 1D sequence of token embeddings. \\[0.1in]
Input image \textbf{x} $\in  \mathbf{R}^{H \times W \times C}$ is reshaped into a sequence of flattened 2D patches $\mathbf{x}_p$ $\in  \mathbb{R}^{N \times (P^2 \cdot C)} $, where $(H \times W)$ denotes the resolution of the original image, \textit{C} the number of channels, $(P \times P)$ the resolution of image patch and $N = HW/P^2$ denotes the resulting number of patches or effective input sequence length for the Transformer. Flattened patches are linearly mapped into \textit{D} dimensioned latent space using trainable matrix \textbf{E} $\in  \mathbb{R}^{(P^2 \cdot C) \times D}$. \\[0.1in]

\begin{figure}[hbt!]
    \centering
    \includegraphics[width=1.0\textwidth]{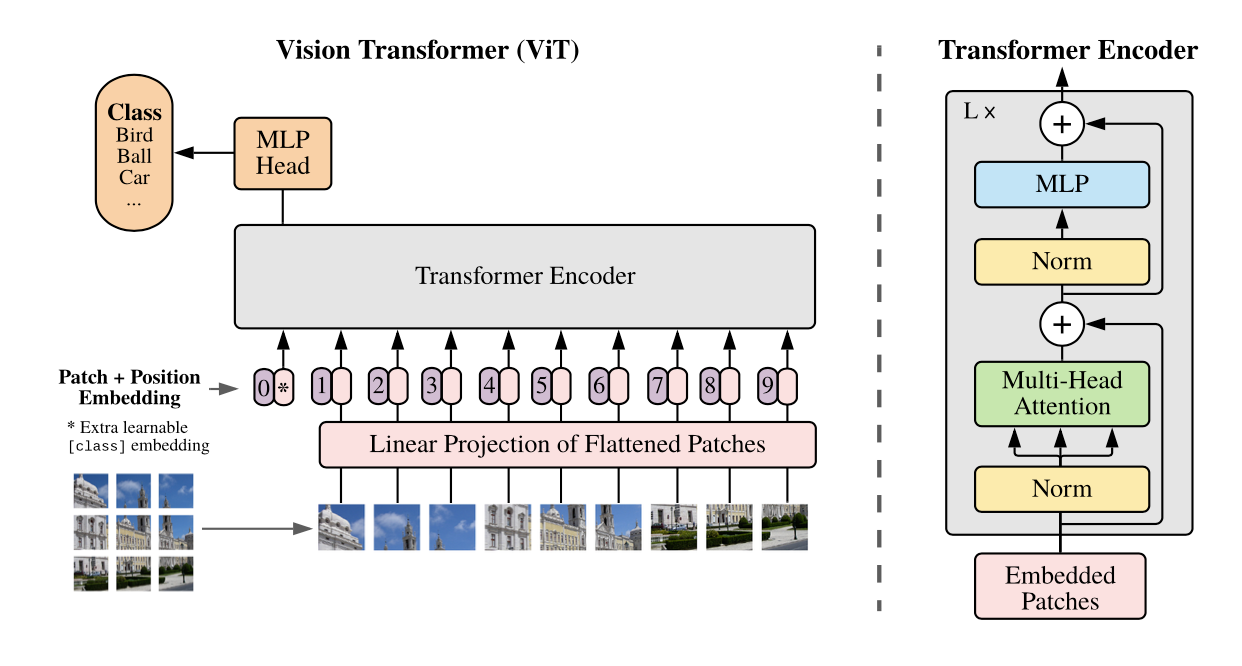}
    \caption{Left: ViT model architecture as proposed as proposed in \cite{dosovitskiy2021an}. Right: Transformer encoder architecture used. (Image source \cite{dosovitskiy2021an})}
    \label{fig:vit_architecture}
\end{figure}

Similar to BERT's \cite{DBLP:journals/corr/abs-1810-04805} classification token, the author's prepend a learnable classification token (${\mathbf{z_{0}^0}} = \mathbf{x}_{class}$) to sequence of embedded patches. Classification-token's value at output of last Transformer-encoder layer ($\mathbf{z}_{L}^0$) serves as the corresponding image representation and fed into the final classification head. The classification head is usually implemented by a single hidden layer MLP. \\[0.1in]
1D Positional embeddings are added to the patch embeddings to retain positional information. Resulting sequence of embedding vectors are fed into a multilayer Transformer-encoder consisting of alternate sublayers of multiheaded self-attention(MHA) and MLP layers. \\[0.1in]

\textbf{Architecture modifications compared to Vanilla-transformer} \\
Compared to conventional transformers \cite{NIPS2017_3f5ee243}, authors slightly modified the encoder architecture such that the LayerNorm (LN) is applied before every MHA-MLP sublayer, instead of after, followed by residual \cite{DBLP:journals/corr/HeZRS15} connections. Coupled with residual connections \cite{DBLP:journals/corr/HeZRS15}, there exist now a gradient path that flows from output to input without any transformations. The same idea has also been proposed by Parisotto et al. in  \cite{DBLP:journals/corr/abs-1910-06764}. Major advantage of this reordering is that it allows for an identity mapping from the input of the transformer at the first layer to the output after last layer. The MLP blocks (Figure \ref{fig:vit_architecture}) consist of two linear layers with GELU nonlinearity \cite{DBLP:journals/corr/HendrycksG16} in between. \\[0.1in]

\begin{equation}
    \mathbf{z_0} = [\mathbf{x_{class}}; \mathbf{x}^1\mathbf{E};  \mathbf{x}^2\mathbf{E}; ... ; \mathbf{x}^N\mathbf{E}] + \mathbf{E}_{pos}
\end{equation}
where, \textbf{E} $\in \mathbf{R}^{(P^2 \cdot C) \times D}$ denotes patch embedding projection, $(\mathbf{x}^1, \mathbf{x}^2, ..., \mathbf{x}^N)$ denotes N patches of size $(P \times P)$, $\mathbf{x_{class}} = \mathbf{z_{0}^0}$ denotes the classification token and $\mathbf{E}_{pos}$ $\in \mathbf{R}^{(N+1) \times D}$ denotes position encoding.

\begin{equation}
    \mathbf{z}_l' = \texttt{MHA}(\texttt{LN}(\mathbf{z}_{l-1})) + \mathbf{z}_{l-1}, \quad l=1...L
\end{equation}

\begin{equation}
    \mathbf{z}_l = \texttt{MLP}(\texttt{LN}(\mathbf{z}_l')) + \mathbf{z}_l', \quad l=1...L
\end{equation}

where $l$ denotes the layer number varying from 1 to $L$, $L$ is the total number of encoder layers. The final classification output is extracted from the classification token in last layer $\mathbf{z}_L ^ 0$ as follows:
\begin{equation}
    \mathbf{y} = \texttt{LN}(\textbf{z}_L ^ 0)
\end{equation}

\textbf{Hybrid Architecture} \\
Raw image patches of an image are generated by passing an image through the patch-embedding network which is basically a single layer 2D convolutional neural network (CNN) with stride value set the same as the kernel width. Hence, in normal architecture, patch size is same as kernel width of patch-embedding CNN network. \\[0.1in]
Instead of extracting raw image patches from simple CNN layer, the authors \cite{dosovitskiy2021an} propose a hybrid architecture using a more advanced CNN, like Resnet \cite{DBLP:journals/corr/HeZRS15}. In the hybrid model, the input sequence is obtained by simply flattening the spatial dimension of Resnet's feature map. It could be considered as a special case of normal architecture with image dimensions matching Resnet's feature map's dimensions and patch size being $(1 \times 1)$. \\[0.1in]

\textbf{GELU activation function} \\
Gaussian Error Linear Unit (GELU) \cite{DBLP:journals/corr/HendrycksG16}, offers a high-performing neural network activation function. The GELU activation function is given by $x\Phi(x)$ (Figure \ref{fig:gelu_vs_relu_plot}), where $\Phi(x)$ is the standard Gaussian cumulative distribution function. Unlike ReLUs where inputs are weighed only by their sign, GELU nonlinearity weighs them by their value\cite{agarap2018learning} too.

\begin{figure}[hbt!]
    \centering
    \includegraphics[width=0.6\textwidth]{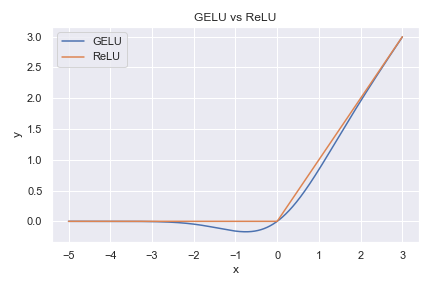}
    \caption{GELU vs ReLU comparison of output values. GELU are increasingly used in popular Transformer architectures \cite{dosovitskiy2021an}, \cite{DBLP:journals/corr/abs-2102-05095}}
    \label{fig:gelu_vs_relu_plot}
\end{figure}

\section{TimeSformer: Video Understanding using ViT}
\label{sec:timesformer_theory}
TimeSformer (Time-Space transformer) \cite{DBLP:journals/corr/abs-2102-05095} adapts Transformer architecture to perform video classification via spatio-temporal feature learning, directly from a sequence of frame level patches. Video understanding is similar to NLP in the sense that both videos and sentences are temporally sequential in nature. Also, similar to word meaning being interpreted by relating it to previous words or context in the sentence, actions in single frames need to be contextualized with previous frames of the video to be correctly classified \cite{DBLP:journals/corr/abs-2102-05095}. Hence, long-range self-attention models from NLP could be expected to work effectively with video modelling as well \cite{DBLP:journals/corr/abs-2102-05095}. \\[0.1in]
In the paper, authors present the possibility of a convolution-free video classification architecture by replacing the 2D or 3D convolution operator with self-attention. Benefits of such a design change are the following:
\begin{itemize}
    \item Transformers lack strong inductive biases inherent to CNNs, such as translation equivariance and local connectivity. Inductive bias is helpful for small training sets, but limits model's expressivity when there is sufficient data available. Hence, Transformers are better able to fit big-data regimes compared to CNNs \cite{DBLP:journals/corr/abs-2102-05095}, \cite{DBLP:journals/corr/abs-2004-13621}, \cite{DBLP:journals/corr/abs-1911-03584}.
    \item While CNN kernels are designed to capture short-range spatiotemporal relations, they cannot precisely model dependencies that extend beyond the effective receptive field. Whereas, self-attention mechanism can be applied to capture both local as well as global long-range dependencies, by directly comparing activations at all space-time locations, way beyond the receptive field of conventional CNNs.
    \item Recent works \cite{dosovitskiy2021an}, \cite{DBLP:journals/corr/abs-2004-13621}, \cite{DBLP:journals/corr/abs-2005-12872} demonstrate that Transformers offer faster training and inference compared to CNNs, making it possible to scale the models to larger learning capacity for comparable computational budgets.
\end{itemize}

Authors adapt Vision Transformer \cite{dosovitskiy2021an} image model to video domain by extending self-attention mechanism from image space to space-time 3D volume. The proposed Time-Space Transformer model TimeSformer \cite{DBLP:journals/corr/abs-2102-05095}, views video as a sequence of patches extracted from temporally correlated individual frames. \\[0.1in]

\textbf{Method} Input \textbf{X} $\in  \mathbf{R}^{H \times W \times 3 \times F}$ consist of F RGB frames of dimension $(H \times W \times 3)$ sampled from the original video. Similar to ViT \cite{dosovitskiy2021an}, each frame is decomposed into \textit{N} non-overlapping patches, each of size $(P \times P)$, such that the \textit{N} patches cover the entire frame, i.e, $N = HW / P^2$. Patches are flattened into vectors $\mathbf{x}_{(p,t)}$ $\in  \mathbf{R}^{3P^2}$ with $p = 1, ..., N$ denoting the spatial locations and $t = 1, ..., F$ denoting indexing over time. \\[0.1in]

\textbf{Patch Embedding} Each patch $\mathbf{x}_{(p,t)}^{0}$ is linearly mapped into an embedding vector $\mathbf{z}_{(p,t)}^0$ via learnable matrix \textbf{E} $\in  \mathbf{R}^{D \times 3P^2}$ according to:

\begin{equation}
    \mathbf{z}_{(p,t)}^{0} = \mathbf{E}\mathbf{x}_{(p,t)}^{0} + \mathbf{e}_{(p,t)}^{pos}
\end{equation}
where $\mathbf{e}_{(p,t)}^{pos}$ $\in  \mathbf{R}^{D}$ represents learnable positional embedding, \textit{D} the latent space dimension. The resulting sequence of embedding vectors $\mathbf{z}_{(p,t)}^{0}$ for $p = 1, ..., N$ and $t = 1, ..., F$ represents the input to Transformer, analogous to a sequence of embedded tokens in NLP. \\[0.1in]

\textbf{Self-attention calculation} Let latent space dimension per head be denoted by $D_h$ such that $D_h = D / A$, where $A$ represents total number of attention heads. At each block \textit{l} of \textit{L} Encoding blocks, query-key-value vectors are computed for each patch using model parameters $\mathbf{W}_{Q}^{(l,a)}$ $\in  \mathbf{R}^{\mathbf{D}_h \times D}$, $\mathbf{W}_{K}^{(l,a)}$ $\in  \mathbf{R}^{\mathbf{D}_h \times D}$, $\mathbf{W}_{V}^{(l,a)}$ $\in  \mathbf{R}^{\mathbf{D}_h \times D}$ as follows:

\begin{equation}
    \mathbf{q}_{(p,t)}^{(l,a)} = \mathbf{W}_{Q}^{(l,a)}\texttt{LN}(\mathbf{z}_{(p,t)}^{(l-1)})
\end{equation}

\begin{equation}
    \mathbf{k}_{(p,t)}^{(l,a)} = \mathbf{W}_{K}^{(l,a)}\texttt{LN}(\mathbf{z}_{(p,t)}^{(l-1)})
\end{equation}

\begin{equation}
    \mathbf{v}_{(p,t)}^{(l,a)} = \mathbf{W}_{V}^{(l,a)}\texttt{LN}(\mathbf{z}_{(p,t)}^{(l-1)})
\end{equation}

where \texttt{LN()} denotes LayerNorm \cite{ba2016layer}, $a = 1, ..., A$ iterates over attention heads and $l = 1, ..., L$ denotes the considered block index. \\[0.1in]

\textbf{Joint Space-Time Attention} \\
In this scheme, authors \cite{DBLP:journals/corr/abs-2102-05095} present a spatio-temporal or 3D attention model where attention is paid both over space and time simultaneously. Self-attention weight $\alpha_{(p,t)}^{(l,a)}$ $\in  \mathbf{R}^{1 \times NF}$ for query patch $(p,t)$ is given by:

\begin{equation}
    \label{eq:spatio_temp}
    \alpha_{(p,t)}^{(l,a)_{\texttt{space}\textunderscore \texttt{time}}} = \texttt{SM} \left( \frac{\mathbf{q}_{(p,t)}^{(l,a)\top}}{\sqrt{D_h}} \left[\{ \mathbf{k}_{(p',t')}^{(l,a)} \}_{\substack{ {p' = 1,...,N} \\ {t' = 1,...,F} } } \right] \right)
\end{equation}

where \texttt{SM} denotes softmax activation function, N number of patches per image, F number of considered frames. In order to reduce computation complexity from $\mathcal{O}(N \times F)$ to $\mathcal{O}(N + F)$, authors \cite{DBLP:journals/corr/abs-2102-05095} propose Divided Space-Time attention model. \\[0.1in]

\textbf{Divided Space-Time Attention} \\
In order to reduce the computational cost for attention calculation in Equation \ref{eq:spatio_temp}, authors replace spatio-temporal attention with the application of temporal attention followed by spatial attention. Computational complexity is significantly reduced when attention is computed only over a single dimension, either temporal or spatial. \\[0.1in]
In temporal attention computation, only F query-key comparisons are required, comparing each patch $(p,t)$ with all patches at the same spatial location $p$ across F frames. Within each block \textit{l}, temporal attention is calculated according to

\begin{equation}
    \label{eq:temporal_attention}
    \alpha_{(p,t)}^{(l,a)_{\texttt{time}}} = \texttt{SM} \left( \frac{\mathbf{q}_{(p,t)}^{(l,a)\top}}{\sqrt{D_h}} \left[\{ \mathbf{k}_{(p,t')}^{(l,a)} \}_{t' = 1,...,F} \right] \right)
\end{equation}

Similarly, for spatial attention, only N query-key comparisons are made, using keys from the same frame as the query. In spatial attention, each patch $(p,t)$ is attended among all patches within the same frame corresponding to time $t$. Self-attention weight $\alpha_{(p,t)}^{(l,a)}$$ \in \mathbf{R}^{1 \times N}$ is given by:

\begin{equation}
    \label{eq:spatial_attention}
    \alpha_{(p,t)}^{(l,a)_{\texttt{space}}} = \texttt{SM} \left( \frac{\mathbf{q}_{(p,t)}^{(l,a)\top}}{\sqrt{D_h}} \left[\{ \mathbf{k}_{(p',t)}^{(l,a)} \}_{p' = 1,...,N} \right] \right)
\end{equation}

Resulting vector is fed into MLP to generate the final encoding of patch $(p,t)$. \\[0.1in]

Compared to $(N \times F)$ comparisons per patch in Joint Space-Time attention model, the new model performs only $(N + F)$ comparisons, thereby significantly speeding up model training. Divided Space Time  model has distinct query-key-value matrices denoted by $\{ W_{Q^{\texttt{time}}}^{(l,a)}, W_{K^{\texttt{time}}}^{(l,a)}, W_{V^{\texttt{time}}}^{(l,a)} \}$ and $\{ W_{Q^{\texttt{space}}}^{(l,a)}, W_{K^{\texttt{space}}}^{(l,a)}, W_{V^{\texttt{space}}}^{(l,a)} \}$, over temporal and spatial space.

\section{Saliency mapping techniques}
\label{sec:saliency_map_info}
Saliency maps are frequently used to support explanations of the behaviour of deep reinforcement learning agents \cite{Atrey2020Exploratory}. We use perturbation based saliency map techniques proposed in \cite{DBLP:journals/corr/abs-1711-00138} to visualize which regions of image influence the agent behaviour. Also, we look out for overlapping between attention map and saliency map projections to check if they both point to similar image artifacts.\\[0.1in]

\textbf{Perturbation based Saliency methods}\\
The idea behind the perturbation-based saliency method \cite{DBLP:journals/corr/abs-1711-00138} is to measure how a model's output changes when the model's input(image) state is altered. The authors do saliency map analysis for both actor (policy $\pi$) and critic (value estimate $V^{\pi}$) at every time step. Saliency map for policy $\pi(I_{t})$ is intended to identify key information in frame $I_t$ that the policy uses to select action $a_t$. Similarly, saliency map for value function ${V^\pi}(I_{t})$ is intended to identify the key information in frame $I_t$ for assigning value at time t. \\[0.1in]

\textbf{Mathematical formulation} \\
Given an image $I_t$ $\in  \mathbf{R}^{H \times W}$ at time t, perturbed image $\Phi(I_t, i, j)$ with perturbation centered at pixel coordinates $(i, j)$ is given by:

\begin{equation}
    \label{eq:saliency_map_equation}
    \Phi(I_t, i, j) = I_t\circ (1-M(i,j)) + A(I_t, {\sigma}_{A})\circ M(i,j)
\end{equation}

where $\circ$ denotes Hadamard product. The blur is generated by interpolating between the original image $I_t$ and Gaussian blur $A(I_t, {\sigma}_{A}=3)$ of the same image, using interpolation mask $M(i,j)$ $\in  (0,1)^{H\times W}$ (Figure \ref{fig:saliency_map_example}). The mask $M(i,j)$ corresponds to a 2D Gaussian centered at $\mu = (i,j)$ with $\sigma=5$, and $(H,W)$ represents image dimension.

\begin{figure}[hbt!]
    \centering
    \includegraphics[width=1.0\textwidth]{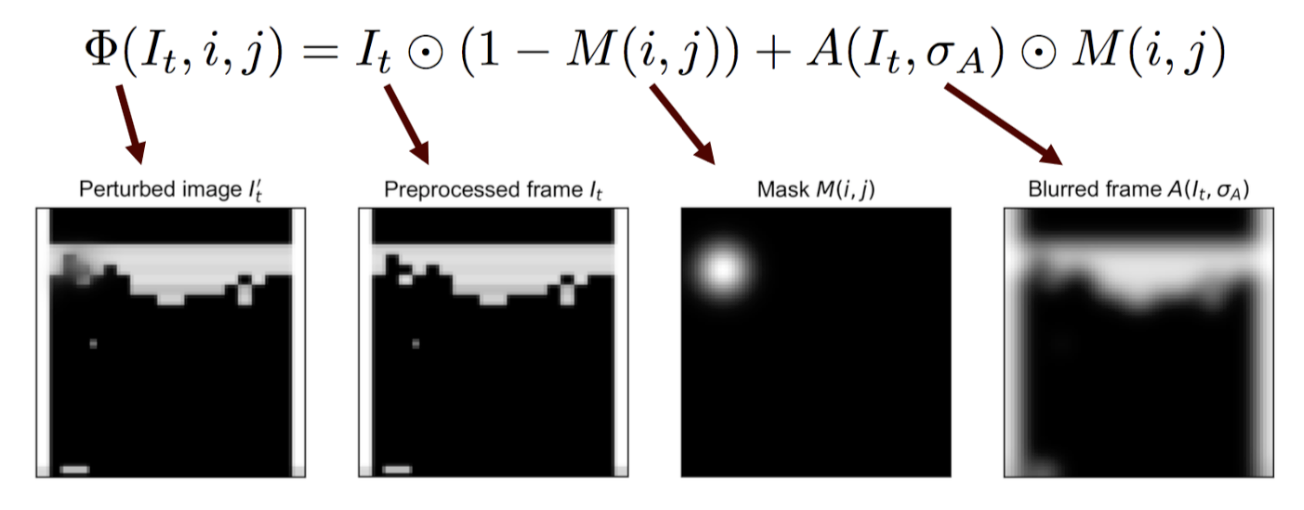}
    \caption{An example of how perturbation method selectively blurs a region applied to Atari Breakout environment. (Image source: \cite{DBLP:journals/corr/abs-1711-00138})}
    \label{fig:saliency_map_example}
\end{figure}

The question authors try to answer here is ``How much does removing information from region around $(i,j)$ impact the policy or value function?". Let $\tilde{\pi(I_{t})}$ denotes the perturbed and $\pi(I_{t})$ the unperturbed policy logits. Saliency metric for policy function at image location $(i,j)$ at time t is given by:

\begin{equation}
    \label{eq:saliency_policy_metric}
    \mathcal{S}_{\pi}(t, i, j) = \frac{1}{2}\lVert \pi(I_{t}) - \tilde{\pi(I_{t})} \rVert^2
\end{equation}

Similarly, value function saliency metric is given by:

\begin{equation}
    \label{eq:saliency_policy_metric}
    \mathcal{S}_{V^{\pi}}(t, i, j) = \frac{1}{2}\lVert V^{\pi}(I_{t}) - \tilde{V^{\pi}(I_{t})} \rVert^2
\end{equation}

\textbf{Saliency map implementation} \\
With these definitions, one can construct a saliency map for either policy $\pi$ or value function $V_{\pi}$ by computing $\mathcal{S}_{\pi}(t, i, j)$ for every pixel in image $I_t$. To lower the computational cost, we sample one in every five pixels and resize the generated saliency map to the original frame size with bilinear interpolation using open-cv library \cite{opencv_library}. We convert the map into heat map using open-cv library \cite{opencv_library} and alpha-blend with the corresponding image.

    
    \chapter{Methods}
    
In the following sections, we present various temporal architectures we have experimented with. Most of them make use of attention mechanism to model long-term time dependencies. We are using attention-based models in an attempt to monitor the information used by an agent to act and thereby make our models more interpretable than the traditional models. \\[0.1in]

\section{Mott's model}
\label{sec:motts_architecture_section}
In \cite{DBLP:journals/corr/abs-1906-02500}, the authors present a soft spatial attention mechanism using image data in a reinforcement learning setting. The model (Figure \ref{fig:motts_architecture}, \ref{fig:motts_process_flow}) extracts task-relevant information from image inputs by sequentially querying the current view of the environment and generating appropriate outputs from the output of attention mechanism. We used an open source implementation of the paper by Lovering et al. \cite{lovering2020reproducing} and improved on it. Model consists of the following functional blocks: \\[0.1in]

\begin{figure}[hbt!]
    \centering
    \includegraphics[width=0.6\textwidth]{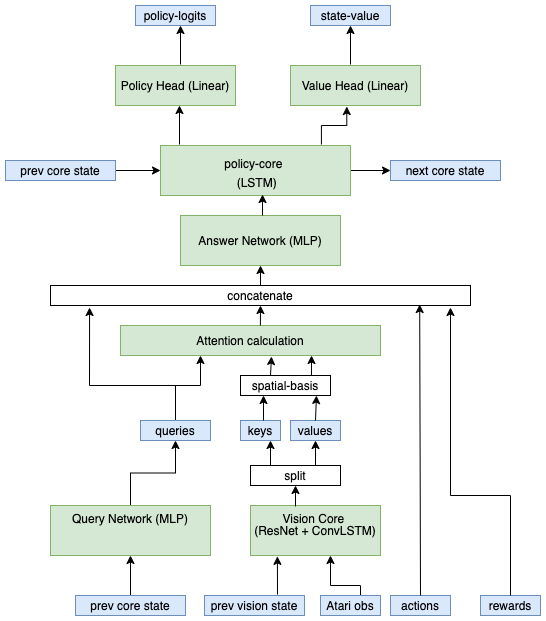}
    \caption{Attention networks architecture proposed in Mott et al. \cite{DBLP:journals/corr/abs-1906-02500}}
    \label{fig:motts_architecture}
\end{figure}

\textbf{Vision core} \\
Vision core process input images from environment. It consist of a convolutional neural network (CNN) followed by a ConvLSTM \cite{DBLP:journals/corr/ShiCWYWW15}. We had two configurations of CNN: one using a plain multilayer convolutional network as given in the original work \cite{DBLP:journals/corr/abs-1906-02500} and other using a more advanced ResNet \cite{DBLP:journals/corr/HeZRS15}. At time t, observation \textbf{X} $\in \mathbf{R}^{H\times W\times C}$ from Atari-environment is fed into the vision core to generate output as follows:
\begin{equation}
    \mathbf{O}_{\texttt{vis}}, s_{\texttt{vis}}(t) = \texttt{vis}_{\theta}(\mathbf{X}_t, s_{\texttt{vis}}(t-1))
\end{equation}
where $\textbf{O}_{\texttt{vis}}$, $\textbf{s}_{\texttt{vis}}(t)$, $\textbf{s}_{\texttt{vis}}(t-1)$ respectively denotes output, current hidden state, previous hidden state tensors of ConvLSTM \cite{DBLP:journals/corr/ShiCWYWW15} layer. Vision core output $\textbf{O}_{\texttt{vis}}$ $\in \mathbf{R}^{h \times w \times c}$ is split along channel dimension into two tensors: Keys: \textbf{K} $\in \mathbf{R}^{h \times w\times c_{K}}$  and Values: \textbf{V} $\in \mathbf{R}^{h \times w\times c_{V}}$, such that $c_K + c_V = c$. \\[0.1in]

\textbf{Spatial basis module} \\
In order to inject spatial information into attention vectors, a static non-trainable tensor: Spatial basis \textbf{S} $\in \mathbb{R} ^ {h \times w \times c_S}$ (Figure \ref{fig:spatial_encodings_motts}) is concatenated to \textbf{K} and \textbf{V} along channel dimension such that the new dimensions are, respectively
 $(h \times w \times {(c_K + c_S)})$ and $(h \times w \times {(c_V + c_S)})$. Each channel of \textbf{S} corresponding to spatial frequencies $(u, v)$, is an outer product of two Fourier basis vectors \cite{DBLP:journals/corr/abs-1906-02500}. For example, one channel of \textbf{S} with spatial frequencies \textit{u} and \textit{v} for two even Fourier basis functions would be:
\begin{equation}
    \mathbf{S}_{i,j,(u, v)} = \texttt{cos}(\pi u*i/h)\texttt{cos}(\pi v*j/w)
\end{equation}

where \textit{i, j} are spatial locations in the tensor. Authors \cite{DBLP:journals/corr/abs-1906-02500} generate all possible outer products such that number of channels in \textbf{S} is $(U+V)^2$. Here, $U$ and $V$ respectively denoting the number of spatial frequencies used for even and odd components, are both set to 4.  

\begin{figure}[hbt!]
    \includegraphics[width=1.0\textwidth]{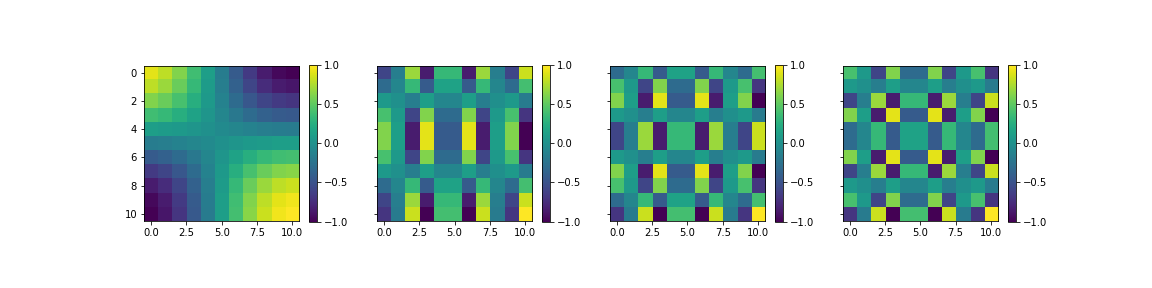}
    \caption{Visualization of four symmetrical spatial encodings with $u=v$ out of the total 64 channels. Left to right, spatial frequency $(u, v)$ increase from $(1, 1)$ to $(4, 4)$.}
    \label{fig:spatial_encodings_motts}
\end{figure}

\textbf{Query network} \\
Query network $Q_\psi$ parametrised by $\psi$, consist of a multi-layer perceptron (MLP) whose output is reshaped into H query vectors $q^i$, with $i = 1, 2,..., H$, $q^i$ $\in \mathbf{R} ^ {1 \times (c_S + c_K)}$ and H the number of attention heads. Query vectors are generated by feeding  the previous timestep output $s_{\texttt{LSTM}}(t-1)$ back into the MLP. The recurrent nature between previous output and the current query encodes temporal dependencies in the model. \\[0.1in]
\begin{equation}
    \label{eq:motts_attention_calculation_equation}
    q^1...q^H = Q_{\psi}(s_{\texttt{LSTM}}(t-1))
\end{equation}

Taking inner product between each query vector $q^i \in \mathbf{R} ^ {1 \times (c_S + c_K)}$ and keys tensor \textbf{K}$ \in \mathbf{R}^{h \times w \times (c_S + c_K)}$ over channel dimension, n-th attention logit map $\tilde{\textbf{A}}\textsuperscript{n}$ $\in  \mathbf{R}^{h \times w}$ is computed as
\begin{equation}
    \tilde{\mathbf{A}}_{i, j}^n = \sum \nolimits_{l} q_{l}^n\times \mathbf{K_{\mathit{i}, \mathit{j}, \mathit{l}}}
\end{equation}
where $l=1, ..., (c_{S} + c_{K})$. Softmax is performed spatially over spatial indexes $(i, j)$ to produce normalized attention map $\textbf{A} \textsuperscript{n}$ $\in \mathbf{R}^{h \times w}$. \\[0.1in]
\begin{equation}
    \label{eq:motts_spatial_attention_expression}
    \mathbf{A}_{i, j}^n = \frac{\exp(\tilde{\mathbf{A}}_{i, j}^n)}{\sum \nolimits_{i', j'} \exp(\tilde{\mathbf{A}}_{i', j'}^n)}
\end{equation}

Each attention map $\textbf{A}^{n}$ $\in \mathbf{R}^{h \times w}$ is broadcasted along channel dimension, point wise multiplied with \textbf{V} $\in \mathbf{R}^{h \times w \times (c_S + c_V)}$ and summed across space  to produce the n-th answer vector $\textit{a}^n$ $\in \mathbf{R}^{1 \times 1 \times (c_S + c_V)}$,

\begin{equation}
    \label{eq:motts_spatial_sum}
    a_{c}^n = \sum \nolimits_{i, j} \mathbf{A}_{i, j}^n \times \mathbf{V}_{i,j,c}
\end{equation}
where $n=1, ..., H$ denotes the attention-head index, $(i, j)$ varies over the spatial indexes and $c=1, ..., (c_S + c_V)$ varies over the channel dimension. The spatial sum (Equation \ref{eq:motts_spatial_sum}) acts as a severe bottleneck, preserving the most relevant information for decision making in answer vector in $a^n$ $\in \mathbf{R}^{1 \times 1 \times (c_S + c_V)}$. \\[0.1in]

\begin{figure}[hbt!]
    \centering
    \includegraphics[width=0.7\textwidth]{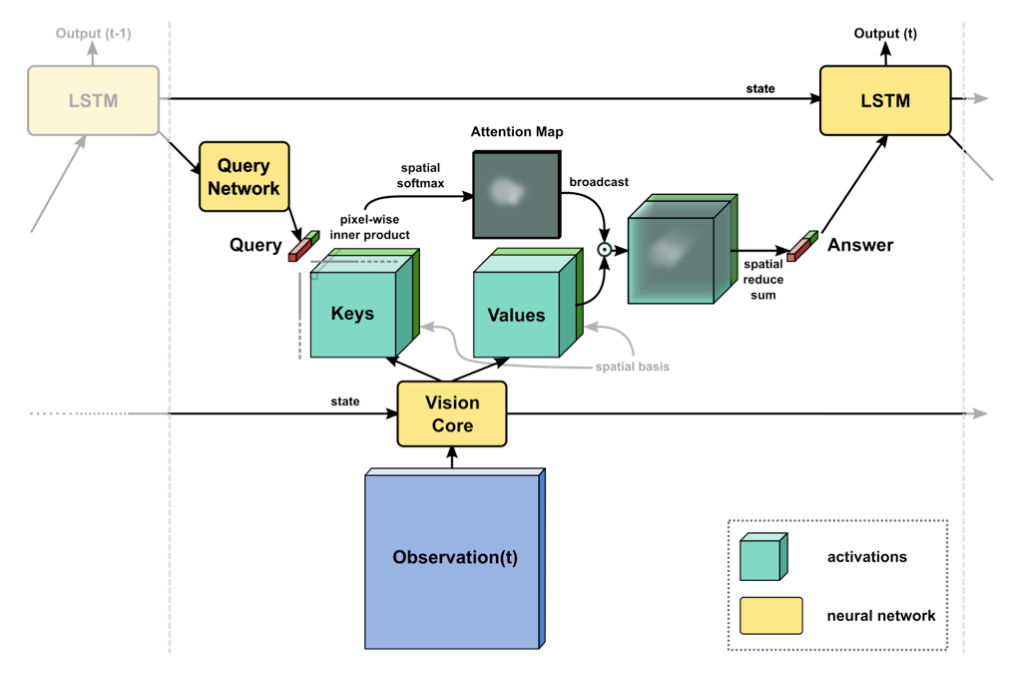}
    \caption{ Outline of model proposed in Mott et al. \cite{DBLP:journals/corr/abs-1906-02500}. Observations are passed through a recurrent Vision core which generates \textbf{keys} and \textbf{values} tensors. Inner product is calculated between each query vector and each location in the keys tensor, followed by spatial softmax to generate attention map per query. The attention map is broadcast along the channel dimension, point-wise multiplied with values tensor and result summed across space to produce an answer vector. Answer vector is sent to top core-LSTM to generate output and next state of the LSTM. (Image source: \cite{DBLP:journals/corr/abs-1906-02500})}
    \label{fig:motts_process_flow}
\end{figure}

\textbf{Answer processor and Policy core} \\
The H answer vectors $\textit{a}^n$, H query vectors $\textit{q}^n$, previous reward $r_{t-1}$ and previous policy-logit $\pi_{t-1}$ are concatenated and fed into answer processor $\texttt{MLP}_{\theta}$: a two-layer MLP. The MLP owns a major fraction of the total model's parameter volume: 0.66M out of the total 2.18M parameters, and influences the training time to an extent. Output of the answer processor is fed into the policy core which is a single layer, fully connected LSTM cell \cite{HochSchm97}.
\begin{equation}
    \texttt{input}_{\texttt{core}}(t) = \texttt{MLP}_{\theta}([a^1, ..., a^n], [q^1, ..., q^n], r_{t-1}, \pi_{t-1})
\end{equation}
\begin{equation}
    o(t), s_{\texttt{LSTM}}(t) = \texttt{LSTM}_{\phi}(\texttt{input}_{\texttt{core}}(t), s_{\texttt{LSTM}}(t-1))
\end{equation}
\\[0.1in]

\textbf{Policy and Value head} \\
The output of policy core LSTM $o(t)$ is processed through a one-layer MLP and splitted into policy logit and value estimate. \\[0.1in]

\textbf{ResNet architecture used} \\
The ResNet architecture we tried with Mott \cite{DBLP:journals/corr/abs-1906-02500} model is the large ResNet architecture (Figure \ref{fig:resnet_architecture}) from IMPALA \cite{DBLP:journals/corr/abs-1802-01561}. It consist of 15 convolutional layers with residual connections and comes to around 97.7k parameters. \\[0.1in]

\begin{figure}[hbt!]
    \centering
    \includegraphics[width=0.6\textwidth]{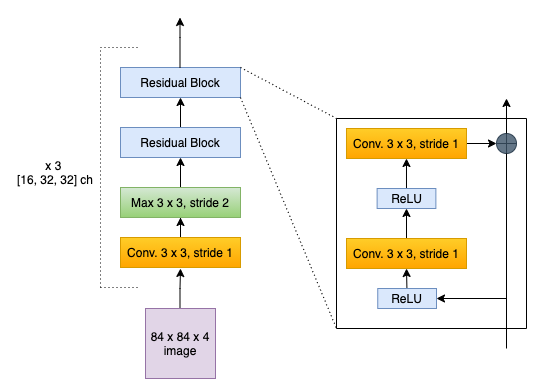}
    \caption{ResNet architecture from Impala \cite{DBLP:journals/corr/abs-1802-01561} used in most of our implementations.}
    \label{fig:resnet_architecture}
\end{figure}


\section{Adaptive architecture: Using Transformers for temporal attention}
\label{sec:adaptive_transformer_xl}
RL tasks share similarity with language modelling in terms of high correlation between consecutive states. Hence, it would be natural to assume that Transformers (Vaswani et al. \cite{NIPS2017_3f5ee243}) which are the state of art in Natural Language Processing, would also work in a partially observable RL environment. There have been many previous works in this line: Parisotto et al. \cite{DBLP:journals/corr/abs-1910-06764} and Kumar et al. \cite{DBLP:journals/corr/abs-2004-03761} are a few to mention. In this section, we are using Transformer-XL \cite{DBLP:journals/corr/abs-1901-02860} variant of Transformer family to impart memory to RL agents. We used previous works by Parisotto et al. \cite{DBLP:journals/corr/abs-1910-06764} and Kumar et al. \cite{DBLP:journals/corr/abs-2004-03761} as our starting point. We use a similar naming convention as Kumar et al. \cite{DBLP:journals/corr/abs-2004-03761} and will be referring to the model as Adaptive architecture in later sections\\[0.1in]

\textbf{Model Architecture} \\
Similar to \cite{DBLP:journals/corr/abs-1910-06764} and \cite{DBLP:journals/corr/abs-2004-03761}, we use a encoder-only configuration of Transformer-XL in our architecture. We are using the Encoder architecture proposed in \cite{DBLP:journals/corr/abs-1910-06764}. The authors \cite{DBLP:journals/corr/abs-1910-06764} claim substantial improvement in stability and learning speed with reordering of layer normalization \cite{ba2016layer}. The layer normalization layers are reordered from output side of Multi-Head Attention unit and position-wise multilayer perceptron (MLP) to their respective input sides. Combined with residual connections, this results in gradients flowing from input to output without any transformation. \\[0.1in]

\begin{figure}[hbt!]
    \centering
    \includegraphics[width=0.3\linewidth]{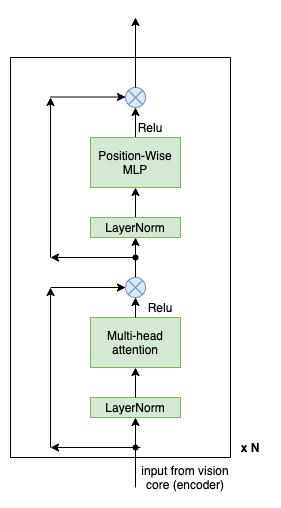}
    \caption{Transformer Encoder used in our model.}
    \label{fig:adaptive_transformer_decoder}
\end{figure}

The Adaptive architecture (Figure \ref{fig:adaptive_architecture}) is simpler in design compared to Mott architecture (Figure: \ref{fig:motts_architecture}). The first major difference is that the LSTM policy core in Mott architecture (Figure \ref{fig:motts_architecture}) is replaced with Transformer-XL \cite{DBLP:journals/corr/abs-1901-02860} policy-core. Another major difference is the Adaptive architecture's Vision core consist of just the three-layer ResNet (Figure \ref{fig:resnet_architecture}) without ConvLSTM \cite{DBLP:journals/corr/ShiCWYWW15}. ConvLSTM was removed since the Adaptive architecture was learning faster without it. Finally, the spatial attention calculation performed in Mott et al. \cite{DBLP:journals/corr/abs-1906-02500} is no longer applicable since encoded data no longer has the spatial structure. \\[0.1in]

Output from Vision core is flattened and fed into a single layer MLP to shrink the encoded image's latent space dimension. Spatial structure associated with image inputs are lost after flattening. This rules out the scope of spatial segmentation via projecting attention maps back onto the images. Output from single layer MLP is concatenated with previous-step reward and previous-step policy logits before feeding into Transformer-XL policy core. At each step, Transformer's output from previous time-step is cached and fed into model along with other inputs as described in Figure: \ref{fig:transformer_xl_processing}. The output of the transformer is linearly mapped to policy and value heads to generate policy logits and state values, respectively. Entire Adaptive architecture is summarised in Figure \ref{fig:adaptive_architecture}. \\[0.1in]

\begin{figure}[hbt!]
    \centering
    \includegraphics[width=0.7\linewidth]{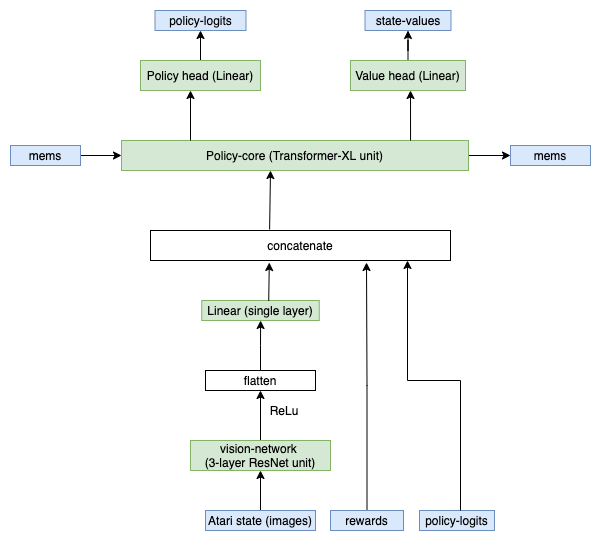}
    \caption{Combined block diagram of Adaptive architecture.}
    \label{fig:adaptive_architecture}
\end{figure}

\textbf{Data processing} \\
Embedding size is kept constant within the encoder unit to allow for the residual skip connections. Batch of input embedding is summed with position encoding from Figure \ref{fig:encodings_transformer_xl} to impart a temporal structure to the data. Forward masking of batch input is done to prevent using future tokens for current token's attention calculation. Unlike LSTM policy core, Transformer based policy core can replace the recurrence between inputs and outputs with self-attention. Consequently, Transformer policy core can process data in one shot and hence allows easy parallelization during training. \\[0.1in]
Actors perform experience rollout of length \texttt{unroll\textunderscore length}(240 time steps for Atari env) and push these trajectories into memory buffers (Figure: \ref{fig:adaptive_pipeline}) as explained in Monobeast algorithm [\ref{alg:monobeast_data_processing}]. The trajectories are split into smaller chunks (80 time steps for Atari env) and concatenated with Transformer's memory of size \texttt{mem\_len}. After processing each chunk, Transformer caches \texttt{mem\_len} tokens of the current chunk into a buffer \texttt{mems}. While processing the next chunk, previous chunk cached in \texttt{mems} buffer acts as the extended context or memory of Transformer (Section \ref{sec:transformer_xl_background}, Figure \ref{fig:transformer_xl_processing}). The central learner calculates V-trace returns and backpropagates on the loss function as explained in Section \ref{section:monobeast_loss_function}. \\[0.1in]

\begin{figure}[hbt!]
    \centering
    \includegraphics[width=0.7\textwidth]{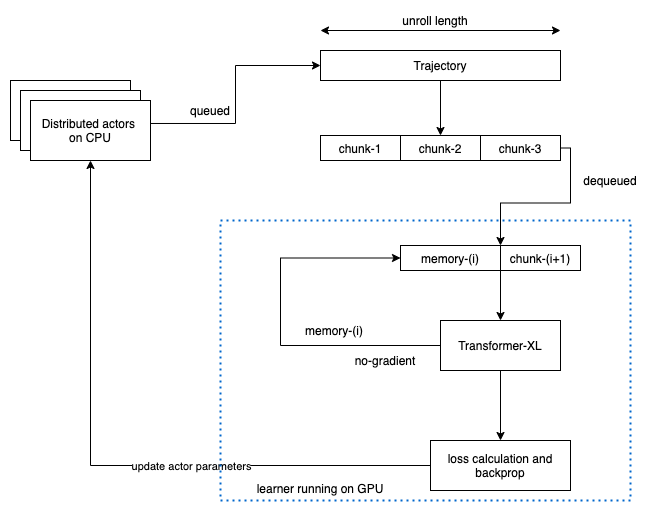}
    \caption{IMPALA data processing pipeline used in Kumar et al. \cite{DBLP:journals/corr/abs-2004-03761}}
    \label{fig:adaptive_pipeline}
\end{figure}


\section{Spatio-temporal attention: Combining Mott and Adaptive architecture} 
\label{sec:combine_motts_transformer_xl}
In this architecture, we combine spatial attention from Mott's architecture (Section: \ref{sec:motts_architecture_section}) and temporal attention from Adaptive architecture (Section: \ref{sec:adaptive_transformer_xl}) to generate 3D-attention or spatio-temporal attention. Architecturally, model (Figure \ref{fig:spatio_temporal_architectures} Left) is very similar to Mott's architecture(Figure: \ref{fig:motts_architecture}). Compared to the previous Adaptive architecture (Figure \ref{fig:adaptive_architecture}), Spatio-Temporal architecture incorporated Mott style spatial attention calculation(Section \ref{sec:motts_architecture_section}) in addition to Transformer's temporal attention, and hence, opened the possibilities of spatial segmentation. \\[0.1in]

\textbf{Sequential data processing} \\
Similar to Mott model (Figure \ref{fig:motts_architecture}), in Spatio-Temporal model too, queries for spatial-attention calculation (Equation \ref{eq:motts_spatial_attention_expression}) are generated sequentially from previous-step policy core output. Hence, input data from the buffer could only be processed sequentially and acts as a severe bottleneck for training speed. This was not the case with Adaptive architecture (Figure \ref{fig:adaptive_architecture}) since it did not involve any spatial attention calculation. \\[0.1in]

Spatial attention is applied first as explained in Section \ref{sec:motts_architecture_section}, followed by temporal attention in Adaptive architecture core. Spatially attended encodings are concatenated with previous action-logit and reward before feeding into Transformer. In Adaptive architecture core, temporal attention calculation is done in one shot(Section \ref{sec:transformer_self_attention_calculation}). \\[0.1in]

\textbf{Sub Architecture to overcome sequential spatial attention bottleneck} \\
We propose a minor workaround to speed up training in the spatio-temporal architecture. For spatial attention, instead of using Transformer-core output to generate query every time-step, we cache Actor core's output at the corresponding time step into a buffer \texttt{actor\textunderscore agent\textunderscore state}(Figure \ref{fig:spatio_temporal_architectures} Right). During training time, model uses \texttt{agent\textunderscore state} from  \texttt{actor\textunderscore agent\textunderscore state} to generate queries for spatial attention computation. Generated queries are delayed compared to sequential queries generated by learner-core since actor lags behind learner model by several updates. We assume that off-policy loss calculation using V-trace is able to compensate the error. In spite of increased memory complexity and delayed queries, we will later see that the agent learns and shows comparable performance to the one trained with the former sequential architecture. \\[0.1in]
The new architecture also increases training speed by approximately 6 times since it allows for parallelization in spatial attention calculation. \\[0.1in]

\begin{figure}[hbt!]
\centering
\begin{subfigure}{0.5\textwidth}
    \centering
    \includegraphics[width=1.0\textwidth]{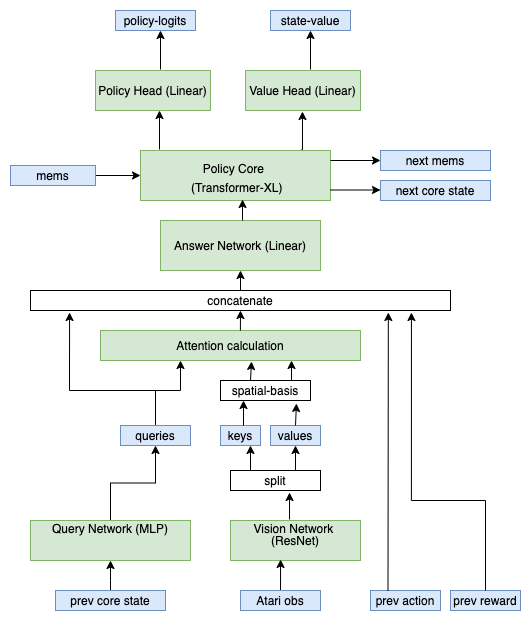}
    \label{fig:sp_temp_seq_architecture}
\end{subfigure}%
\begin{subfigure}{0.5\textwidth}
    \centering
    \includegraphics[width=1.0\textwidth]{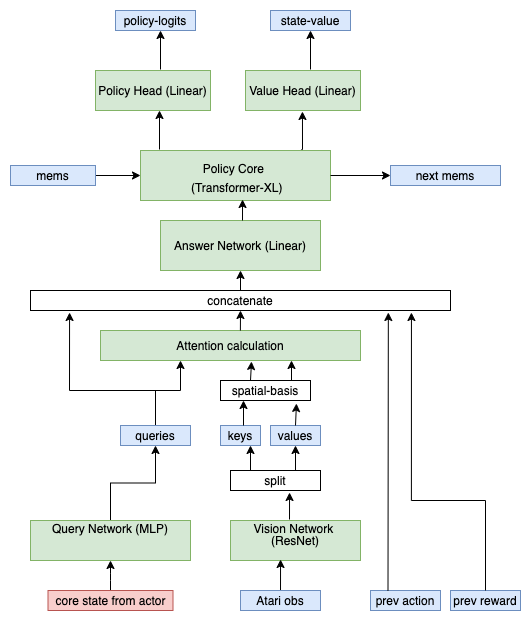}
    \label{fig:sp_temp_batched_architecture}
\end{subfigure}%

\caption{Left: Spatio-temporal sequential attention architecture. Right: With modified spatio-temporal one-shot architecture, parallel processing is again possible for spatial attention calculation. The key difference between both models is in how \texttt{prev\textunderscore core\textunderscore state} is sourced to Query Network to generate spatial-attention queries.}
\label{fig:spatio_temporal_architectures}
\end{figure}


\section{Spatio-temporal attention using Vision Transformer}
\label{sec:timesformer_method}
In the final architecture, we come up with a design using Vision Transformer \cite{dosovitskiy2021an} (ViT), in RL environment. Our model is closely similar to video classification model: TimeSformer (Section: \ref{sec:timesformer_theory}) by Gedas et al. \cite{DBLP:journals/corr/abs-2102-05095}. In video classification, atomic actions in short-term segments need to be contextualized with the rest of the video to be fully disambiguated \cite{DBLP:journals/corr/abs-2102-05095}. Video classification tasks might share spatio-temporal similarity with RL tasks on partially observable RL environments, in the sense that agent's current actions can be better contextualized using previous states. \\[0.1in]

\begin{figure}[hbt!]
    \includegraphics[width=0.8\textwidth]{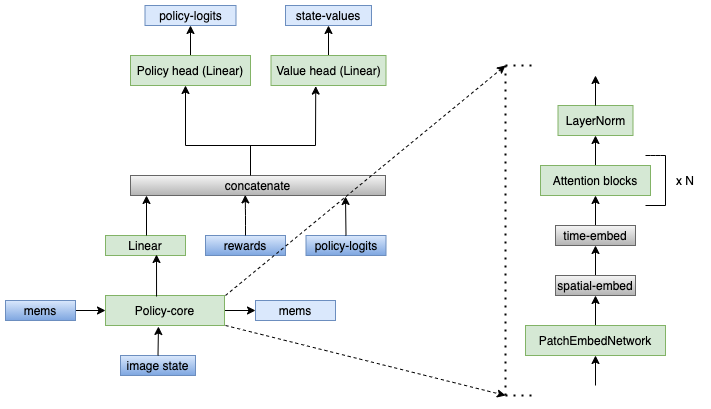}
    \caption{Our architecture using ViT is closely similar to adaptive architecture(Figure: \ref{fig:adaptive_architecture})}
    \label{fig:vit_block_diagram}
\end{figure}

\textbf{Model architecture} \\
Unlike former architectures, TimeSformer model(Figure \ref{fig:vit_block_diagram} Left) is relatively simpler in design with no Vision-core and Query network. Image batch is fed into Policy-core which performs spatial and temporal self-attention. Policy core(Figure \ref{fig:vit_block_diagram} Right) consist of Patch embedding network followed by attention blocks. Sequences are embedded spatially and temporally before feeding into attention blocks in Policy core. Similar to Adaptive architecture (Section \ref{sec:adaptive_transformer_xl}), output of Policy core is passed through a single layer Linear network to shrink the encoded image's latent space dimension. Linearly mapped output of Policy core is concatenated along with previous rewards, and policy logits before feeding into separate policy and value heads. As we will see later, the Linear layer owns a major chunk of the TimeSformer model parameters(0.482M out of the model's total 0.558M parameters). Compared to previous architectures, this is a key difference: previous models concatenated encoded image, previous rewards, and policy-logits before passing them to policy core, whereas the ViT based architecture concatenate the inputs after the policy core. \\[0.1in]
We experimented with both TimeSformer (Section \ref{sec:timesformer_theory}) models: Joint Space-Time model(Figure \ref{fig:vit_attention_scheme} Left) and Divided Space-Time (Figure \ref{fig:vit_attention_scheme} Right). \\[0.1in]

In Divided Space-Time scheme, frames are first attended temporally followed by spatial attention according to Equation \ref{eq:temporal_attention} and Equation \ref{eq:spatial_attention}. On the other hand, in Joint Space-Time scheme, attention is applied simultaneously over space-time(Equation \ref{eq:spatio_temp}). \\[0.1in]

\textbf{Major differences compared to TimeSformer model} \\
As mentioned before, we build upon TimeSformer architecture \cite{DBLP:journals/corr/abs-2102-05095} and introduced architecture modifications to improve sample efficiency of the model. Primary change was, to allow for autoregressive data processing. To provide contextual memory to ViT, we performed `Transformer-XL' style mem-caching (Figure \ref{fig:transformer_xl_processing}) as explained in Section \ref{sec:adaptive_transformer_xl}. Moreover, during temporal and spatio-temporal attention, we use forward attention masking (Section \ref{sec:adaptive_transformer_xl}) as proposed in Transformer-XL \cite{DBLP:journals/corr/abs-1901-02860} to mask out future tokens during attention calculations for current tokens. Finally, since ours is not a classification task and since we need a one-to-one mapping between frames and actions, we are not using BERT \cite{DBLP:journals/corr/abs-1810-04805} like classification token in our design, but draw output from every frame. \\[0.1in]

\begin{figure}[hbt!]
\centering
\begin{subfigure}{0.2\textwidth}
    \centering
    \includegraphics[width=1.0\linewidth]{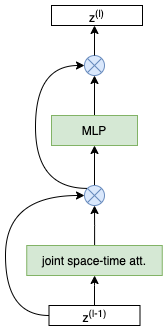}
\end{subfigure}%
\hspace{0.3\textwidth}
\begin{subfigure}{0.2\textwidth}
    \centering
    \includegraphics[width=1.0\linewidth]{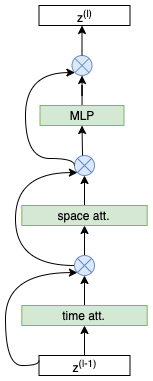}
\end{subfigure}%
\caption{Self-attention blocks from TimeSformer \cite{DBLP:journals/corr/abs-2102-05095} used in our implementation. Policy core in Figure \ref{fig:vit_block_diagram} constitute of $N$ number of these self-attention blocks connected serially. Left: Joint Space-Time attention block. Right: Divided Space-Time attention block.}
\label{fig:vit_attention_scheme}
\end{figure}

\textbf{Positional encoding} \\
For both spatial and temporal positional encoding, we tried initializing using advanced sinusoidal techniques used in Mott's architecture (Figure \ref{fig:spatial_encodings_motts}) and Adaptive architecture (Figure \ref{fig:encodings_transformer_xl}). Similar to Transformer-XL encodings (Figure \ref{fig:encodings_transformer_xl}), they were also made part of model's learnable parameters. We noticed that there was no additional performance improvement with sinusoidal initialization as compared to Gaussian distribution initialization in TimeSformer \cite{DBLP:journals/corr/abs-2102-05095}. Hence, for all future experiments, we just used trainable positional encodings initialized from a Gaussian distribution, for both spatial and temporal encoding. 

    
    \chapter{Results}
    In this chapter, for the first half, we present environment wise results of our models. In the second half, we present an analysis section wherein we justify our network design and selection of hyperparameters, with empirical results. Experiments were run on our cluster using five GeForce RTX 2080 GPU units.

\section{Environment wise results}
In this section, we compare the performance of our different architectures on OpenAI Gym \cite{brockman2016openai} Atari-2600 game suite. We mostly experimented with the following four environments: Pong, Enduro, Pacman and Breakout. Variation of both episode-returns and episode-lengths, with training steps are plotted. Episode return and length are calculated as moving average over previous 100 training episodes. To smooth noisy episode-returns and episode-lengths plot, we have used 1-D Savgol filter \cite{savgol_filter}. Same seed value was used for all environments to generate the performance plots. For all four environments, we experimented the following six architectures detailed in the previous sections:
\begin{itemize}
    \item Mott architecture (Section \ref{sec:motts_architecture_section}).
    \item Adaptive architecture (Section \ref{sec:adaptive_transformer_xl}).
    \item Spatio-Temporal sequential architecture (Figure \ref{fig:spatio_temporal_architectures} Left).
    \item Spatio-Temporal one-shot architecture using actor queries (Figure \ref{fig:spatio_temporal_architectures} Right).
    \item TimeSformer: Divided Space-Time architecture (Figure \ref{fig:vit_attention_scheme} Right).
    \item TimeSformer: Joint Space-Time architecture (Figure \ref{fig:vit_joint_3d_attention}).
\end{itemize}

\subsection{Results on Pong environment}
We ran all architectures on relatively easier Pong environment. Except for the Mott architecture (Section \ref{sec:motts_architecture_section}), all models learned to complete Pong game within 10M training steps (Figure \ref{fig:pong_result}) with Adaptive architecture(Section \ref{sec:adaptive_transformer_xl}) learning it the fastest. Faster learning with the Pong environment was the main reason it was selected for most of the analysis and debugging.

\begin{figure}[hbt!]
\centering
\begin{subfigure}{0.5\textwidth}
    \centering
    \includegraphics[width=1.0\linewidth]{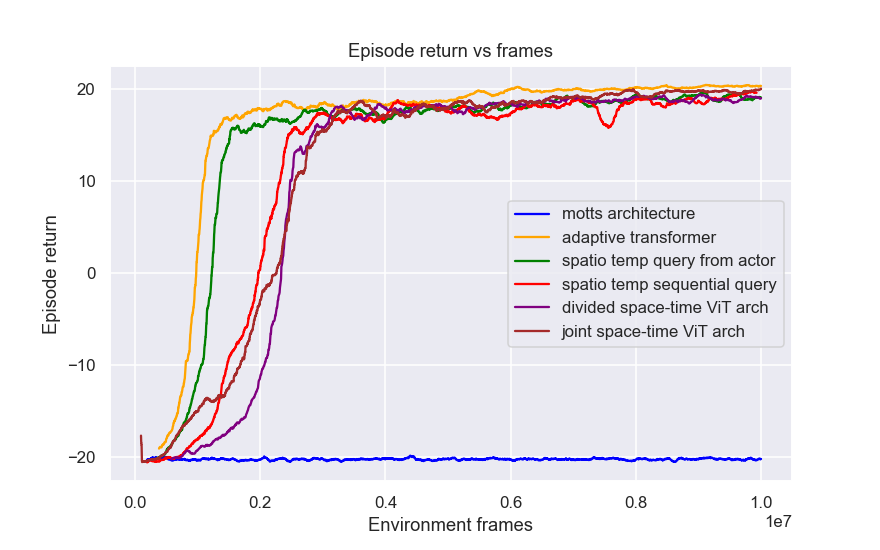}
\end{subfigure}%
\begin{subfigure}{0.5\textwidth}
    \centering
    \includegraphics[width=1.0\linewidth]{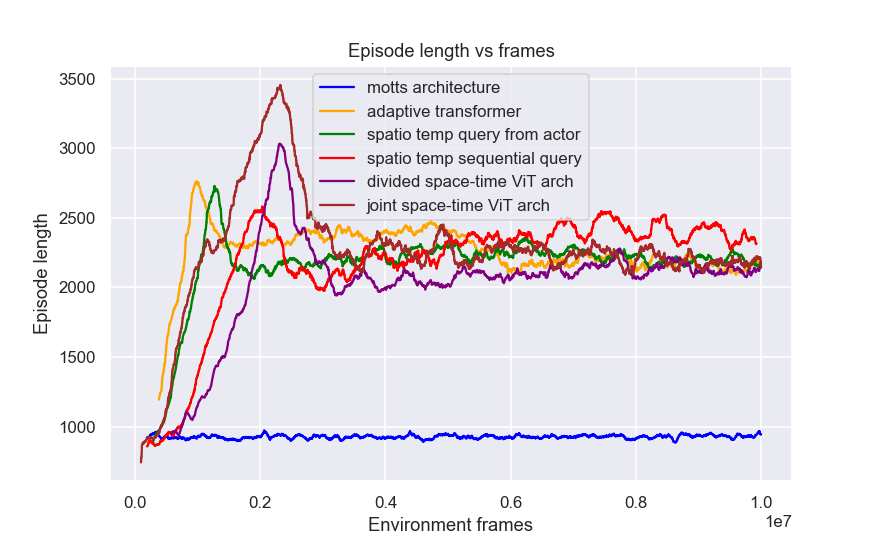}
\end{subfigure}%
\caption{Performance plots for Pong environment trained for 10M steps.}
\label{fig:pong_result}
\end{figure}

\begin{table}[hbt!]
    \caption{Comparison of performance metrics of all models on Pong environment trained for 10M steps.}
    \centering
        \begin{tabular}{p{3.5cm}|p{1.3cm}|p{1.5cm}|p{1.3cm}|p{1.3cm}|p{1.3cm}|p{1.3cm}}
         Metric & Mott & Adaptive & sp\_temp-seq & sp\_temp-batch & Div Space-Time & Joint Space-Time\\
         \hline \hline
         Model size & 2.176M & 1.92M & 1.95M & 1.95M & 0.56M & 0.56M\\
         Training time(h) & 2.80 & 1.63 & 5.74 & 2.55 & 3.60 & 10.1\\
         Inference time(ms) & 19 & 13 & 17 & 22 & 17 &  67\\
         Final return & -20.24 & 20.38 & 20 & 19.79 & 20 & 20.1\\
         AUC & -20.13 & 15.59 & 13.04 & 14.74 & 8.6 & 10\\
        \end{tabular}
    \label{tab:model_comparison_pong}
\end{table}

\subsection{Results on Enduro environment}
With Enduro, all models were able to learn the environment in 30M training steps (Figure \ref{fig:enduro_result}). Mott architecture (Section \ref{sec:motts_architecture_section}) was the slowest to learn. Joint Space-Time model (Figure:[\ref{fig:vit_joint_3d_attention}]) learned the fastest, closely followed by Spatio-Temporal sequential architecture (Figure \ref{fig:spatio_temporal_architectures} Left), in terms of training steps. The unusually flat region in the episode length plot(Figure \ref{fig:enduro_result} Right) represents time instances where an agent has not learned yet and episodes terminate early, roughly around the same number of steps.

\begin{figure}[hbt!]
\centering
\begin{subfigure}{0.5\textwidth}
    \centering
    \includegraphics[width=1.0\linewidth]{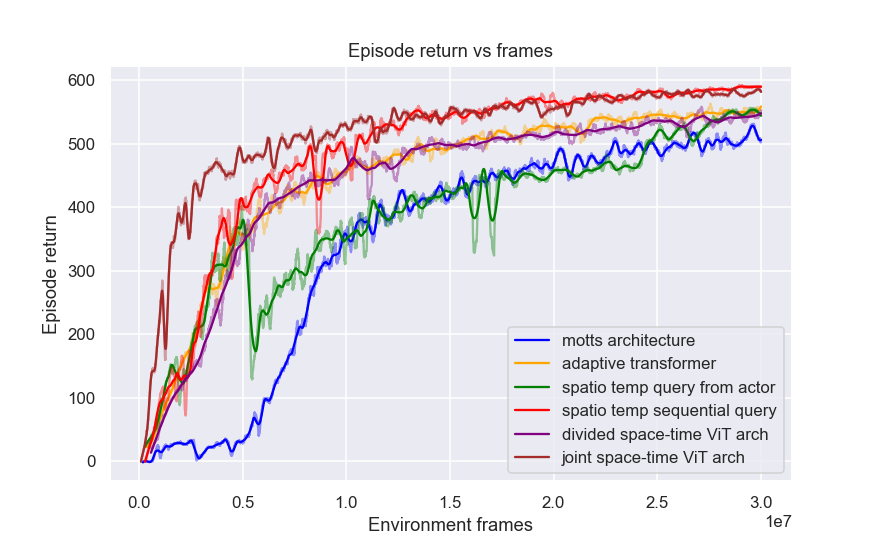}
\end{subfigure}%
\begin{subfigure}{0.5\textwidth}
    \centering
    \includegraphics[width=1.0\linewidth]{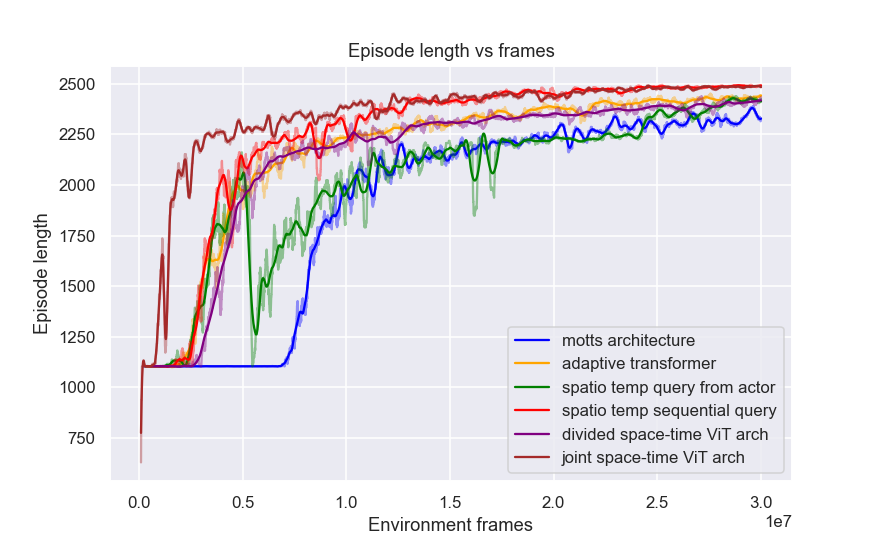}
\end{subfigure}%
\caption{Performance plots for Enduro environment trained for 30M steps.}
\label{fig:enduro_result}
\end{figure}

\begin{table}[hbt!]
    \caption{Comparison of performance metrics of all models on Enduro environment trained for 30M steps.}
    \centering
        \begin{tabular}{p{3.5cm}|p{1.3cm}|p{1.5cm}|p{1.3cm}|p{1.3cm}|p{1.3cm}|p{1.3cm}}
         Metric & Mott & Adaptive & sp\_temp-seq & sp\_temp-batch & Div Space-Time & Joint Space-Time\\
         \hline \hline
         Model size & 2.18M & 1.90M & 1.95M & 1.95M & 0.55M & 0.55M\\
         Training time(h) & 8.78 & 5.34 & 17.14 & 8.27 & 6.675 & 31.41\\
         Inference time(ms) & 35 & 11 & 17 & 22 & 16 &  66\\
         Final return & 504 & 553 & 590 & 548 & 552 & 582\\
         AUC & 340 & 446 & 482 & 385 & 436 & 513\\
        \end{tabular}
    \label{tab:model_comparison_enduro}
\end{table}

\subsection{Results on Pacman environment}
On Pacman, a relatively challenging environment compared to Pong and Enduro, all models except Mott model(Section \ref{sec:motts_architecture_section}) learned to navigate through the environment, dodging hostile agents. In Pacman too, Spatio-Temporal sequential architecture (Figure \ref{fig:spatio_temporal_architectures} Left) learned the environment fastest, closely followed by Divided Space-Time model (Figure \ref{fig:vit_attention_scheme} Right), in terms of number of training steps. Also, here the models needed to be trained longer till 60M steps to observe learning.

\begin{figure}[hbt!]
\centering
\begin{subfigure}{0.5\textwidth}
    \centering
    \includegraphics[width=1.0\linewidth]{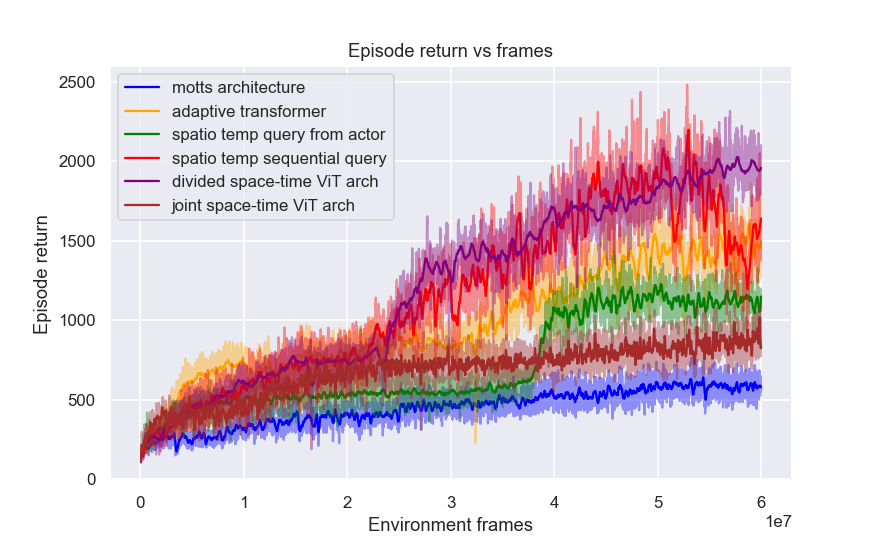}
\end{subfigure}%
\begin{subfigure}{0.5\textwidth}
    \centering
    \includegraphics[width=1.0\linewidth]{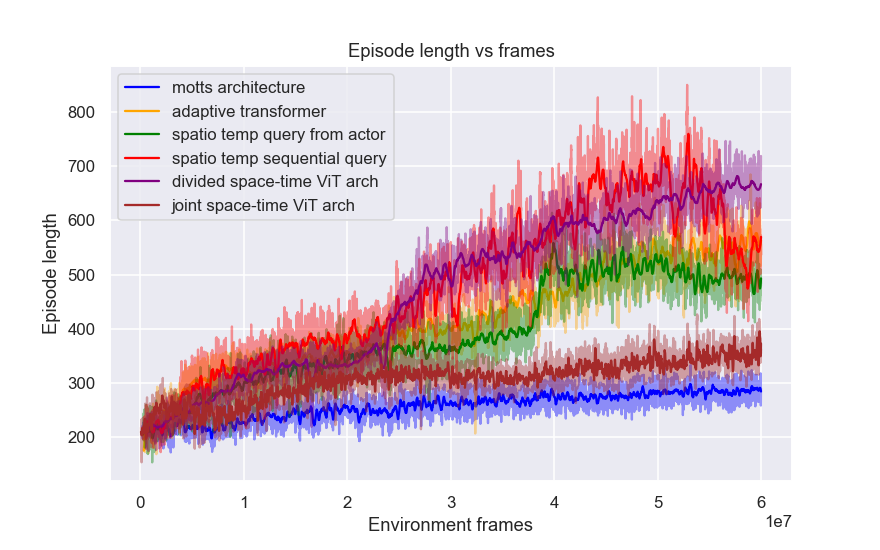}
\end{subfigure}%
\caption{Performance plots for Pacman environment trained for 60M steps.}
\label{fig:pacman_result}
\end{figure}

\begin{table}[hbt!]
    \caption{Comparison of performance metrics of all models on Pacman environment trained for 60M steps. (Joint space model is with env\_shape $(42 \times 42)$}
    \centering
        \begin{tabular}{p{3.5cm}|p{1.3cm}|p{1.5cm}|p{1.3cm}|p{1.3cm}|p{1.3cm}|p{1.3cm}}
         Metric & Mott & Adaptive & sp\_temp-seq & sp\_temp-batch & Div Space-Time & Joint Space-Time\\
         \hline \hline
         Model size & 2.18M & 1.90M & 1.95M & 1.95M & 0.55M & 0.22M\\
         Training time(h) & 18.42 & 9.81 & 34.42 & 15.76 & 12.36 & 9.70\\
         Inference time(ms) & 19 & 12 & 17 & 22 & 17 & 10\\
         Final return & 584 & 1333 & 1472 & 1007 & 2100 & 855\\
         AUC & 443 & 978 & 1147 & 766 & 1204 & 674\\
        \end{tabular}
    \label{tab:model_comparison_pacman}
\end{table}

\subsection{Results on Breakout environment}
With Breakout environment, models learned in varying extent, with TimeSformer Divided Space-Time model (Figure \ref{fig:vit_attention_scheme} Right) performing the best (Figure \ref{fig: breakout_result}), closely followed by Spatio-Temporal sequential architecture (Figure \ref{fig:spatio_temporal_architectures} Left). Mott model(Section \ref{sec:motts_architecture_section}) displayed the slowest learning in terms of number of training steps. Similar to Pacman, Breakout required the models to be trained longer, till 60M steps. Compared to other environments, Breakout's episode-returns and episode-lengths were very noisy in nature. Hence, the smoothing window had to be increased from last-100 to last-500 episodes, to generate distinct plots for the architectures.

\begin{figure}[hbt!]
\centering
\begin{subfigure}{0.5\textwidth}
    \centering
    \includegraphics[width=1.0\linewidth]{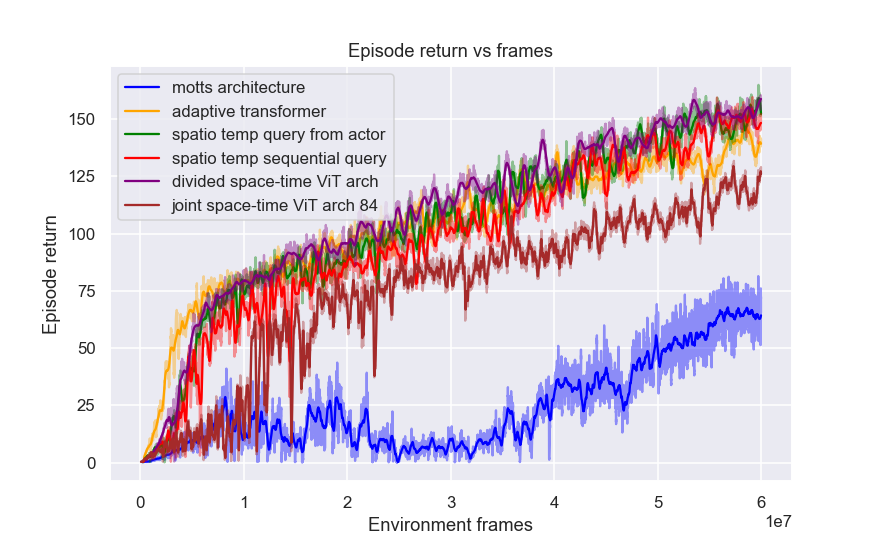}
\end{subfigure}%
\begin{subfigure}{0.5\textwidth}
    \centering
    \includegraphics[width=1.0\linewidth]{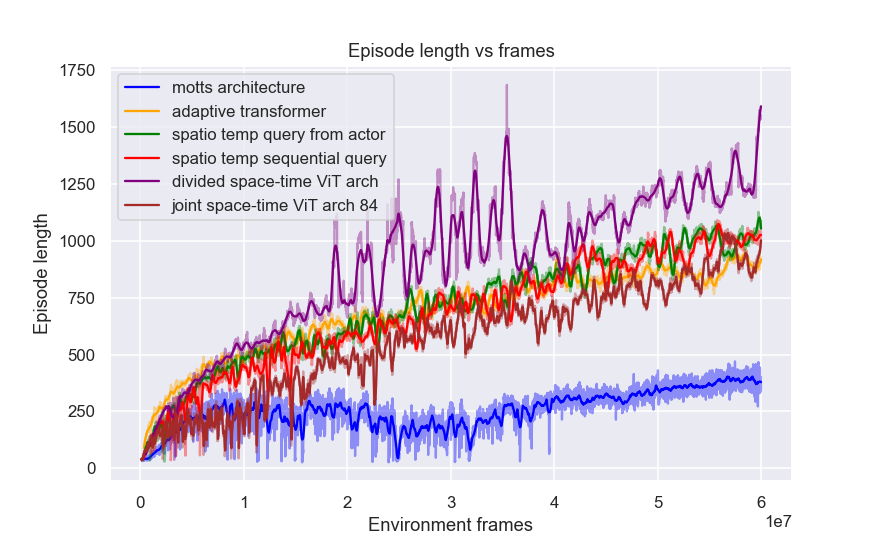}
\end{subfigure}%
\caption{Performance plots for Breakout environment trained for 60M steps.}
\label{fig: breakout_result}
\end{figure}

\begin{table}[hbt!]
    \caption{Comparison of performance metrics of all models on Breakout environment trained for 60M steps. (Joint space model is with env\_shape $(42 \times 42)$}
    \centering
        \begin{tabular}{p{3.5cm}|p{1.3cm}|p{1.5cm}|p{1.3cm}|p{1.3cm}|p{1.3cm}|p{1.3cm}}
         Metric & Mott & Adaptive & sp\_temp-seq & sp\_temp-batch & Div Space-Time & Joint Space-Time\\
         \hline \hline
         Model size & 2.18M & 1.93M & 1.95M & 1.95M & 0.56M & 0.22M\\
         Training time(h) & 17.16 & 9.68 & 34.40 & 15.36 & 11.16 & 59.62\\
         Inference time(ms) & 19 & 12 & 17 & 18 & 16 & 64\\
         Final return & 66 & 149 & 149 & 142 & 141 & 108\\
         AUC & 24 & 90 & 103 & 97 & 104 & 69\\
        \end{tabular}
    \label{tab:model_comparison_breakout}
\end{table}

\begin{table}[hbt!]
    \caption{Mean of model size, training time and inference time for all 5 models across all environments.}
    \centering
        \begin{tabular}{p{3.5cm}|p{1.3cm}|p{1.5cm}|p{1.3cm}|p{1.3cm}|p{1.3cm}|p{1.3cm}}
         Metric & Mott & Adaptive & sp\_temp-seq & sp\_temp-batch & Div Space-Time & Joint Space-Time\\
         \hline \hline
         Model size & 2.18M & 1.91M & 1.95M & 1.95M & 0.56M & 0.39M\\
         Training time(h) & 11.8 & 6.62 & 22.92 & 10.49 & 8.45 & 27.70\\
         Inference time(ms) & 23 & 12 & 17 & 21 & 16.5 & 52\\
        \end{tabular}
    \label{tab:model_comparison_breakout}
\end{table}


\section{Analysis performed}
We performed various analysis on our architectures to come up with the optimum network design and hyper parameters. In the following subsections, we present some of the analysis and impact of network design decisions. \\[0.1in]
Training steps per second (SPS), approximately calculated as $\texttt{SPS} = \texttt{total\textunderscore{steps}}/\texttt{training\textunderscore{time}}$ is considered a measure of training speed. We performed most of our analysis on OpenAI Gym \cite{brockman2016openai} Atari-2600 Pong, and Enduro environments. Besides average episode returns and episode lengths, other metrics considered are: training speed (SPS), total training duration, total network-parameter size and inference time. Inference time is defined as the time lapsed between two consecutive agent steps during inference with a trained model. Inference time is observed and averaged over an episode. Episode return and length are computed as moving average over last 100 episodes.

\subsection{Impact of using ResNet in Vision Core}
We experimented with both a smaller convolutional network (ConvNet) as well the larger ResNet (Figure \ref{fig:resnet_architecture}) in vision core for Mott architecture (Section \ref{sec:motts_architecture_section}). The model with ResNet was bigger in terms of parameter volume and learned faster in terms of training steps, compared to ConvNet(Figure \ref{fig:motts_resnet_frame_stacking_impact} Left). Since there was significant improvement in learning speed with ResNet, for all future architectures, we decided to go ahead with ResNet(Figure \ref{fig:resnet_architecture}). Both networks were trained for 40M steps with similar hyper parameters. Findings are summarized in the Table \ref{tab:resnet_vs_convnet}.

\begin{table}[hbt!]
    \caption{Comparison of performance metrics between ResNet and Convnet Mott models trained on Atari-Enduro environment for 40M steps.}
    \centering
        \begin{tabular}{p{4cm}|p{1.5cm}|p{1.5cm}}
         Metric & ResNet & ConvNet\\
         \hline \hline
         total parameter size   & 2.18M    & 2.01M\\
         total training time &   11.70h  & 6.20h\\
         training speed (SPS) &   949.27  & 1791.16\\
         inference time per cycle & 40.4ms & 18.84ms\\
        \end{tabular}
    \label{tab:resnet_vs_convnet}
\end{table}

\subsection{Impact of pre-processing technique:  stacking}
A common pre-processing technique proposed in Mnih et al.\cite{Mnih2015}, involves downsizing images, converting them to grey scale and stacking last $m$ frames along channel dimension such that observation space \textbf{X} $\in \mathbf{R}^{H \times W \times m}$ with $m = 4$. To analyse impact of frame stacking, two experiments: one with frame stacking with $m=4$ and other without frame stacking such that $m=1$, was run on Mott architecture (Section \ref{sec:motts_architecture_section}). Both model were trained for 40M steps on Atari-Enduro environment. Frame stacking clearly speeds up learning as seen from the performance curves (Figure \ref{fig:motts_resnet_frame_stacking_impact} Right). 

\begin{figure}[hbt!]
\centering
\begin{subfigure}{0.5\textwidth}
    \centering
    \includegraphics[width=1.0\textwidth]{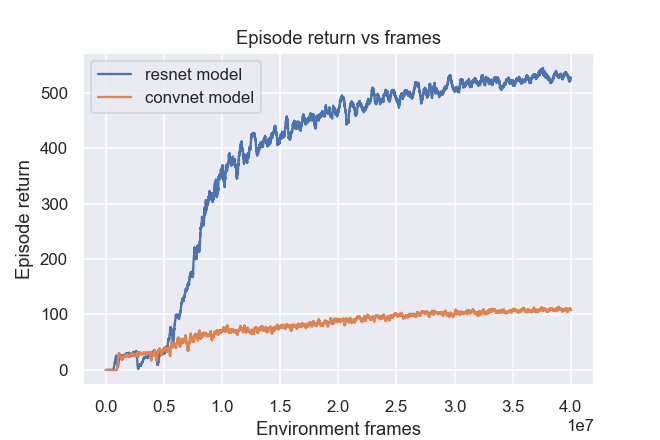}
    \label{fig:mean_returns_resnet_vs_convnet}
\end{subfigure}%
\begin{subfigure}{0.5\textwidth}
    \centering
    \includegraphics[width=1.0\textwidth]{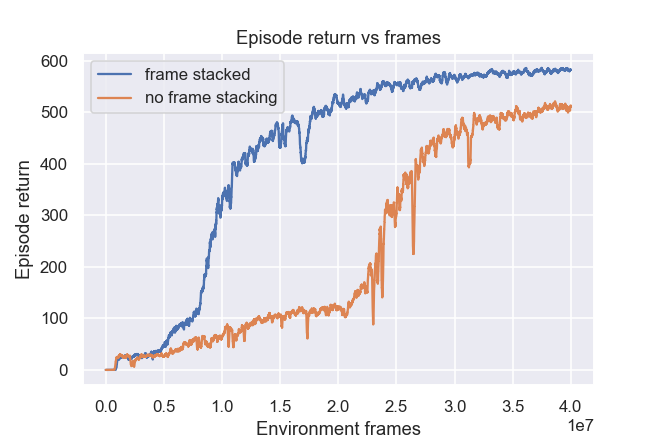}
    \label{fig:mean_returns_frame_stacking}
\end{subfigure}%

\caption{Left: Impact of using ResNet in Vision core. Right: Impact of frame stacking pre-processing technique.}
\label{fig:motts_resnet_frame_stacking_impact}
\end{figure}

\subsection{Effect of varying number of encoder layers in Adaptive architecture}
\label{sec:effect_num_layers_transformer}
We varied the number of encoder layers from 1 to 3 for Adaptive architecture(Section \ref{sec:adaptive_transformer_xl}) on Pong environment trained for 10M steps. Contrary to expectation, increasing the number of layers did not significantly improve the convergence speed, in spite of increasing network capacity. In fact, we noticed that architecture with one and two layers converged faster than the three layers variant(Figure \ref{fig:effect_num_layers_transformer}). We believe that the reason is reactive environments like Pong where time dependencies do not exceed more than 100 time-steps, are not sufficiently memory intensive to fully utilize the Transformer's memory capabilities. As the number of layers increase, total model parameters increases and result in both slower training and inference times(Table \ref{tab:variation_layers_transformer}). Peak in episode-length curve(Figure \ref{fig:effect_num_layers_transformer} Right) represents the training instant where the opponent cpu-agent is able to prolong the game duration by scoring points against the model agent, thereby extending the episode duration. Once the model figures out the optimum move, episode steps are brought down with cpu-agent loosing 0-21 every time thereafter.

\begin{table}[hbt!]
    \centering
    \caption{Effect of variation of number of encoder-layers in Adaptive architecture(Section \ref{sec:adaptive_transformer_xl}) trained for 10M steps.}
        \begin{tabular}{p{4cm}|p{1.5cm}|p{1.5cm}|p{1.5cm}}
         Metric & Layers 1 & Layers 2 & Layers 3\\
         \hline \hline
         total parameter size   & 1.92M  & 2.77M  & 3.63M\\
         total training time &   2.57h  & 3.32h  & 4.02h\\
         training speed (SPS) &   1080.26  & 834.65  & 691.12\\
         inference time per cycle & 14.84ms & 17.46ms & 19.87ms\\
        \end{tabular}
    \label{tab:variation_layers_transformer}
\end{table}

\begin{figure}[hbt!]
\centering
\begin{subfigure}{0.5\textwidth}
    \centering
    \includegraphics[width=1.0\textwidth]{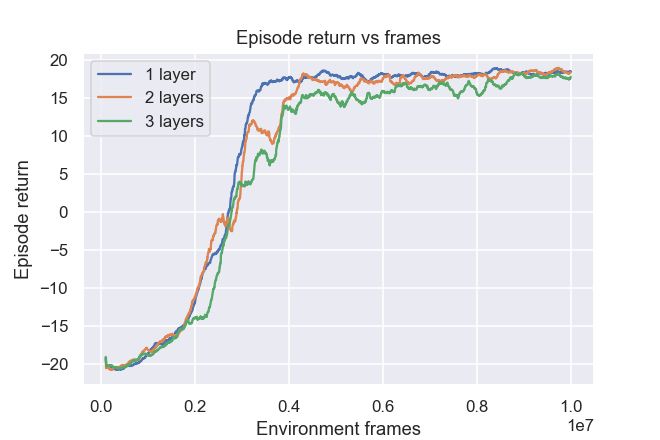}
\end{subfigure}%
\begin{subfigure}{0.5\textwidth}
    \centering
    \includegraphics[width=1.0\textwidth]{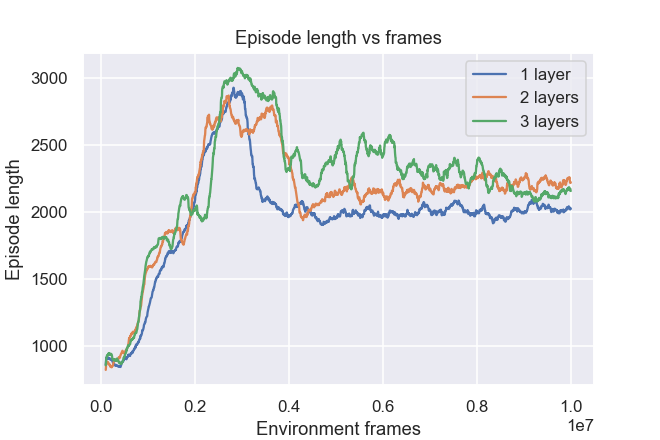}
\end{subfigure}%

\caption{The models \texttt{layer-1}, \texttt{layers-2} and \texttt{layers-3} converge with decreasing speeds with \texttt{layers-3} being the slowest.}
\label{fig:effect_num_layers_transformer}
\end{figure}

\subsection{Effect of scaling images to [0, 1]} 
In this section, we analyze the impact of value normalization technique in which images are scaled from $(0,255)$ to $(0,1)$ via dividing with maximum pixel brightness value 255. To observe the effect of scaling, two Adaptive architecture models: one with and other without rescaling, were tested. Surprisingly, a slight drop in performance is observed in the rescaled case compared to nonscaled ones (Figure \ref{fig:effect_scaling_by_255}). Hence, for all future experiments we decided not to use rescaling. \\[0.1in]

\begin{figure}[hbt!]
\centering
\begin{subfigure}{0.5\textwidth}
    \centering
    \includegraphics[width=1.0\textwidth]{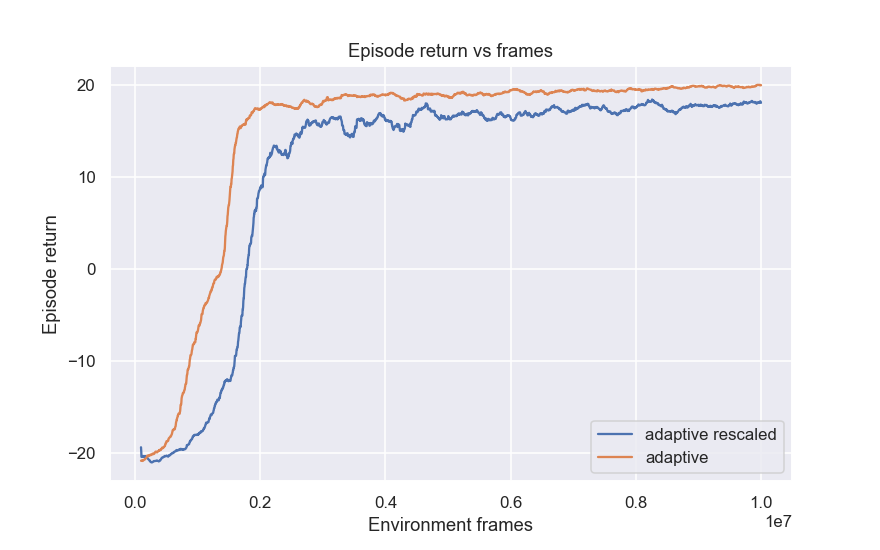}
    \label{fig:effect_scaling_returns}
\end{subfigure}%
\begin{subfigure}{0.5\textwidth}
    \centering
    \includegraphics[width=1.0\textwidth]{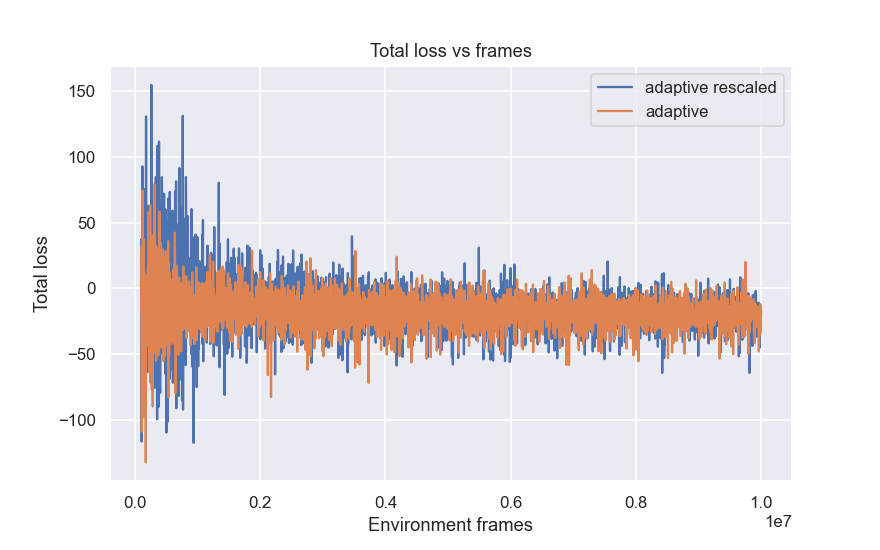}
    \label{fig:effect_scaling_loss}
\end{subfigure}%

\caption{Impact of scaling images to $(0,1)$ on models trained for 10M steps with Adaptive architecture (Section \ref{sec:adaptive_transformer_xl}).}
\label{fig:effect_scaling_by_255}
\end{figure}

\subsection{TimeSformer architecture: Effect of patch size variation}
In TimeSformer variant(Section \ref{sec:timesformer_method}), each frame is decomposed into \textbf{N} non-overlapping patches of size $P \times P$. Total number of patches per frame is given by $N = HW / P^2$. We experimented with three different patch values: 7, 14, and 28 on Pong environment for 60M steps on Divided Space-Time architecture (Figure \ref{fig:vit_attention_scheme} Right). Decreasing patch size would result in more spatial granularity but increases effective sequence length, leading to an increase in training time(Table \ref{tab:vit_patch_size_variation}). In terms of number of training steps, learning speed decreases with an increase in patch size. For instance, patch-size: 7 model learns the fastest, followed by 14, and 28 being the slowest (Figure \ref{fig:patch_frame_size_impact} Left). With patch-size 28, image is split into just 9 patches and the agent might be loosing valuable pixel information in coarse patch embedding operation. We have chosen patch size 7 for further experiments for better attention-visualisation and spatial resolution. \\[0.1in]

\begin{table}[hbt!]
    \caption{Impact of patch size variation in ViT Divided Space-Time model for Pong environment.}
    \centering
        \begin{tabular}{p{4cm}|p{2cm}|p{2.2cm}|p{2.2cm}}
         Metric & patch-size 7 & patch-size 14 & patch-size 28\\
         \hline \hline
         total parameter size & 0.589M    & 0.166M    & 0.095M\\
         total training time &  7.24h    & 3.60h    & 2.60h\\
         training speed (SPS) & 2301.22  & 4627.01  & 6432.63\\
         inference time per cycle & 6.8ms & 3.3ms & 2.64ms\\
        \end{tabular}
    \label{tab:vit_patch_size_variation}
\end{table}

\begin{figure}[hbt!]
\centering
\begin{subfigure}{0.5\textwidth}
    \centering
    \includegraphics[width=1.0\textwidth]{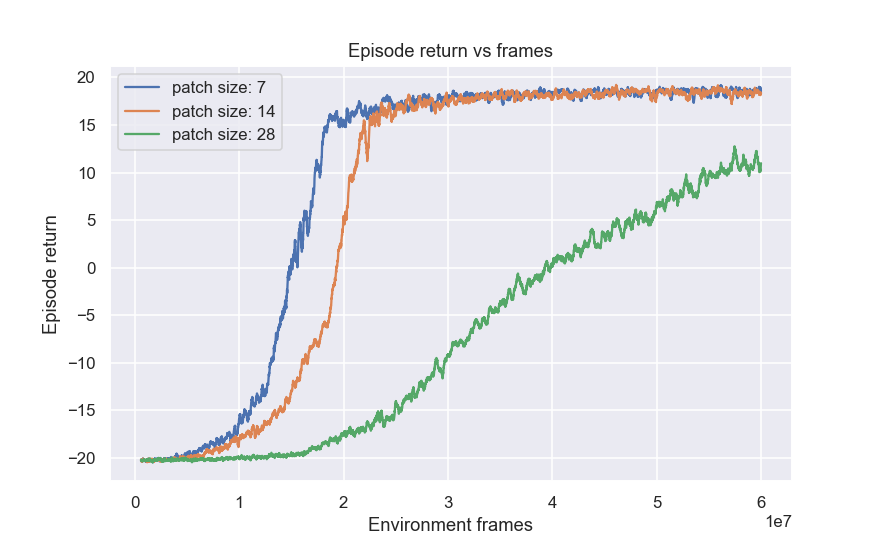}
\end{subfigure}%
\begin{subfigure}{0.5\textwidth}
    \centering
    \includegraphics[width=1.0\textwidth]{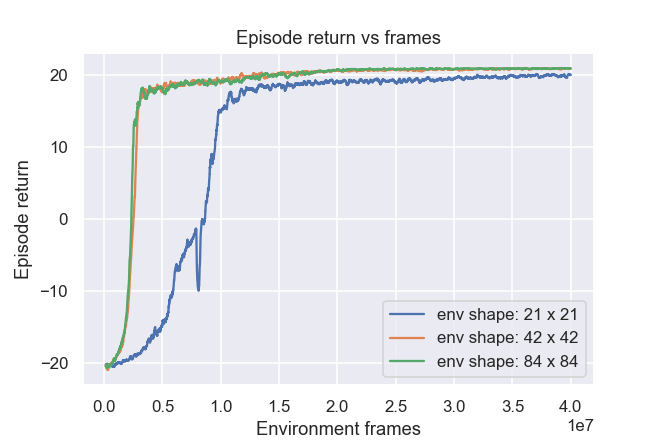}
\end{subfigure}%
\caption{Impact of patch size(Left) and frame size variation(Right) for TimeSformer Divided Space-Time model(Section \ref{sec:timesformer_method}) in Pong environment.}
\label{fig:patch_frame_size_impact}
\end{figure}

\subsection{TimeSformer architecture: Effect of environment size variation}
\label{sec:vit_env_size_impact}
Following the preprocessing techniques used in Mnih et al. \cite{Mnih2015}, we downsize, convert to gray scale and frame-stack the last 4 RGB Atari frames from $(210 \times 160 \times 3)$ to $(84 \times 84 \times 4)$. Higher image resolution results in more number of patches and thus longer sequence length(N) per frame. We additionally experimented with frame sizes 42 and 21 using Divided Space-Time model(Figure \ref{fig:vit_attention_scheme} Right). Training time increases with environment size since time complexity for Divided Space-Time model scales with $(N + F)$ (Table \ref{tab:vit_env_size_impact}). The impact is more profound for Joint Space-Time attention model (Figure \ref{fig:vit_attention_scheme} Left) where the run time directly scales with $(N \times F)$. Hence down-scaling from $(84 \times 84)$ to $(42 \times 42)$ sped up training tremendously for Joint Space-Time attention model (Figure \ref{fig:vit_attention_scheme} Left). The increase in total parameter size with environment size is due to the growth of parameter volume of Linear layer following Policy-core (Section \ref{sec:timesformer_method}). The Linear layer is responsible for flattening and shrinking Policy core output to a lower-dimensional space. Hence, the total parameter size of Linear layer scales quadratically with the environment size. Also, the training performance deteriorated on reducing the environment shape to $(21 \times 21)$ (Figure \ref{fig:patch_frame_size_impact} Right).

\begin{table}[hbt!]
    \caption{Impact of variation in environment frame shape for Divided Space-Time architecture (Figure \ref{fig:vit_attention_scheme} Right)}
    \centering
        \begin{tabular}{p{4cm}|p{2cm}|p{2cm}|p{2cm}}
         Metric & env-size 21 & env-size 42 & env-size 84\\
         \hline \hline
         total parameter size   & 0.11M    & 0.22M & 0.56M\\
         total training time &   2.36h  & 3.68h  & 7.61h\\
         training speed (SPS) &   4696.37  & 3018.12  & 1459.69\\
         inference time per cycle & 4.5ms & 6.5ms & 17ms\\
        \end{tabular}
    \label{tab:vit_env_size_impact}
\end{table}

\subsection{TimeSformer architecture: Effect of Hybrid architecture}
Similar to Hydrid architecture used in ViT (Section \ref{sec:vision_transformer_background}), we compare the effect of using ResNet \cite{DBLP:journals/corr/HeZRS15} and a single-layer 2D-convolutional(ConvNet) network similar to TimeSformer model \cite{DBLP:journals/corr/abs-2102-05095} to generate image-patches sequences. Similar to the trends observed in Mott (Figure \ref{fig:motts_resnet_frame_stacking_impact} Left), significant improvement in training performance has been observed with ResNet \cite{DBLP:journals/corr/HeZRS15} configuration (Figure \ref{fig:vit_resnet_convnet_embsize_impact} Left). Hence for all subsequent experiments, we used ResNet as the patch-embedding network. However, the improvement in training performance comes with a price of increase in total training time and slower inference (Table \ref{tab:vit_effect_resnet}). Another interesting observation here is that the total parameter size is more for ConvNet based model compared to ResNet based model, even though ResNet (\ref{fig:resnet_architecture}) is more parameter intensive that the single layer CNN. Similar to observation in Section \ref{sec:vit_env_size_impact}, this is again attributed to substantial increase in the parameter size of the Linear layer responsible for shrinking Policy core's output. Output frame size of ConvNet layer is $(12 \times 12)$ whereas it is $(11 \times 11)$ for ResNet. Since size of the mapping Linear layer scales quadratically with the output frame size, this leads to Hybrid architecture using ResNet having fewer parameters than the normal architecture using ConvNet.

\begin{table}[hbt!]
    \caption{Impact of Hydrid architecture with ResNet compared to normal convolutional network for Divided Space-Time model.}
    \centering
        \begin{tabular}{p{4cm}|p{2cm}|p{2cm}}
         Metric & ConvNet & ResNet\\
         \hline \hline
         total parameter size   & 0.59M    & 0.56M\\
         total training time &   4.84h  & 7.61h\\
         training speed (SPS) &   2294.42  & 1459.7\\
         inference time per cycle & 6.5ms & 17ms\\
        \end{tabular}
    \label{tab:vit_effect_resnet}
\end{table}

\begin{figure}[hbt!]
\centering
\begin{subfigure}{0.5\textwidth}
    \centering
    \includegraphics[width=1.0\textwidth]{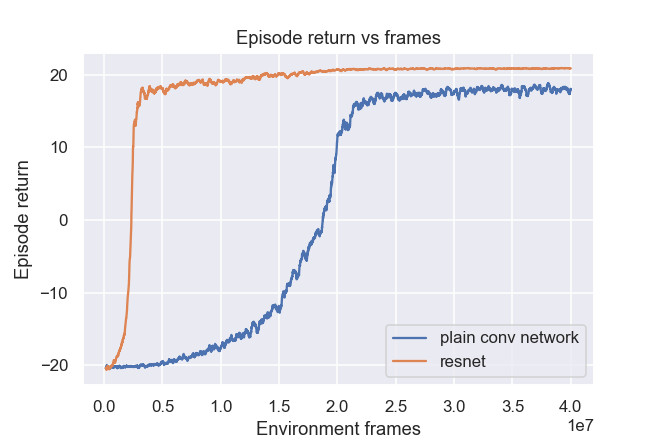}
\end{subfigure}%
\begin{subfigure}{0.5\textwidth}
    \centering
    \includegraphics[width=1.0\textwidth]{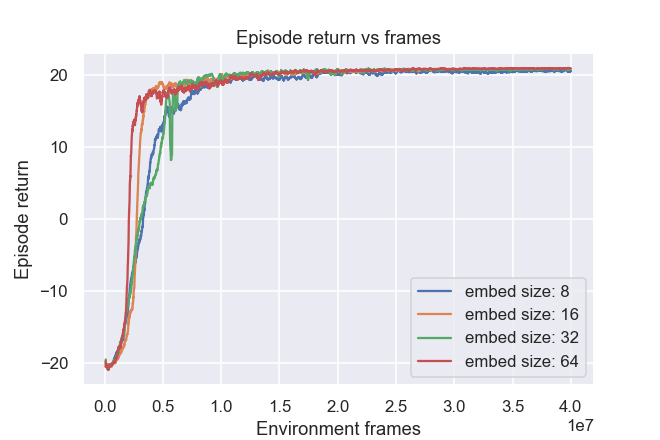}
\end{subfigure}%
\caption{Impact of using ResNet as patch embedding network(Left) and variation of \texttt{emb\textunderscore{size}}(Right) for TimeSformer Divided Space-Time model(Section \ref{sec:timesformer_method}) with Pong environment.}
\label{fig:vit_resnet_convnet_embsize_impact}
\end{figure}

\subsection{TimeSformer architecture: Effect of embedded dimension for Patch embedding network}
Patch Embedding network encodes input frame $X$ $\in \mathbf{R}^{H \times W \times C}$ to $X'$ $\in \mathbf{R}^{h \times w \times emb\textunderscore{size}}$ which are in turn flattened to $X''$ $\in \mathbf{R}^{N \times P^2 \times emb\textunderscore{size}}$. Hyper parameters \texttt{emb\textunderscore{size}} determines latent space size of encoded input frame. Increasing \texttt{emb\textunderscore{size}} increases network capacity at the expense of training time. We experimented with values: $[8, 16, 32, 64]$ for \texttt{emb\textunderscore{size}} on Pong environment for 40M training steps. Training time increases significantly with higher values of \texttt{emb\textunderscore{size}} with no proportionate improvement in returns or learning speed (Table \ref{tab:vit_emb_dim_variation} and Figure \ref{fig:vit_resnet_convnet_embsize_impact} Right). We have tuned \texttt{emb\textunderscore{size}} to 16 to trade-off training speed and model capacity for our subsequent experiments. \\[0.1in]

\begin{table}[hbt!]
    \caption{Impact of variation in \texttt{emb\textunderscore{size}} for Divided Space-Time model with Pong environment.}
    \centering
        \begin{tabular}{p{4cm}|p{1.5cm}|p{1.5cm}|p{1.5cm}|p{1.5cm}}
         Metric & emb-8 & emb-16 & emb-32 & emb-64\\
         \hline \hline
         total parameter size & 0.30M  & 0.56M & 1.09M & 2.24M\\
         total training time & 6.33h  & 7.64h  & 11.36h & 17.89h\\
         training speed (SPS) &  1754.24 & 1459.69 & 977.55 & 620.86\\
         inference time per cycle & 13.5ms & 16ms & 21.5ms & 33.5ms\\
        \end{tabular}
    \label{tab:vit_emb_dim_variation}
\end{table}

\subsection{TimeSformer architecture: Effect of varying number of attention layers}
Similar to Adaptive architecture experiment (Section \ref{sec:effect_num_layers_transformer}), we varied \texttt{num\textunderscore {layers}} in $[1, 2, 3, 4]$. Increasing the number of layers slightly increased the total parameter size with minor improvement in agent performance (Table \ref{tab:vit_effect_num_layers_observations} and Figure \ref{fig:num_layers_div_join_comparison} Left). This trend is similar to that observed with the Adaptive architecture (Figure \ref{fig:effect_num_layers_transformer}) and could be again attributed to less memory-intensive nature of Atari-Pong environment. Also, adding layers significantly slowed down both training and inference. For all later experiments, we used a single-layer configuration. \\[0.1in]

\begin{table}[hbt!]
    \caption{Impact of variation in number of attention layers in TimeSformer model (Section \ref{sec:timesformer_method}) for Pong environment.}
    \centering
        \begin{tabular}{p{4cm}|p{1.5cm}|p{1.5cm}|p{1.5cm}|p{1.5cm}}
         Metric & layers:1 & layers:2 & layers:3 & layers:4\\
         \hline \hline
         total parameter size & 0.5589M  & 0.5635M & 0.5681M & 0.5727M\\
         total training time & 7.64h  & 10.23h  & 12.63h & 16.15h\\
         training speed (SPS) &  1459.69 & 1086.69 & 879.5 & 688.0\\
         inference time per cycle & 16ms & 22.5ms & 29ms & 35ms\\
        \end{tabular}
    \label{tab:vit_effect_num_layers_observations}
\end{table}

\begin{figure}[hbt!]
\centering
\begin{subfigure}{0.5\textwidth}
    \centering
    \includegraphics[width=1.0\textwidth]{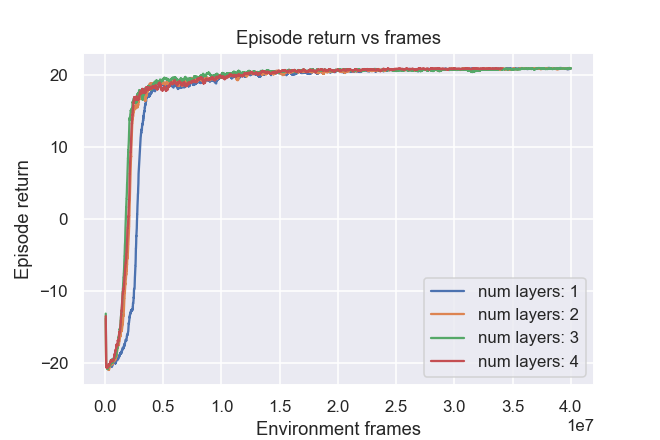}
\end{subfigure}%
\begin{subfigure}{0.5\textwidth}
    \centering
    \includegraphics[width=1.0\textwidth]{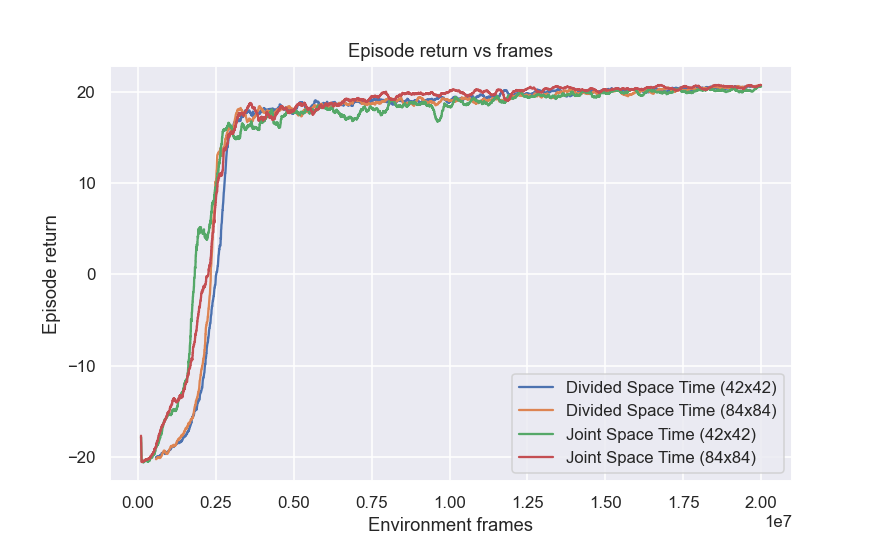}
\end{subfigure}%
\caption{Effect of varying number of attention layers(Left) and performance comparison between Divided Space-Time and Joint Space-Time TimeSformer(Section \ref{sec:timesformer_method}) agents trained on Pong environment.}
\label{fig:num_layers_div_join_comparison}
\end{figure}

\subsection{TimeSformer architecture: Comparison between Divided and Joint Space-Time models}
\label{sec:divided_vs_joint_space_time}
We compare training time and performance characteristics for Divided Space-Time and Joint Space-Time architectures. As mentioned in previous Section \ref{sec:timesformer_theory}, training time of Joint Space-Time agents is of order $\mathcal{O}(N \times F)$, whereas for Divided Space-Time agents, it reduces to $\mathcal{O}(N + F)$, where $N$ represents number of patches per image and $F$ number of frames considered. For both attention schemes, we are considering two models: first one with environment shape $(84 \times 84)$ and second with reduced shape $(42 \times 42)$. Faster training was observed with reduced environment shapes for both Divided and Joint Space-Time models (Table \ref{tab:vit_div_joint_space_time_observations}). Learning performance wise (Figure \ref{fig:num_layers_div_join_comparison} Right), there was no significant difference between the models. All models were trained on Pong environment for 20M steps. Computation wise, it is expensive to run the Joint Space-Time models without environment resizing. \\[0.1in]

\begin{table}[hbt!]
    \caption{Performance comparison of Divided Space-Time and Joint Space-Time agents.}
    \centering
        \begin{tabular}{p{2.5cm}|p{2.0cm}|p{2.0cm}|p{2.0cm}|p{2.0cm}}
         Metric & divided-42 & divided-84 & joint-42 & joint-84\\
         \hline \hline
         total params & 0.221M & 0.5589M  & 0.2165M & 0.5565M\\
         training time & 1.87h & 3.81h  & 3.17h  & 21.12h\\
         SPS & 3018.12 & 1460.06 & 1750.70 & 263.1\\
         inference time & 4.5ms & 16ms & 9.5ms & 81.5ms\\
        \end{tabular}
    \label{tab:vit_div_joint_space_time_observations}
\end{table}

    
    \chapter{Attention Visualizations}
    In the following sections, we visualize various attention schemes used in our architectures. Generally, the technique is to resize the attention map to the original frame size and blend it with the corresponding image using alpha-blending technique. Alpha-blending is the process of overlaying a foreground image with transparency over a background image \cite{opencv_alpha_blending}. The chapter includes following attention visualizations:
\begin{itemize}
    \item Spatial attention visualization for Mott (Section \ref{sec:motts_architecture_section}) architecture.
    \item Temporal attention visualization for Adaptive architecture (Section \ref{sec:adaptive_transformer_xl}).
    \item Spatial attention visualization for Spatio-Temporal sequential architecture  (Figure \ref{fig:spatio_temporal_architectures} Left).
    \item Spatial attention visualization for Spatio-Temporal one-shot architecture using actor query(Figure \ref{fig:spatio_temporal_architectures} Right).
    \item Spatial and temporal attention visualization for TimeSformer: Divided Space-Time architecture (Figure \ref{fig:vit_attention_scheme} Right).
    \item Spatial and Spatio-Temporal attention visualization for TimeSformer: Joint Space-Time architecture (Figure \ref{fig:vit_attention_scheme} Left).
\end{itemize}

We also do perturbation based saliency map \cite{DBLP:journals/corr/abs-1711-00138} analysis detailed in Section \ref{sec:saliency_map_info} for above mentioned architectures.

\section{Attention matrices to Heat maps}
\label{sec:cv_colormap_scale}
We have used open-cv library \cite{opencv_library} to generate RGB attention heat maps $\in \mathbb{R}^{(H\times W\times C)}$ from the attention map $\in \mathbb{R}^{(H\times W\times 1)}$. Attention magnitude increases from minimum:0 (Blue) to maximum:255 (Red) in VIBGYOR order (Figure \ref{fig:cv_colormap_scale}).

\begin{figure}[hbt!]
    \centering
    \includegraphics[width=0.5\textwidth]{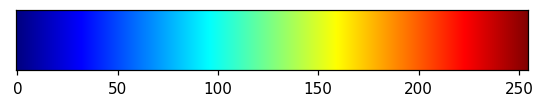}
    \caption{Color map reference in our heat map visualisations.}
    \label{fig:cv_colormap_scale}
\end{figure}

\section{Normalizing attention data}
Before projecting attention (spatial, temporal and spatio-temporal) probability matrices to images, we normalize the attention values in range [0,1] for better visualization. Attention probability data $A$ is normalized as follows,

\begin{equation}
\label{eq:attention_visual_normalization}
    A = \frac{A - A_{min}} {A_{max} - A_{min}}
\end{equation}

\section{Spatial attention visualization of Mott model}
We convert attention map from Equation \ref{eq:motts_spatial_attention_expression} into heat map using open-cv \cite{opencv_library} package and alpha-blend it with corresponding image. Out of all four tested environments, Mott architecture (Section \ref{sec:motts_architecture_section}) could only master the Enduro environment. \\[0.1in]

We normalized the attention matrix in range $[0,1]$ w.r.t to maximum attention value per head, using Equation \ref{eq:attention_visual_normalization}. Four consecutive time-steps (Figure \ref{fig:motts_sp_temp_seq_enduro_sp_attn} Left) are visualized with left most column being the original frame and remaining four columns being the four attention heads. Even though there is attention region on the score-board and on agent at times, most of the time, attention patterns stay static, especially the higher-valued ones in red crimson regions. Nevertheless, generated actor saliency map (Figure \ref{fig:mott_enduro_actor_critic_salmap} Top) focuses on relevant artifacts like incoming cars and curvature of the road. On the other hand, critic saliency map (Figure \ref{fig:mott_enduro_actor_critic_salmap} Bottom) concentrates mostly on agent score. Movies of spatial attention and saliency maps visualization for Mott architecture trained on Enduro environment, can be found at \url{https://imgur.com/a/0El2tmh}.

\begin{figure}[hbt!]
\centering
\begin{subfigure}{0.48\textwidth}
    \centering
    \includegraphics[width=1.0\textwidth]{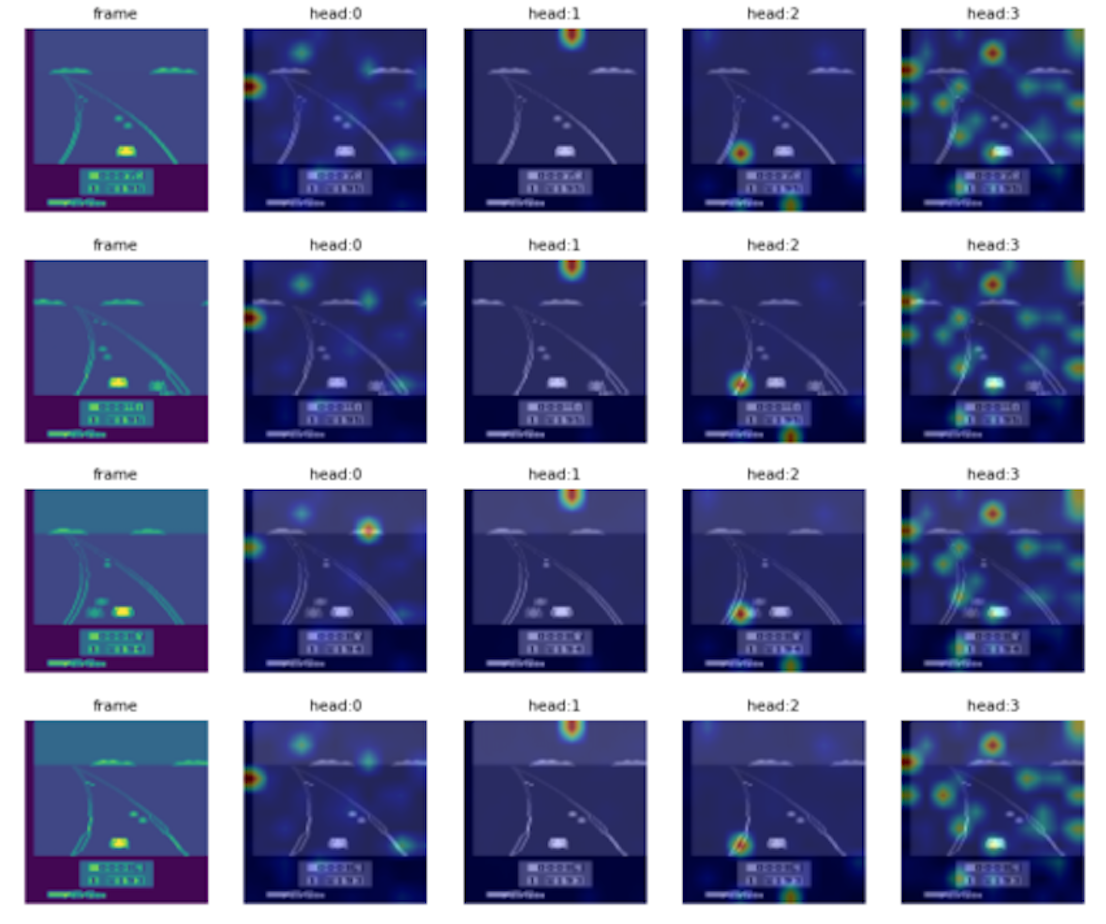}
\end{subfigure}\hspace{0.04\textwidth}%
\begin{subfigure}{0.48\textwidth}
    \centering
    \includegraphics[width=1.0\textwidth]{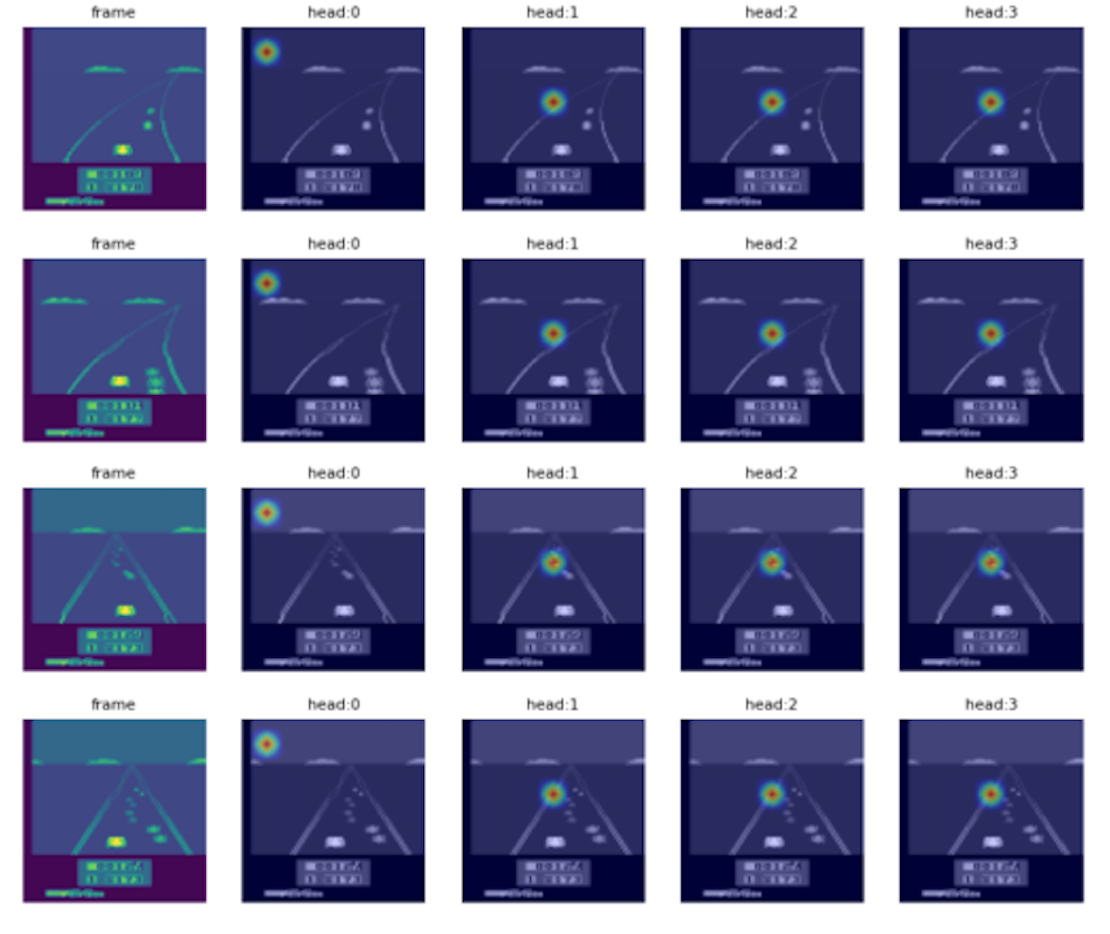}
\end{subfigure}%
\caption{Spatial attention visualization for Mott model (Left) and Spatio-Temporal sequential (Figure \ref{fig:spatio_temporal_architectures} Left) model (Right) on Enduro environment.}
\label{fig:motts_sp_temp_seq_enduro_sp_attn}
\end{figure}

\begin{figure}[hbt!]
\centering
\begin{subfigure}{1.0\textwidth}
    \centering
    \includegraphics[width=1.0\textwidth]{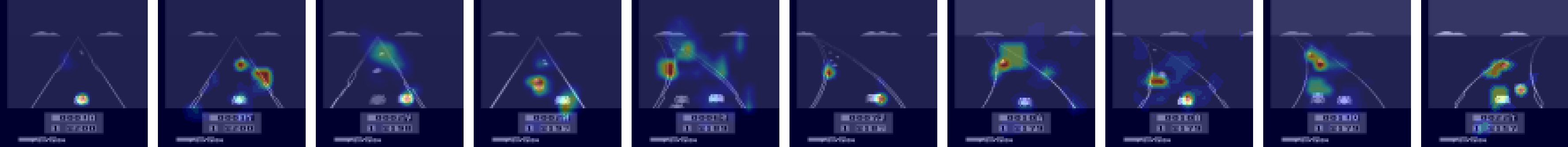}
\end{subfigure}
\par\medskip
\begin{subfigure}{1.0\textwidth}
    \centering
    \includegraphics[width=1.0\textwidth]{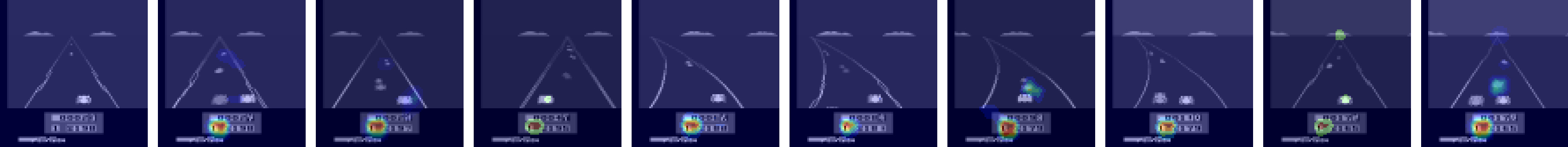}
\end{subfigure}%
\caption{Actor(top) and critic(bottom) saliency map for Mott model on Enduro environment.}
\label{fig:mott_enduro_actor_critic_salmap}
\end{figure}

\section{Spatial attention visualization of Spatio-Temporal sequential model}
Spatio-Temporal sequential architecture (Figure \ref{fig:spatio_temporal_architectures} Left) is closely similar to Mott model (Section \ref{sec:motts_architecture_section}) except that the LSTM core is replaced by a Transformer-XL \cite{DBLP:journals/corr/abs-1901-02860} core. Also, ConvLSTM \cite{DBLP:journals/corr/ShiCWYWW15} is removed from Vision core for Spatio-Temporal architectures. Both models sequentially query current agent state using queries generated from previous time-step outputs. \\[0.1in]

\begin{figure}[hbt!]
\centering
\begin{subfigure}{0.48\textwidth}
    \centering
    \includegraphics[width=1.0\textwidth]{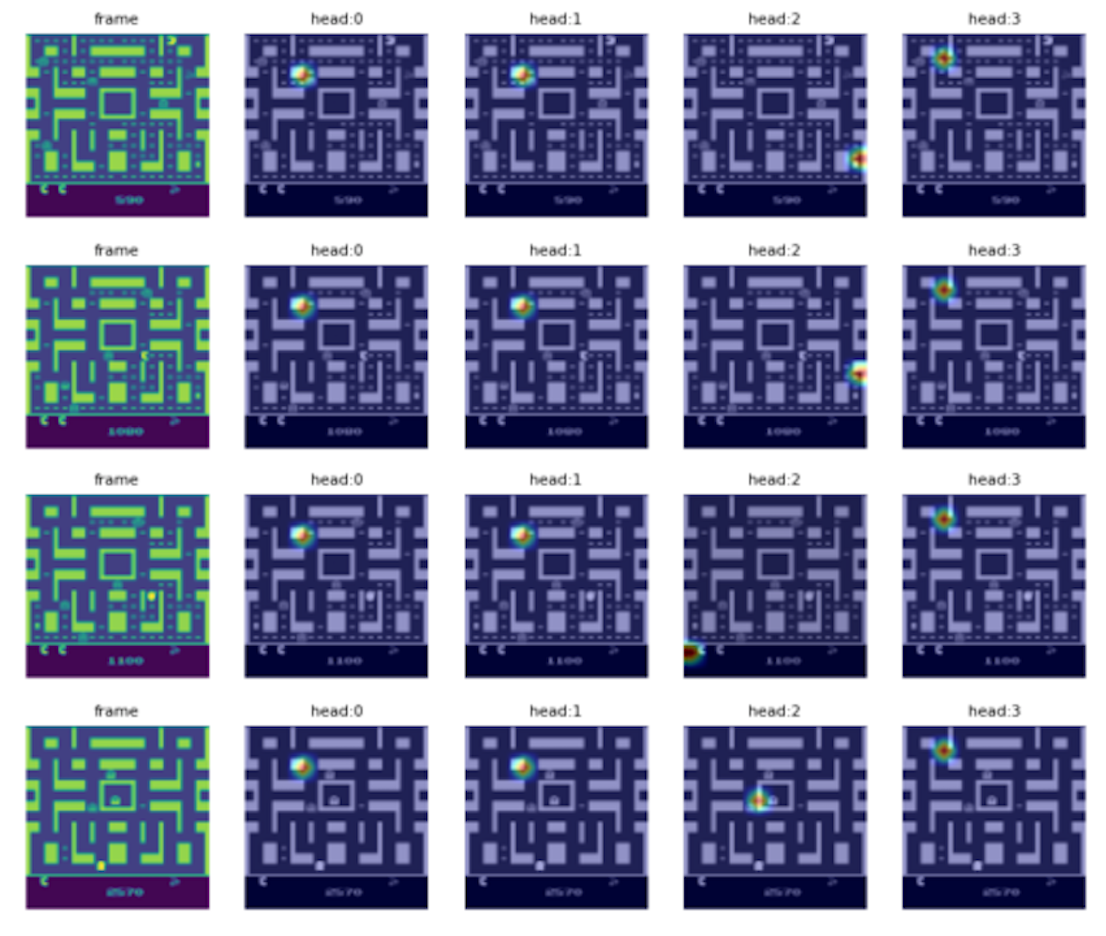}
\end{subfigure}\hspace{0.04\textwidth}%
\begin{subfigure}{0.48\textwidth}
    \centering
    \includegraphics[width=1.0\textwidth]{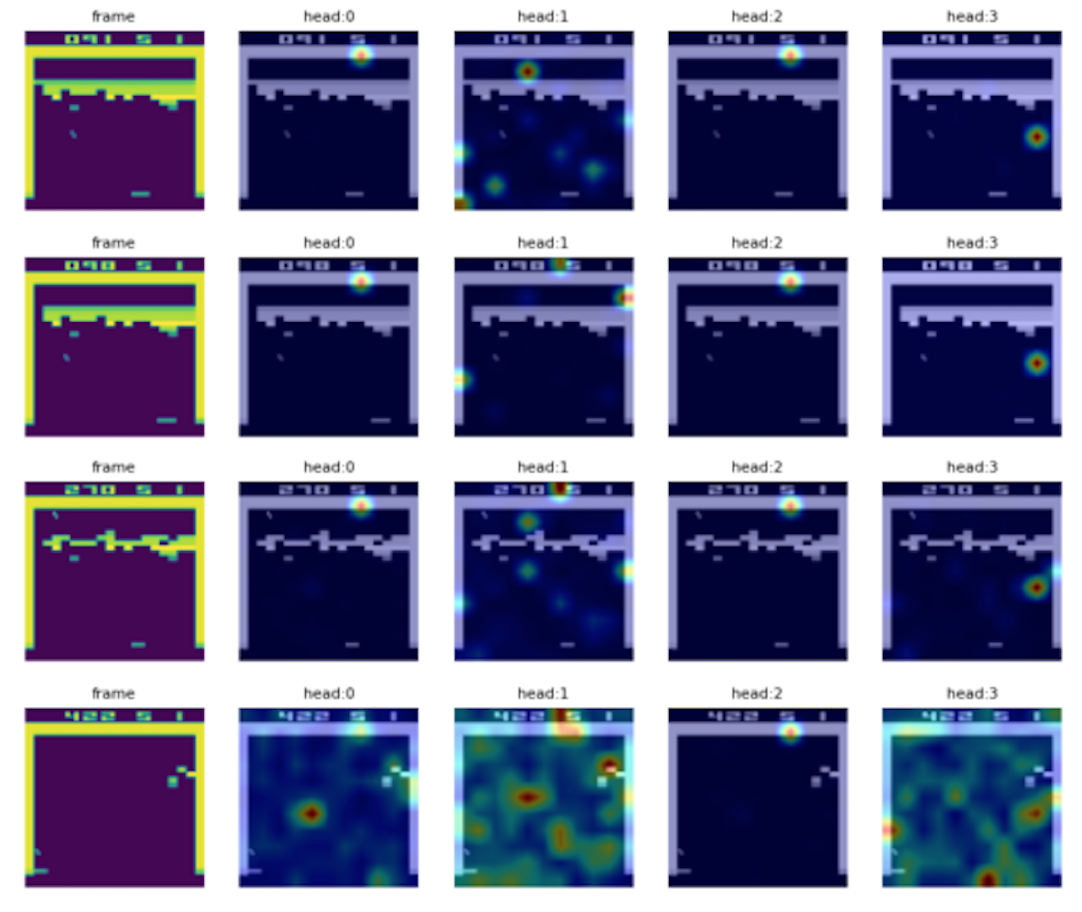}
\end{subfigure}%
\caption{Spatial attention visualization for Spatio-Temporal sequential model on Pacman environment (Left) and Breakout environment (Right).}
\label{fig:sp_temp_seq_pacman_breakout_sp_attn}
\end{figure}

\begin{figure}[hbt!]
\centering
\begin{subfigure}{1.0\textwidth}
    \centering
    \includegraphics[width=1.0\textwidth]{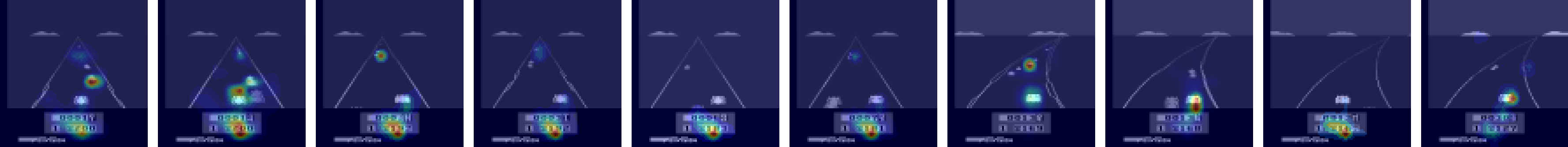}
\end{subfigure}
\par\medskip
\begin{subfigure}{1.0\textwidth}
    \centering
    \includegraphics[width=1.0\textwidth]{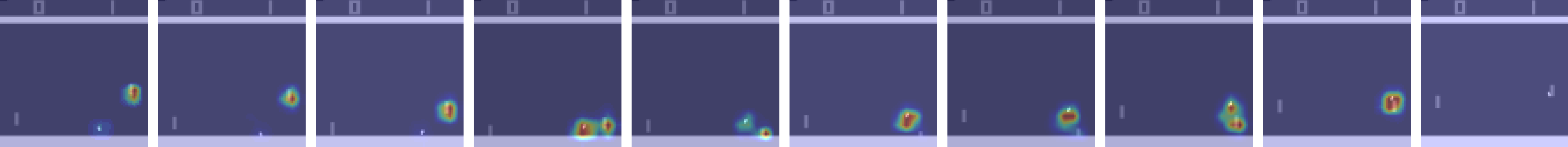}
\end{subfigure}%
\par\medskip
\begin{subfigure}{1.0\textwidth}
    \centering
    \includegraphics[width=1.0\textwidth]{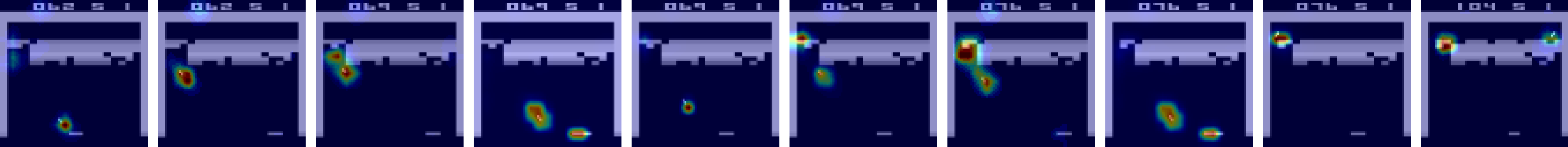}
\end{subfigure}%
\par\medskip
\begin{subfigure}{1.0\textwidth}
    \centering
    \includegraphics[width=1.0\textwidth]{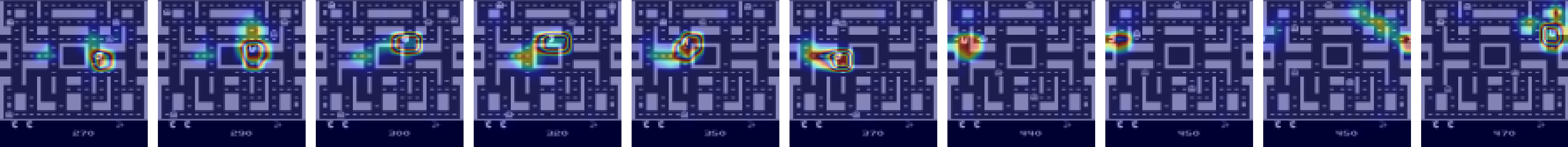}
\end{subfigure}%
\caption{Actor saliency maps for spatio temporal sequential model on Enduro, Pong, Breakout and Pacman (from top to bottom).}
\label{fig:spt_seq_actor_salmap}
\end{figure}

We normalized the attention matrix in range $[0,1]$ w.r.t to maximum attention value per head, using Equation \ref{eq:attention_visual_normalization}. Even though the agent played well on all four environments, attention maps are not actively responding to changes in environment(Figure \ref{fig:sp_temp_seq_pacman_breakout_sp_attn}), similar to the trend observed with Mott model. At times, score relevant artifacts come under attention regions (Figure \ref{fig:motts_sp_temp_seq_enduro_sp_attn} Right, third row, Figure \ref{fig:sp_temp_seq_pacman_breakout_sp_attn} Right, head 1, Figure \ref{fig:spt_seq_batch_pong_attn} Left, head 1), but is not always the case, especially for Enduro(Figure \ref{fig:motts_sp_temp_seq_enduro_sp_attn} Right) and Pacman(Figure \ref{fig:sp_temp_seq_pacman_breakout_sp_attn} Left). Although for Pong (Figure \ref{fig:spt_seq_batch_pong_attn} Left) and Breakout (Figure \ref{fig:sp_temp_seq_pacman_breakout_sp_attn} Right), we do see the attention map responding to state changes. For example, in  Pong(Figure \ref{fig:spt_seq_batch_pong_attn} Left), even though the attention head-0, head-2, and head-3 do not vary much, head-1 changes its distribution in accordance with the environment. For instance, going from second-last to last frame in Figure \ref{fig:spt_seq_batch_pong_attn} Left, one can notice region around the ball drawing more attention. Similarly, in Breakout, there are attention spots over the score-board, bricks and the agent (Figure \ref{fig:sp_temp_seq_pacman_breakout_sp_attn} Right, head-1, second-last and last frame). From the linked clip of Breakout-agent (\url{https://imgur.com/a/G6Bljo2}), agent can be seen to perform `tunneling' operation to maximize returns. Tunneling refers to the agent targeting bricks at top-left and top-right corners to create a path to the inside of the block. However, there are not many convincing patterns in attention visualization to establish the intuition. \\[0.1in]
Nevertheless the saliency map for the model (Figure \ref{fig:spt_seq_actor_salmap}) indicates that agent is sensitive to score relevant artifacts in the image and actively responds to changes.  For instance, in saliency map for Pacman (Figure \ref{fig:spt_seq_actor_salmap} row-iv), last two images show the saliency-score metric changing its focus towards the bonus point at the top right corner in the environment. For Breakout-agent, saliency map shows the agent doing `tunneling' at a very early stage and the saliency map also attends to `tunnel-entrance' region quite strongly (Figure \ref{fig:spt_seq_actor_salmap} row-iii). Movies of spatial attention and saliency maps visualization for Spatio-Temporal sequential model, trained on all four environments can be found at \url{https://imgur.com/a/G6Bljo2}.\\[0.1in]

\textbf{Spatial attention visualization of Spatio-Temporal one-shot sub-architecture} \\[0.1in]
In this section, we visualise spatial attention of Spatio-Temporal one-shot model using actor queries(Figure \ref{fig:spatio_temporal_architectures} Right), which is very similar to Spatio-Temporal sequential model (Figure \ref{fig:spatio_temporal_architectures} Left). Both models differ only in data processing during training, where former uses the actors' cached output and the latter uses the model's sequential output, to query the environment(Section \ref{sec:combine_motts_transformer_xl}). Also, note that during inference, both models use its previous time-step's output to generate query for the current time-step. \\[0.1in]

\begin{figure}[hbt!]
\centering
\begin{subfigure}{0.48\textwidth}
    \centering
    \includegraphics[width=1.0\textwidth]{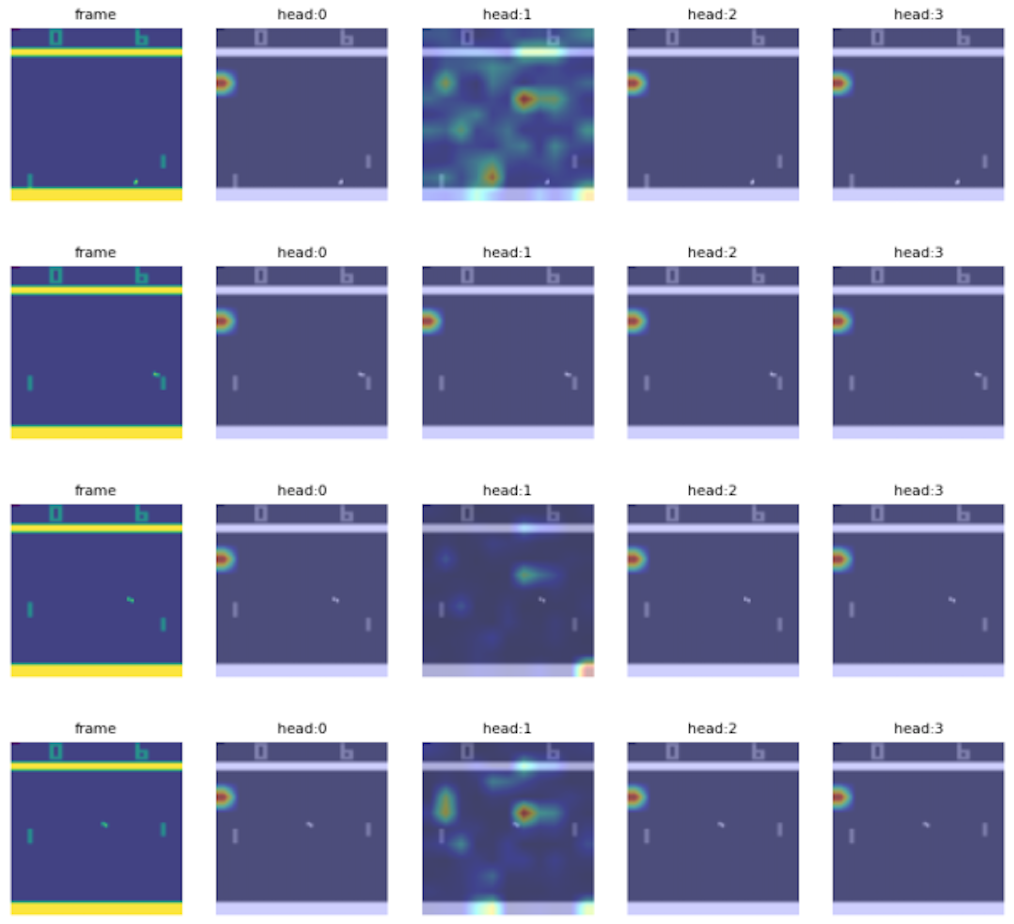}
\end{subfigure}\hspace{0.04\textwidth}%
\begin{subfigure}{0.48\textwidth}
    \centering
    \includegraphics[width=1.0\textwidth]{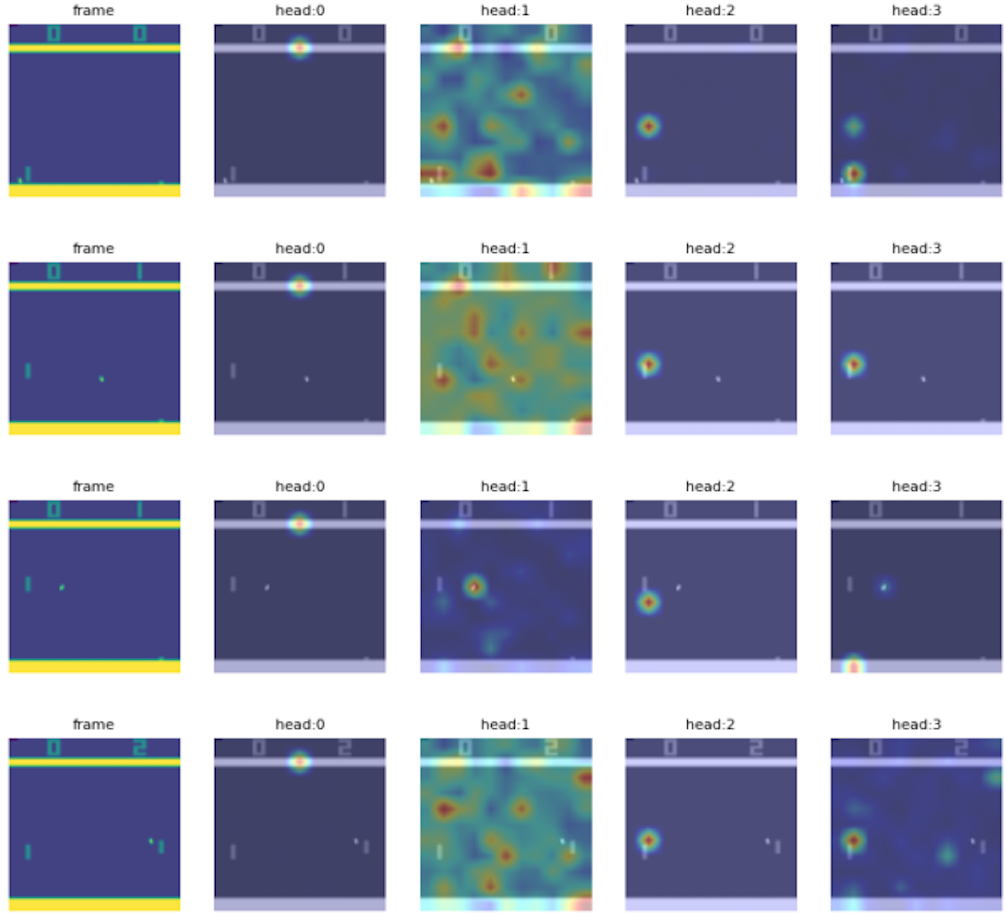}
\end{subfigure}%
\caption{Spatial attention visualization for Spatio-Temporal sequential model(Left) and Spatio-Temporal one-shot model on Pong environment (Right).}
\label{fig:spt_seq_batch_pong_attn}
\end{figure}

\begin{figure}[hbt!]
\centering
\begin{subfigure}{0.48\textwidth}
    \centering
    \includegraphics[width=1.0\textwidth]{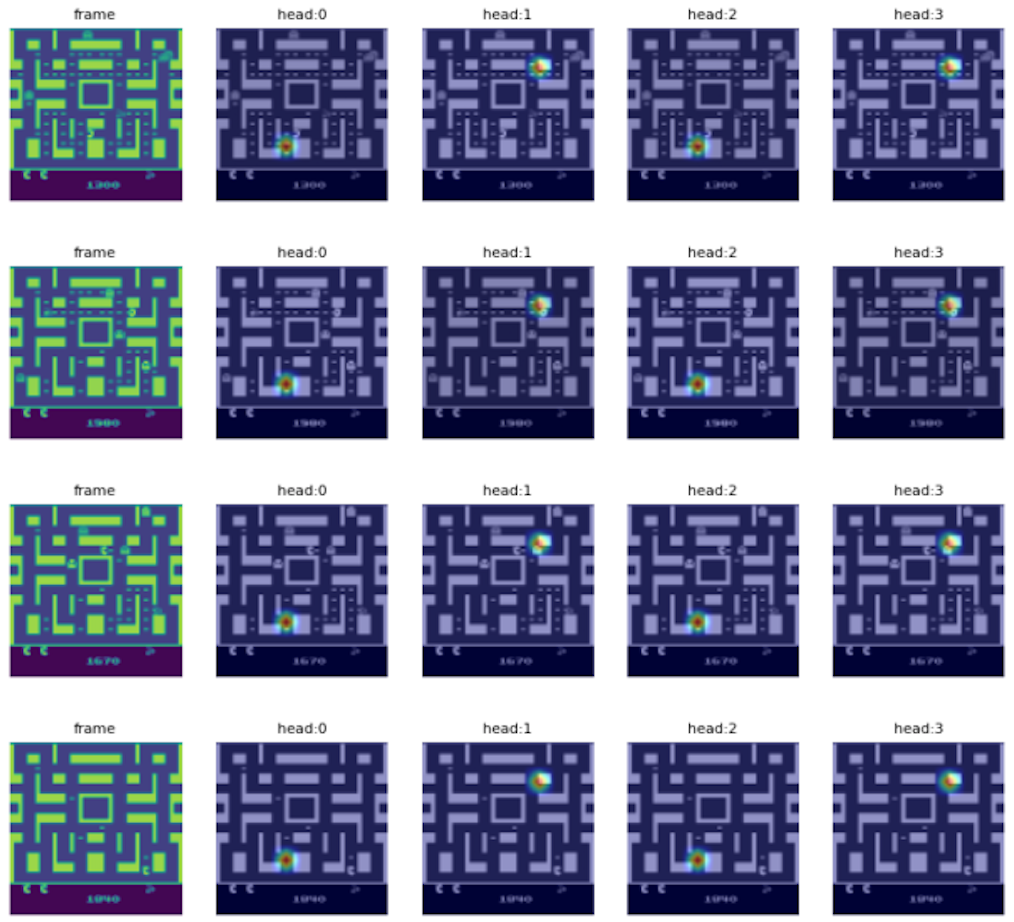}
\end{subfigure}\hspace{0.04\textwidth}%
\begin{subfigure}{0.48\textwidth}
    \centering
    \includegraphics[width=1.0\textwidth]{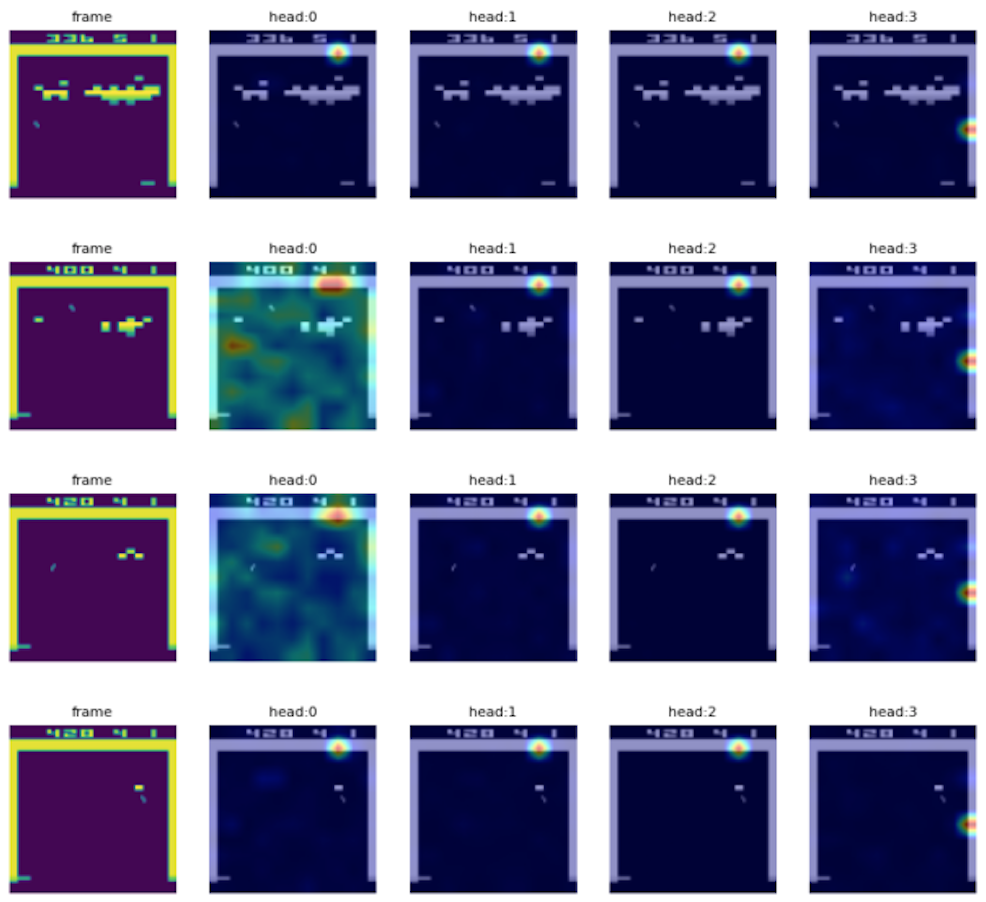}
\end{subfigure}%
\caption{Spatial attention visualization for Spatio-Temporal one-shot model on Pacman (Left) and Breakout (Right) environments.}
\label{fig:spt_batch_pacman_breakout_attn}
\end{figure}

\begin{figure}[hbt!]
\centering
\begin{subfigure}{1.0\textwidth}
    \centering
    \includegraphics[width=1.0\textwidth]{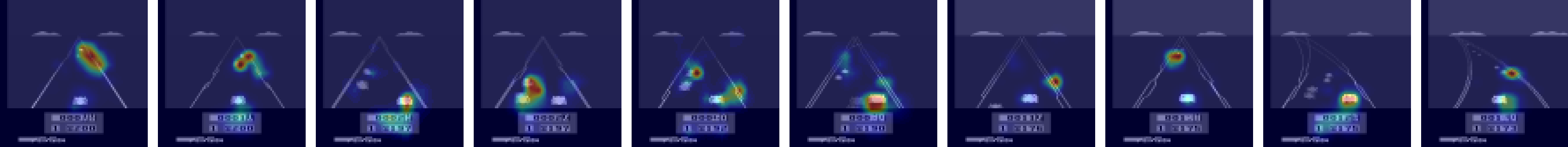}
\end{subfigure}
\par\medskip
\begin{subfigure}{1.0\textwidth}
    \centering
    \includegraphics[width=1.0\textwidth]{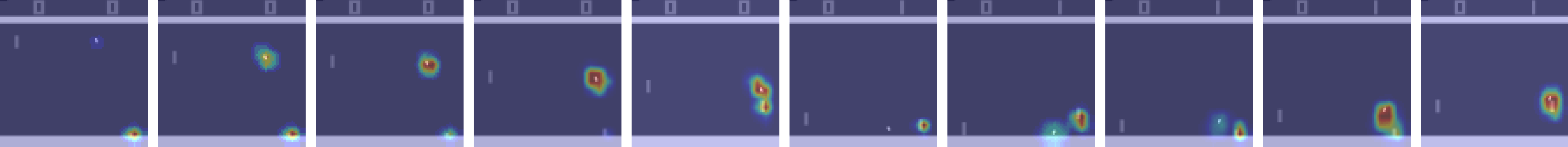}
\end{subfigure}%
\par\medskip
\begin{subfigure}{1.0\textwidth}
    \centering
    \includegraphics[width=1.0\textwidth]{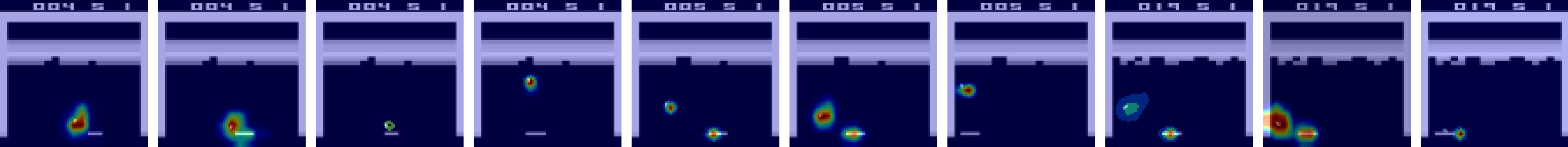}
\end{subfigure}%
\par\medskip
\begin{subfigure}{1.0\textwidth}
    \centering
    \includegraphics[width=1.0\textwidth]{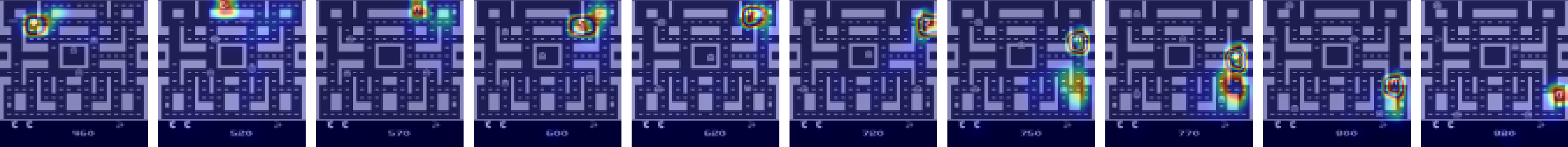}
\end{subfigure}%
\par\medskip
\caption{Saliency maps for Spatio-Temporal one-shot model on Enduro, Pong, Breakout and Pacman(from top to bottom).}
\label{fig:spt_batch_actor_salmap}
\end{figure}

For Spatio-Temporal one-shot model, we notice that except for a few variations with Pong head-1(Figure \ref{fig:spt_seq_batch_pong_attn} Right) and Breakout head-0(Figure \ref{fig:spt_batch_pacman_breakout_attn} Right), attention distribution is more or less static in nature and do not provide much insight into the agent's actions. \\[0.1in]
Nevertheless, saliency map(Figure \ref{fig:spt_batch_actor_salmap}) identifies key regions from the image space for the model. Another interesting observation from the saliency map for both Pong and Breakout environments (Figure \ref{fig:spt_batch_actor_salmap} second and third row) are that the saliency-score progressively increases as the ball approaches the agent in the lower half of the frame and fades out over the upper half. Movies of spatial attention and saliency maps visualization for Spatio-Temporal one-shot model using actor queries, trained on all four environments can be found at \url{https://imgur.com/a/p1Ou7PV}.

\section{Attention visualization of Adaptive architecture}
In this section, we visualize the attention matrix of the Adaptive architecture \texttt{temp\textunderscore attn} $\in \mathbf{R}^{\texttt{qlen} \times \texttt{klen}}$ where \texttt{qlen} denotes current sequence length, \texttt{klen} the total context length including memory tokens and current sequence tokens(Figure \ref{fig:adaptive_pipeline}), i.e. $\texttt{klen} = \texttt{qlen} + \texttt{mlen}$. During inference,  $\texttt{qlen} = 1$ and $\texttt{klen} = 1 + \texttt{mlen}$ where $\texttt{mlen} = 100$ is the number of cached memory tokens from previous step. Hence, attention is applied over $\texttt{mlen} + 1 = 101$ tokens. \\[0.1in]
For Pong environment, we expect the peaks in attention plots (Figure \ref{fig:adaptive_attention_heads}) to represent key-events in the episode like opponent returning the ball or ball bouncing off the walls. Attention probabilities look centered around these key-events in immediate past time-steps. Moreover, as time progresses (along the columns), unless attention splits into multiple sub-peaks, the current attention peak still controls agent's output and slides backwards along time-axis.\\[0.1in]

\begin{figure}[hbt!]
\centering
\begin{subfigure}{0.25\linewidth}
    \centering
    \includegraphics[width=1.0\textwidth]{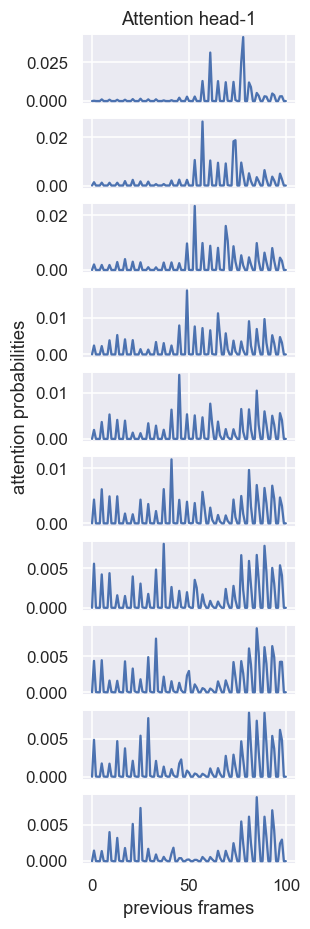}
\end{subfigure}%
\begin{subfigure}{0.25\linewidth}
    \centering
    \includegraphics[width=1.0\textwidth]{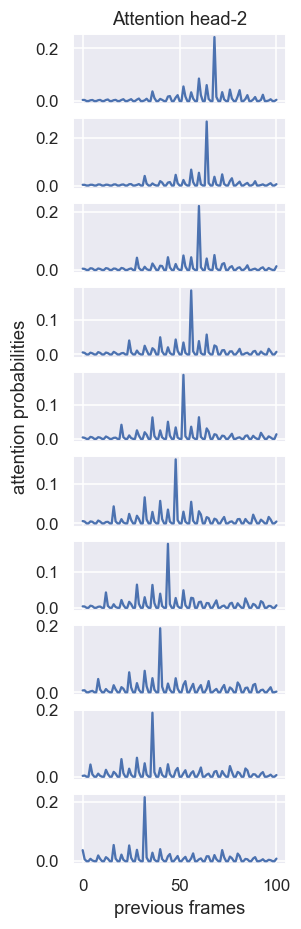}
\end{subfigure}%
\begin{subfigure}{0.25\linewidth}
    \centering
    \includegraphics[width=1.0\textwidth]{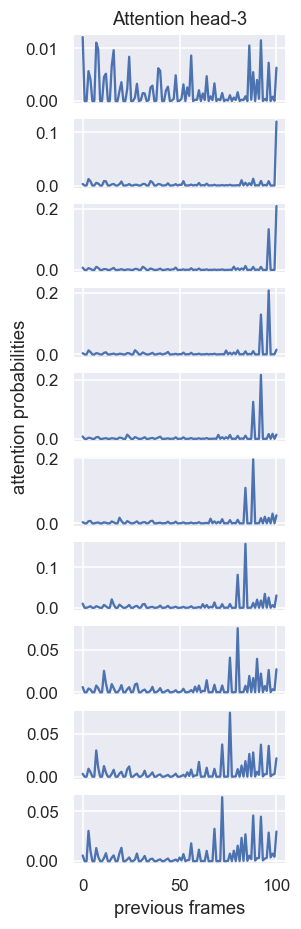}
\end{subfigure}%
\begin{subfigure}{0.25\linewidth}
    \centering
    \includegraphics[width=1.0\textwidth]{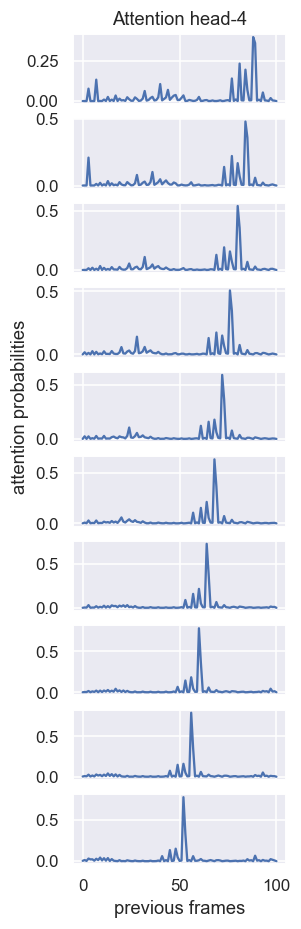}
\end{subfigure}%
\caption{Attention plots for heads 1 to 4 for 10 consecutive time-steps for Pong agent. Topmost row corresponds to $t=t1$ and bottom row to $t=t1+10$. On x-axis, \texttt{frame} varies from (0,100) with $\texttt{frame}=100$ being the current frame.}
\label{fig:adaptive_attention_heads}
\end{figure}

\begin{figure}[hbt!]
\centering
\begin{subfigure}{1.0\textwidth}
    \centering
    \includegraphics[width=1.0\textwidth]{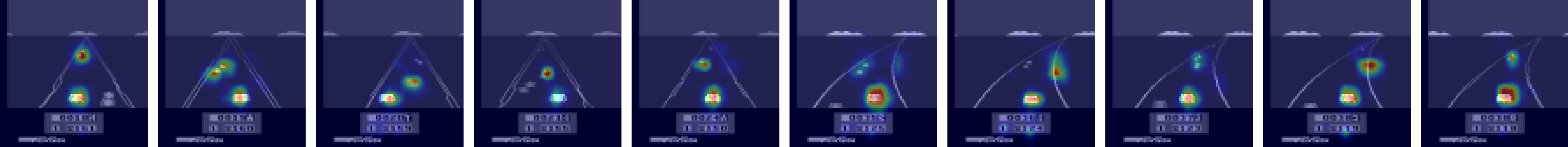}
\end{subfigure}
\par\medskip
\begin{subfigure}{1.0\textwidth}
    \centering
    \includegraphics[width=1.0\textwidth]{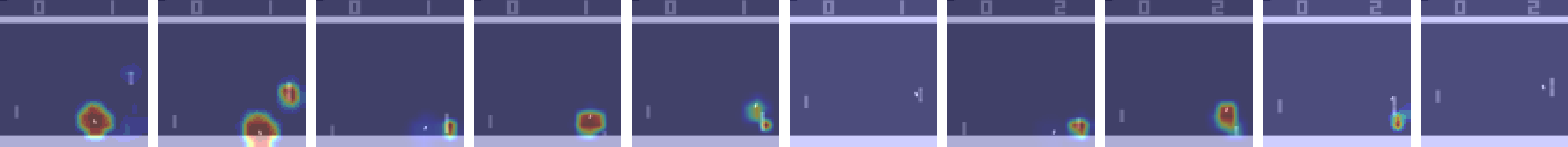}
\end{subfigure}
\par\medskip
\begin{subfigure}{1.0\textwidth}
    \centering
    \includegraphics[width=1.0\textwidth]{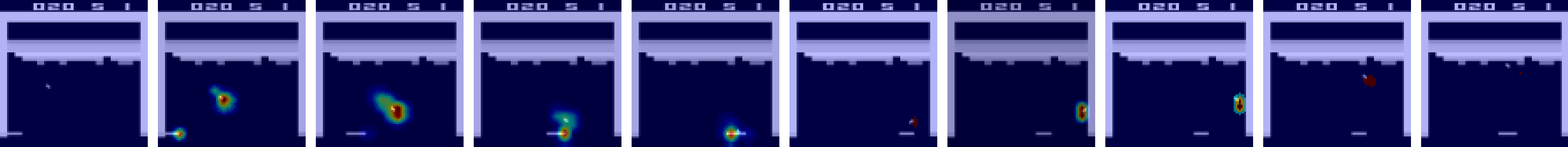}
\end{subfigure}%
\par\medskip
\begin{subfigure}{1.0\textwidth}
    \centering
    \includegraphics[width=1.0\textwidth]{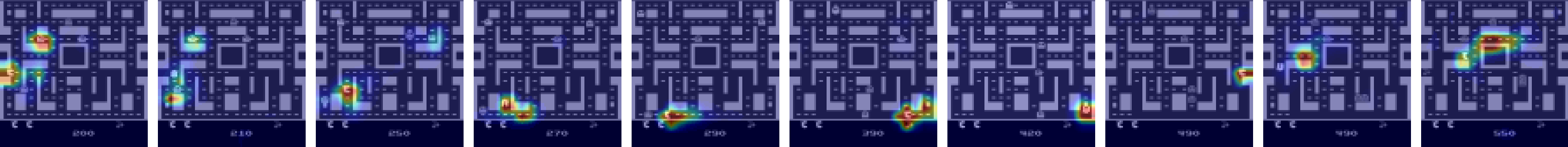}
\end{subfigure}%
\caption{Saliency maps for Adaptive architecture on Enduro, Pong, Breakout and Pacman(from top to bottom).}
\label{fig:adaptive_actor_salmap}
\end{figure}

Additionally, saliency map analysis of the model(Figure \ref{fig:adaptive_actor_salmap}) shows that the agent is actively attending to changes in environment. Also, note that spatial attention map visualization as done in previous models is not possible with Adaptive architecture since the spatial structure of tokens is lost during conversion (Section \ref{sec:adaptive_transformer_xl}). Movies of saliency maps visualization for Adaptive model, trained on all four environments can be found at \url{https://imgur.com/a/aS1sf8X}.

\section{TimeSformer architecture: Visualization of temporal attention}
In this section, we visualize temporal-attention from Divided Space-Time model (Figure \ref{fig:vit_attention_scheme} Right) similar in nature to attention visualization for Adaptive architecture (Figure \ref{fig:adaptive_attention_heads}). TimeSformer temporal-attention matrix $\texttt{temp\textunderscore attn\textunderscore vit}$ $\in \mathbf{R}^{h.w \times klen}$ contains attention probabilities of $(h.w)$ tiles or pixels per image, over $klen$ tokens, where $klen$ represents the total sequence length including the current token and former tokens in memory. Here $(h,w)$ represents the reduced frame size after passing the frames through Patch Embedding network (Figure \ref{fig:vit_block_diagram}) \\[0.1in]
Temporal attention is visualized for all 4 attention heads(Figure \ref{fig:vit_temporal_attention}) over consecutive time frames represented by adjacent rows. To cover more number of time-steps, each row is sampled every $n$th time-step with $n = 8$. Each of the four columns represent the four attention heads \\[0.1in]
\textbf{Observation} Each tile on Figure \ref{fig:vit_temporal_attention} denotes the attention matrix $\texttt{temp\_attn\_vit}$ $\in \mathbf{R}^{h.w \times klen}$ of an image for one head, where $(h.w = 121)$ and $klen = 101$. It can be observed that the sinusoidal looking attention pattern moves backward with time along the rows, similar to the pattern observed with Adaptive architecture (Figure \ref{fig:adaptive_attention_heads}). \\[0.1in]

\begin{figure}[hbt!]
    \centering
    \includegraphics[width=0.5\textwidth]{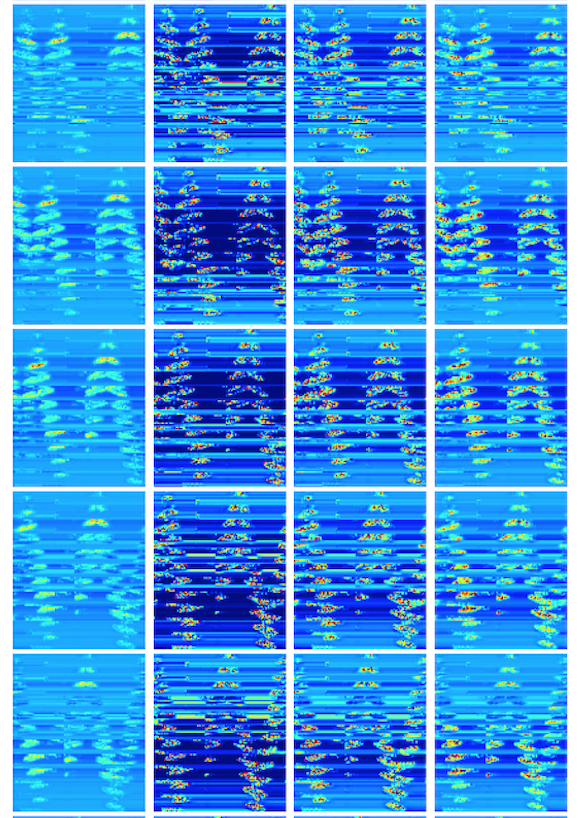}
    \caption{Temporal attention visualization for all 4 attention heads over 40 consecutive time frames, trained on Pong environment.}.
    \label{fig:vit_temporal_attention}
\end{figure}

\section{TimeSformer architecture: Visualization of spatial attention}
In this section, we visualize spatial attention for Divided Space-Time model (Figure \ref{fig:vit_attention_scheme} Right). Attention matrix \texttt{sp\textunderscore attn} $\in \mathbf{R}^{heads\times (h.w)\times (h.w)}$ is reshaped to $(heads\times (h.w)\times h\times w)$ and normalized w.r.t maximum value, where $(h,w)$ represents frame dimensions, \texttt{heads} the number of attention heads. The reshaped tensor is averaged across all the tiles in the image ($\texttt{axis}=1$) to generate final attention matrix \texttt{sp\textunderscore attn\textunderscore mean} $\in \mathbf{R}^{heads\times h\times w}$ which is projected onto images. We normalized the attention matrix in the range $[0,1]$ w.r.t to maximum attention value per head, using Equation \ref{eq:attention_visual_normalization}. \\[0.1in]
In Figure \ref{fig:div_spt_pong_enduro} Left, Pong agent trained for 60M steps is visualized. At the begining of the game, when there is no ball in frame (first-row), head-0 has its attention over the opponent agent. Additionally, patches of `attention-clouds' follow the ball as game progresses (second row, all attention heads and second-last row, head-0, head-1 and head-3). \\[0.1in]
An Enduro agent trained for 30M steps is visualized on Figure \ref{fig:div_spt_pong_enduro} Right. First row, head-0 shows agent having attention on distant incoming cars and on itself too. Attention head-2 does not give much insight since it is having a major share of its attention on the horizon and outside track. Another interesting observation is the arched attention pattern over the bent track in the last three rows, head-3. \\[0.1in]

\begin{figure}[hbt!]
\centering
\begin{subfigure}{0.48\textwidth}
    \centering
    \includegraphics[width=1.0\textwidth]{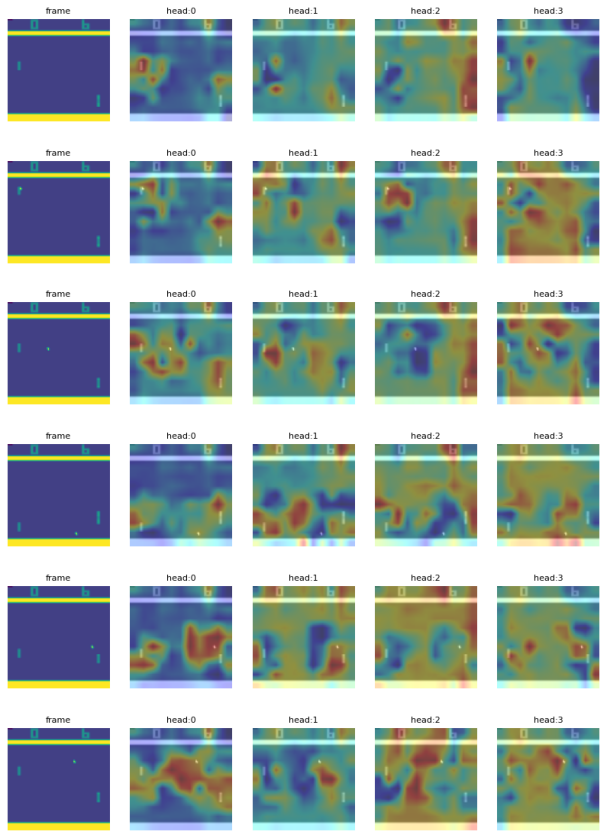}
\end{subfigure}\hspace{0.04\textwidth}%
\begin{subfigure}{0.48\textwidth}
    \centering
    \includegraphics[width=1.0\textwidth]{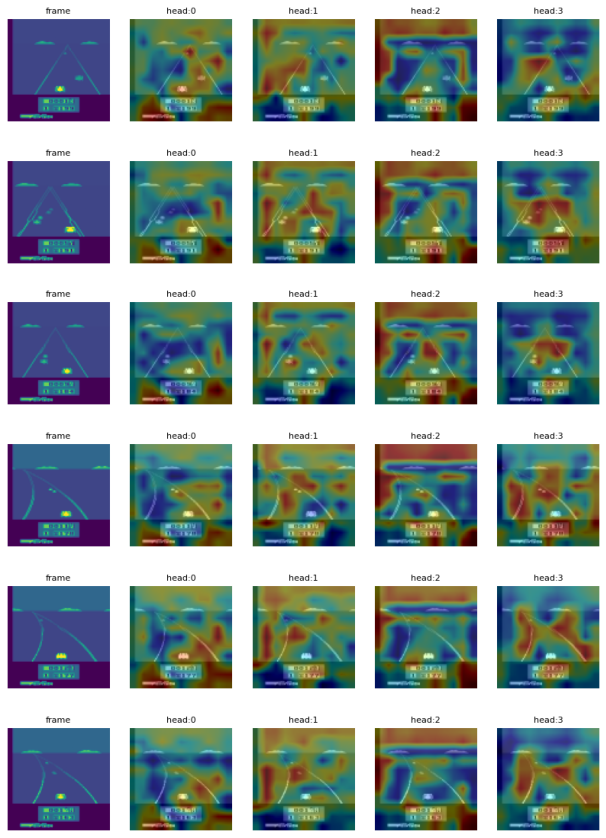}
\end{subfigure}%
\caption{Spatial attention visualization for Divided Space-Time model on Pong (Left) and Enduro (Right).}
\label{fig:div_spt_pong_enduro}
\end{figure}

\begin{figure}[hbt!]
\centering
\begin{subfigure}{0.48\textwidth}
    \centering
    \includegraphics[width=1.0\textwidth]{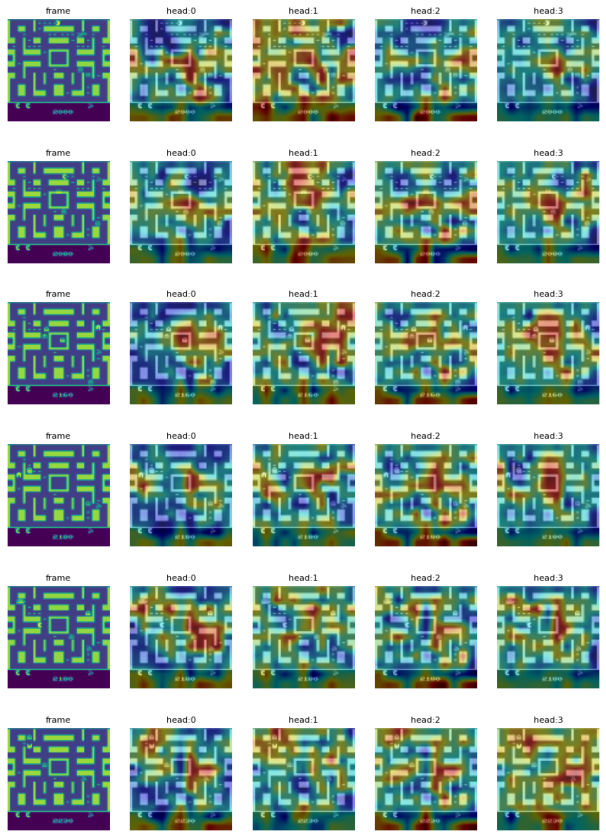}
\end{subfigure}\hspace{0.04\textwidth}%
\begin{subfigure}{0.48\textwidth}
    \centering
    \includegraphics[width=1.0\textwidth]{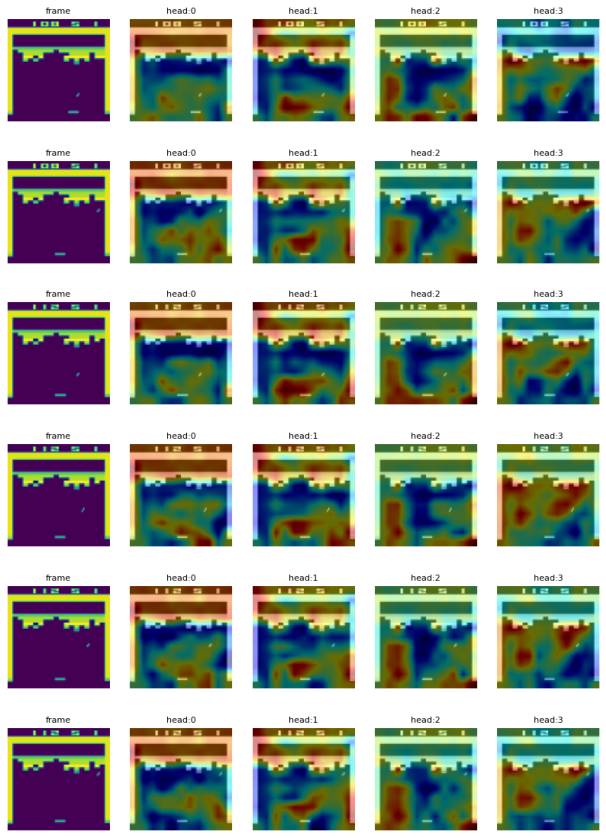}
\end{subfigure}%
\caption{Spatial attention visualization for Divided Space-Time model on Pacman (Left) and Breakout (Right).}
\label{fig:div_spt_pacman_breakout}
\end{figure}

In Figure \ref{fig:div_spt_pacman_breakout} Left, a Pacman agent trained for 60M steps in visualized. We could observe the consistent attention pattern of head-2 over score board. Also, there is an attention pattern over the agent in many instances (first, second and third rows, head-1), and (last row, head-0). \\[0.1in]
In Figure \ref{fig:div_spt_pacman_breakout} Right, a Breakout agent trained for 60M steps in visualized. For all frames, head-3 shows two `attention-clouds' on left and right top corners of the brick-group, where the agent is attempting to drill a `tunnel' to the inner blocks. In the last three rows, the crimson attention-cloud follows the ball to top right corner, where it eventually manages to tunnel inside as observed from later frames.\\[0.1in]

\begin{figure}[hbt!]
\centering
\begin{subfigure}{1.0\textwidth}
    \centering
    \includegraphics[width=1.0\textwidth]{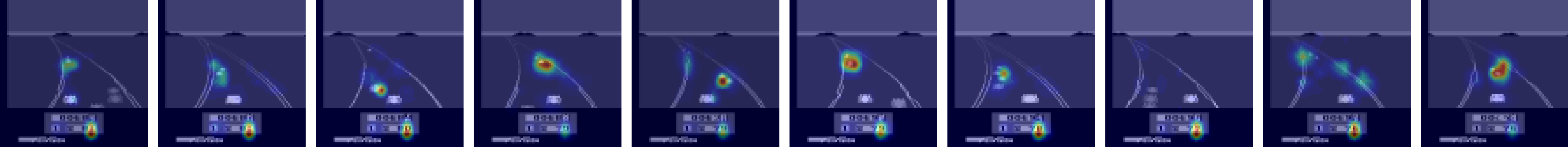}
\end{subfigure}
\par\medskip
\begin{subfigure}{1.0\textwidth}
    \centering
    \includegraphics[width=1.0\textwidth]{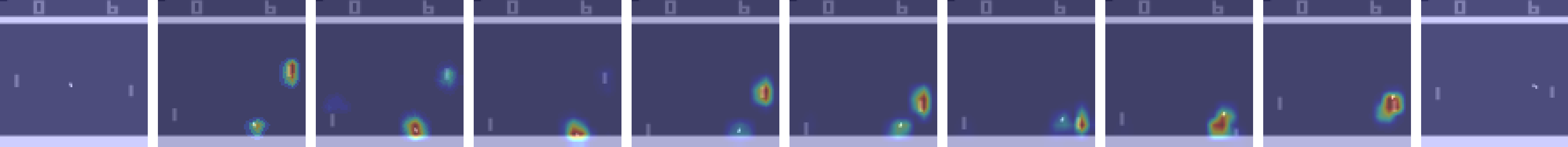}
\end{subfigure}
\par\medskip
\begin{subfigure}{1.0\textwidth}
    \centering
    \includegraphics[width=1.0\textwidth]{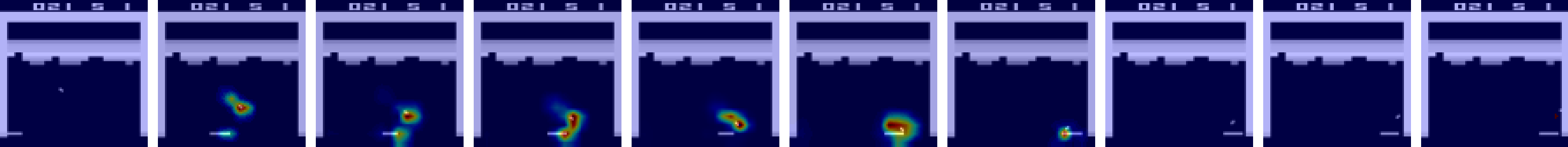}
\end{subfigure}%
\par\medskip
\begin{subfigure}{1.0\textwidth}
    \centering
    \includegraphics[width=1.0\textwidth]{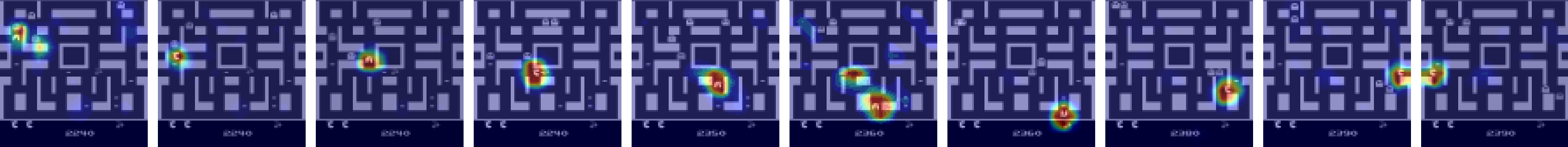}
\end{subfigure}%
\caption{Saliency maps for Divided space time model on Enduro, Pong, Breakout and Pacman(from top to bottom).}
\label{fig:div_spt_actor_salmap}
\end{figure}

Saliency map visualization(Figure \ref{fig:div_spt_actor_salmap}) of Divided Space-Time model reveals similar patterns to previous models. For instance, with both Pong and Breakout environments, the saliency map indicates more active regions when ball approaches the agent in the lower half of frame compared to the upper half. Movies of spatial attention visualization and saliency maps visualization for Divided Space-Time model trained on all four environments can be found at \url{https://imgur.com/a/MLZx3hz}.

\section{TimeSformer architecture: Visualization of Spatio-Temporal attention}
\label{sec:joint_spt_image_visualization}
In this section, we visualize Spatio-Temporal attention $\texttt{attn\_sp\_temp}$ $\in \mathbf{R}^{heads\times (h.w)\times (klen.h.w)}$ of Joint Space-Time model. Here $klen$ denotes the total context length including cached tokens in memory and the current token. We do both a graphical and image-based visualization of the attention matrix $\texttt{attn\_sp\_temp}$. \\[0.1in]

\subsection{Graphical visualization of Spatio-Temporal attention}
In this subsection, we split spatio-temporal attention matrix $\texttt{attn\_sp\_temp}$ $\in \mathbf{R}^{heads\times (h.w)\times (klen.h.w)}$ into its heads. Each head $\texttt{attn\_sp\_temp\_head\_i}$ $\in \mathbf{R}^{(h.w)\times (klen.h.w)}$ is converted to an attention heat map, indicating red regions for high attention and green regions for low attention (Section \ref{sec:cv_colormap_scale}). Attention heads $\texttt{attn\_sp\_temp\_head\_i}$ are normalized to $[0,1]$ using Equation {\ref{eq:attention_visual_normalization}}. \\[0.1in]

Attention heads $\texttt{attn\_sp\_temp\_head\_i}$ $\in \mathbf{R}^{(h.w)\times (klen.h.w)}$ are visualized (Figure \ref{fig:vit_joint_3d_attention}) for 10 consecutive time-steps. Leftmost column represents head-0 and rightmost head-3. Time increases from $t$ at top row to $t + 10$ at last row. Each tile in the figure represents one time instant for one of the four heads. In each tile, right-most color bands represent the latest token and left-most the oldest tokens in memory. There are multiple high attention regions represented by red bands and lower attention regions by scattered yellowish and green segments. Head-3 represented by right-most column has red regions at right edges indicating the agent's attention to the most recent frames. One interesting observation is that the red colored high-attention bands look stationary with a constant offset to the current token, across the time axis. This conflicts with the pattern observed in temporal attention visualization (Figure \ref{fig:adaptive_attention_heads}) of Adaptive architecture (\ref{sec:adaptive_transformer_xl}) where the key-events holding high attention moved back in time. The observation also hints the possibility that the agent is merely attending to random frames that are behind the current frame by a constant time offset, without dynamically responding to changes in environment. In other words, the agent could be possibly `overfitting' to game dynamics by adapting this offset value to different game situations. \\[0.1in]

\begin{figure}[hbt!]
    \centering
    \includegraphics[width=1.0\textwidth]{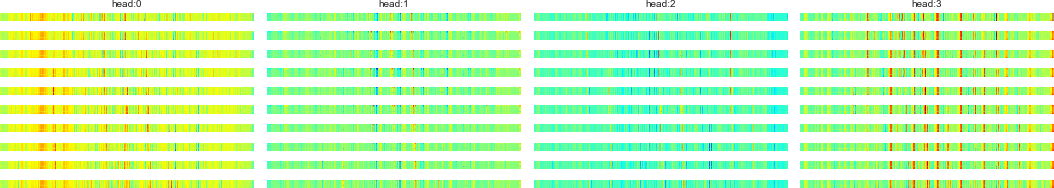}
    \caption{Heat map of attention heads 1, 2, 3 and 4 for 10 consecutive time-steps for Atari Pong environment with Joint Space-Time model (Figure \ref{fig:vit_attention_scheme} Left).}
    \label{fig:vit_joint_3d_attention}
\end{figure}

\subsection{Visualization of Spatio-Temporal attention on images}
In order to visualize Spatio-Temporal attention $\texttt{attn\_sp\_temp}$ $\in \mathbf{R}^{heads\times (h.w)\times (klen.h.w)}$ on corresponding images, we average \texttt{attn\_sp\_temp} across axis=1 to generate \texttt{attn\_sp\_temp\_mean} $\in \mathbf{R}^{heads\times 1\times (klen.h.w)}$. We project back all heads of \texttt{attn\_sp\_temp\_mean} of shape $1\times (klen.h.w)$, on corresponding images by extracting time and spatial coordinates from attention data \texttt{attn\_sp\_temp\_mean}. For ease, we have only taken a fraction of top attention values for visualization. Images are numbered in Figure \ref{fig:joint_sp_temp_pong_head01} - \ref{fig:joint_sp_temp_breakout_head23} based on their corresponding time-step index. For example, in Figure \ref{fig:joint_sp_temp_pong_head23}, Right (attention head-4), last row, for current image at time-step:373, spatio-temporal attention is spread mainly over frames at time-steps: 322, 331, 341, 351 and 363. We normalized the attention matrix in the range $[0,1]$ w.r.t to maximum attention value per head, using Equation \ref{eq:attention_visual_normalization}. \\[0.1in]

\begin{figure}[hbt!]
\centering
\begin{subfigure}{0.48\textwidth}
    \centering
    \includegraphics[width=1.0\textwidth]{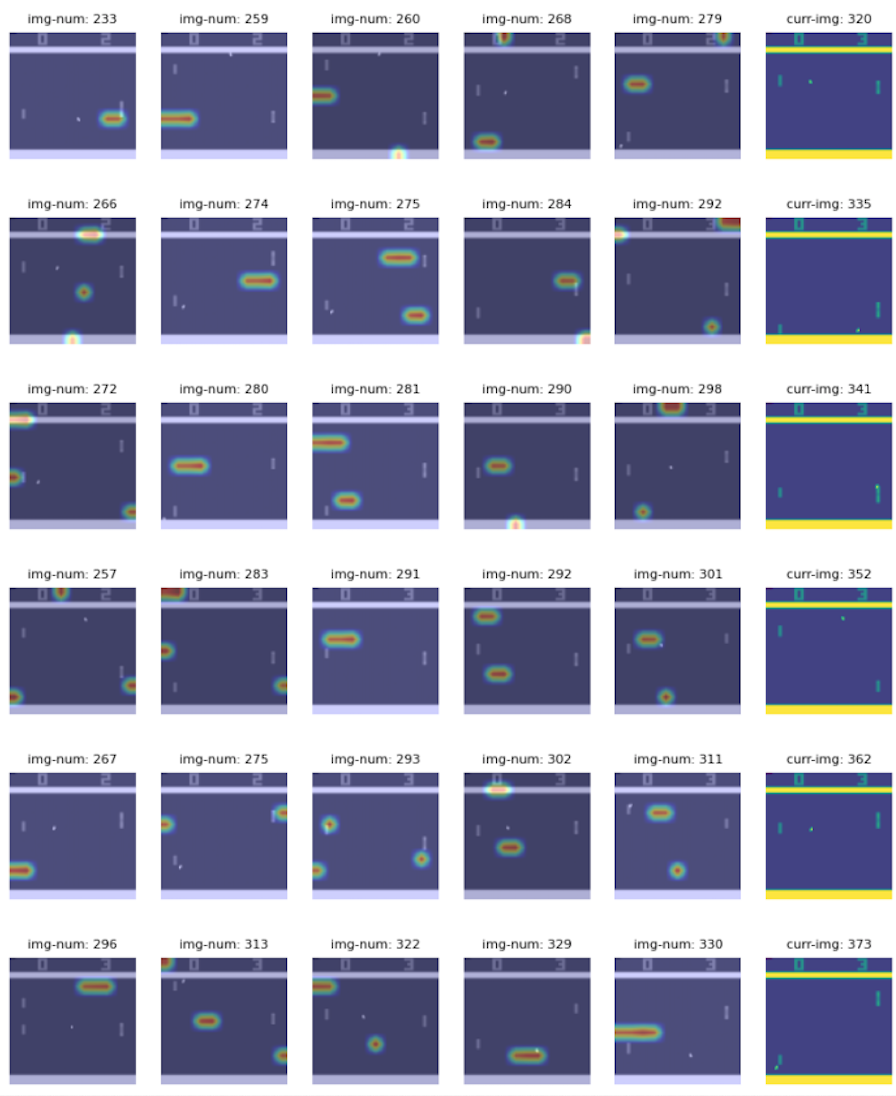}
\end{subfigure}\hspace{0.04\textwidth}%
\begin{subfigure}{0.48\textwidth}
    \centering
    \includegraphics[width=1.0\textwidth]{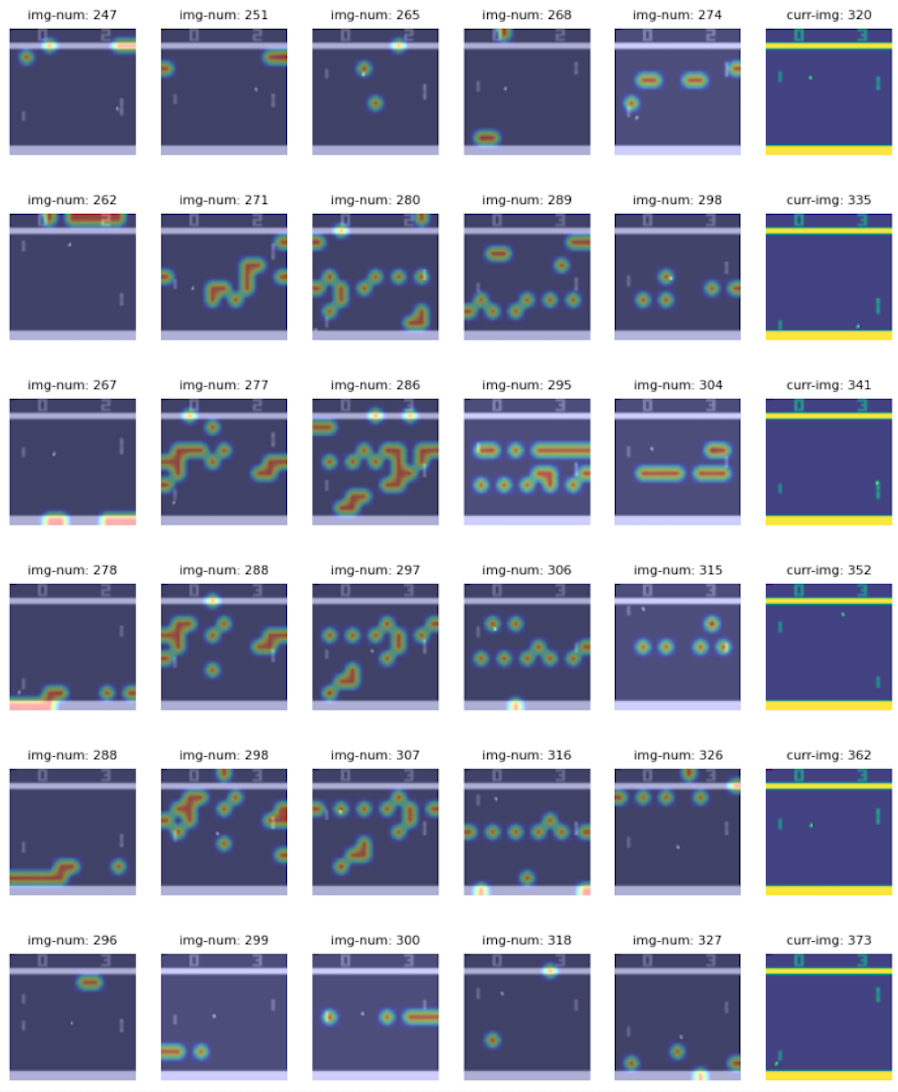}
\end{subfigure}%
\caption{Spatial-temporal attention visualization of head-1(Left) and head-2(Right) for Joint Space-Time model on Pong.}
\label{fig:joint_sp_temp_pong_head01}
\end{figure}

\begin{figure}[hbt!]
\centering
\begin{subfigure}{0.48\textwidth}
    \centering
    \includegraphics[width=1.0\textwidth]{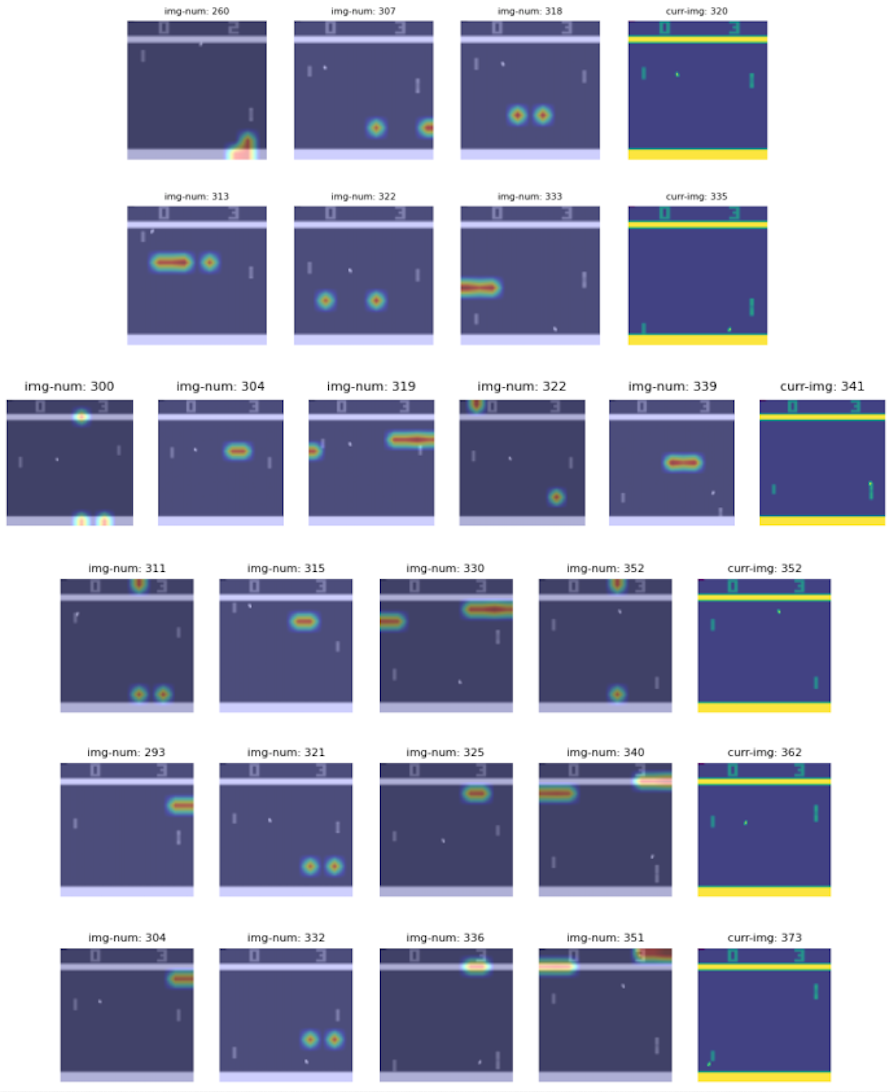}
\end{subfigure}\hspace{0.04\textwidth}%
\begin{subfigure}{0.48\textwidth}
    \centering
    \includegraphics[width=1.0\textwidth]{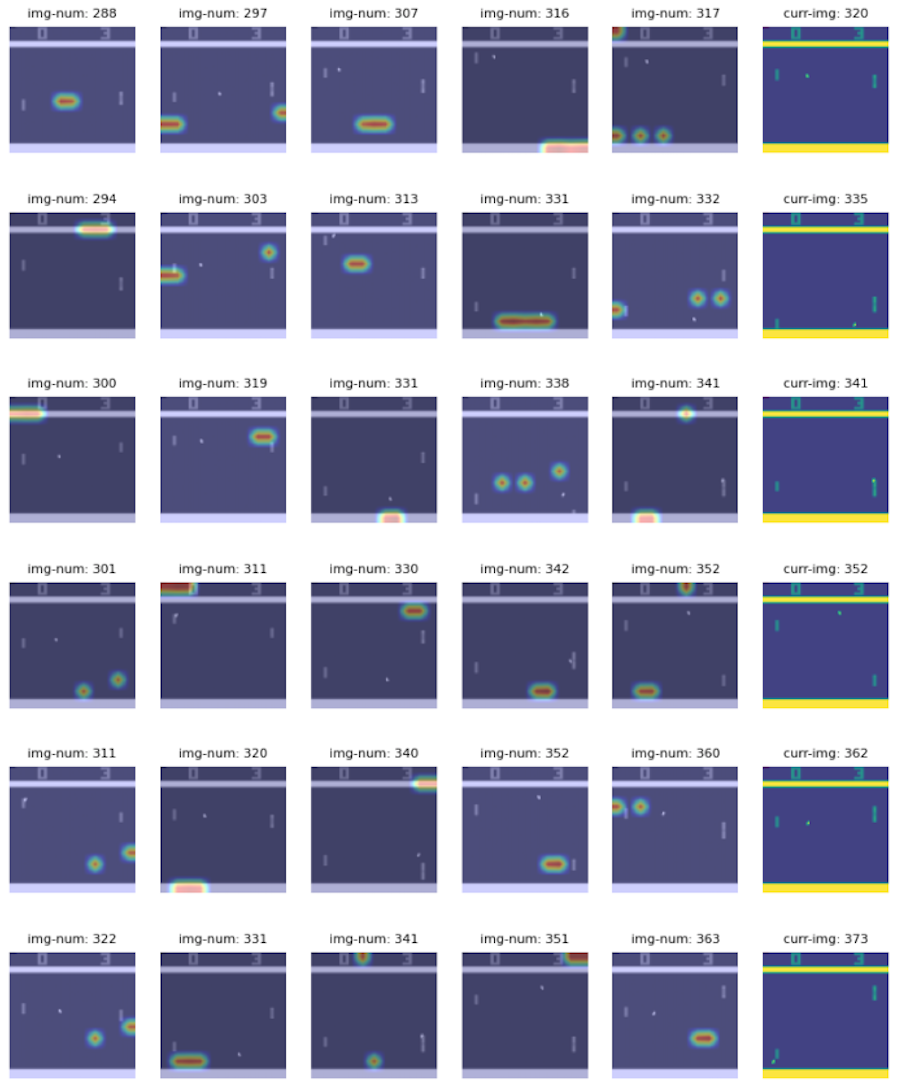}
\end{subfigure}%
\caption{Spatial-temporal attention visualization of head-3(Left) and head-4(Right) for Joint Space-Time model on Pong.}
\label{fig:joint_sp_temp_pong_head23}
\end{figure}

We notice that with Pong environment, for attention heads: 3 and 4(Figure \ref{fig:joint_sp_temp_pong_head23}), immediate past frames are attended more compared to older frames, whereas heads: 1 and 2(Figure \ref{fig:joint_sp_temp_pong_head01}) exhibits a longer span of attention over previous frames. Other than this, we were not able to fully interpret and draw insights from the attention patterns generated. For instance, consider a Pong agent which is about to return a shot. Intuitively, the key moments where the agent could have the majority of its attention are time-space moments when the opponent struck the ball, or when the ball changed direction after hitting the walls and so on. Unfortunately, we could not find clear trends in 3D-attention visualization (Fig \ref{fig:joint_sp_temp_pong_head01} - Fig \ref{fig:joint_sp_temp_pong_head23}) to establish the same. \\[0.1in]

\begin{figure}[hbt!]
\centering
\begin{subfigure}{0.48\textwidth}
    \centering
    \includegraphics[width=1.0\textwidth]{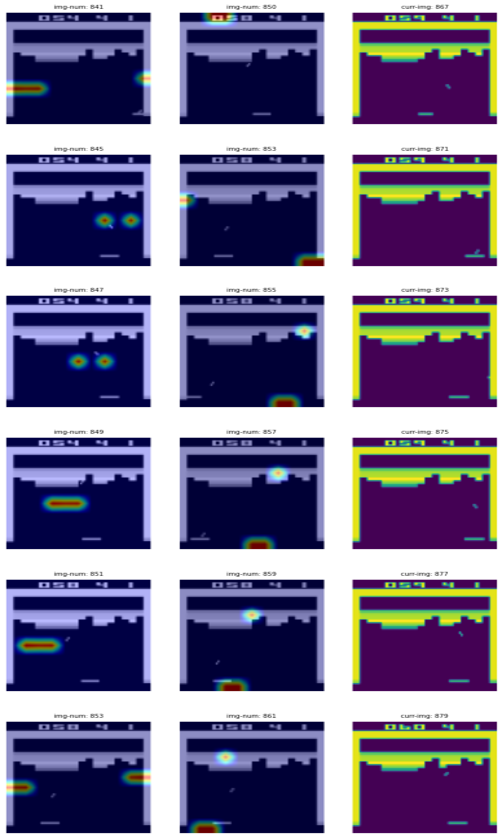}
\end{subfigure}\hspace{0.04\textwidth}%
\begin{subfigure}{0.48\textwidth}
    \centering
    \includegraphics[width=1.0\textwidth]{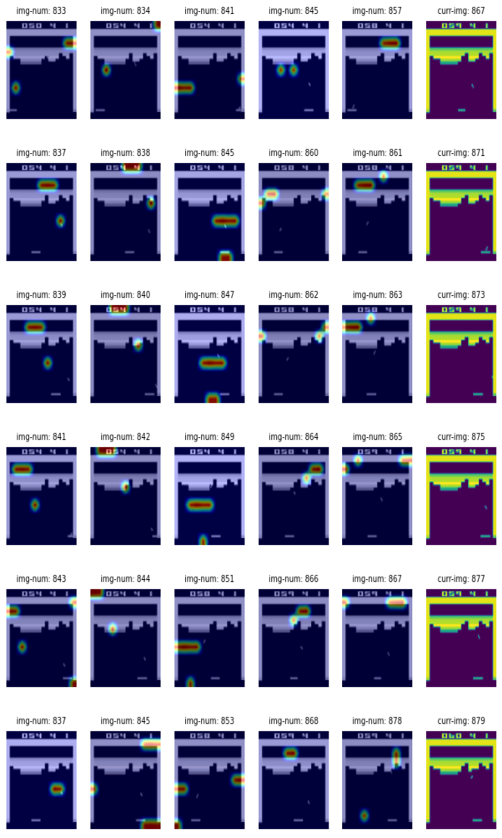}
\end{subfigure}%
\caption{Spatial-temporal attention visualization of head-1(Left) and head-2(Right) for Joint Space-Time model on Breakout.}
\label{fig:joint_sp_temp_breakout_head01}
\end{figure}

\begin{figure}[hbt!]
\centering
\begin{subfigure}{0.48\textwidth}
    \centering
    \includegraphics[width=1.0\textwidth]{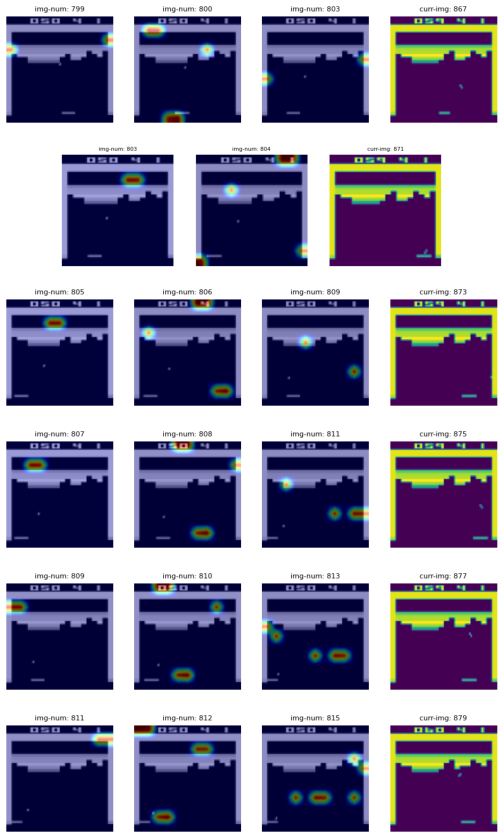}
\end{subfigure}\hspace{0.04\textwidth}%
\begin{subfigure}{0.48\textwidth}
    \centering
    \includegraphics[width=1.0\textwidth]{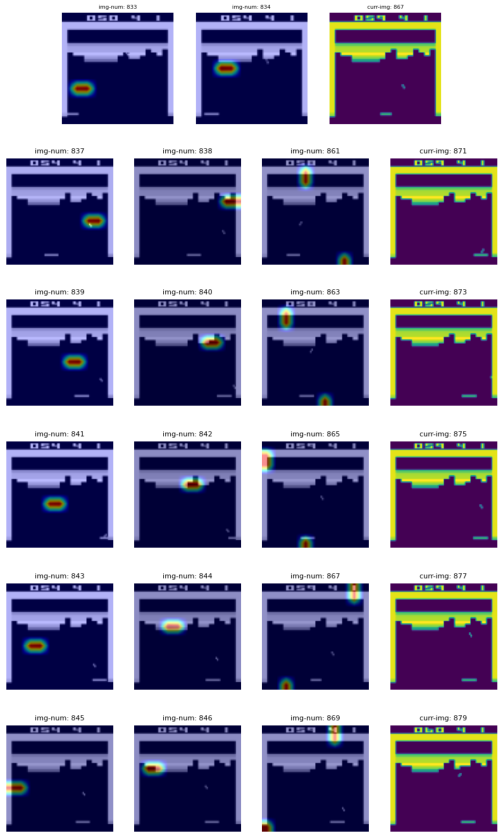}
\end{subfigure}%
\caption{Spatial-temporal attention visualization of head-3(Left) and head-4(Right) for Joint Space-Time model on Breakout.}
\label{fig:joint_sp_temp_breakout_head23}
\end{figure}

With attention visualization in Breakout environment,  (Figure \ref{fig:joint_sp_temp_breakout_head01} - \ref{fig:joint_sp_temp_breakout_head23}), we notice constant offsets between the current frame and the frames that are attended. For example, attention head-2 (Figure \ref{fig:joint_sp_temp_breakout_head01} Right) shows current frames attending to past frames consistently off by a constant offset. This observation also aligns with heat map visualization (Figure \ref{fig:vit_joint_3d_attention}) where the red high-attention regions appeared at constant offsets with respect to the current token. It also strengthens our suspicion that, the agent is just looking out for frames that are lagging behind the current frame by a constant offset to take the right action, instead of actively looking out for changes in environment.\\[0.1in]

\begin{figure}[hbt!]
\centering
\begin{subfigure}{1.0\textwidth}
    \centering
    \includegraphics[width=1.0\textwidth]{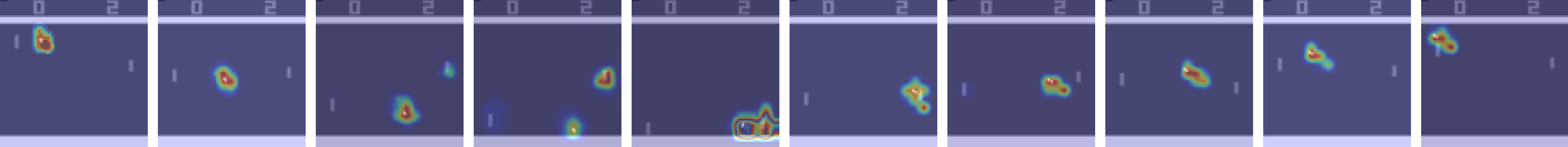}
\end{subfigure}
\par\medskip
\begin{subfigure}{1.0\textwidth}
    \centering
    \includegraphics[width=1.0\textwidth]{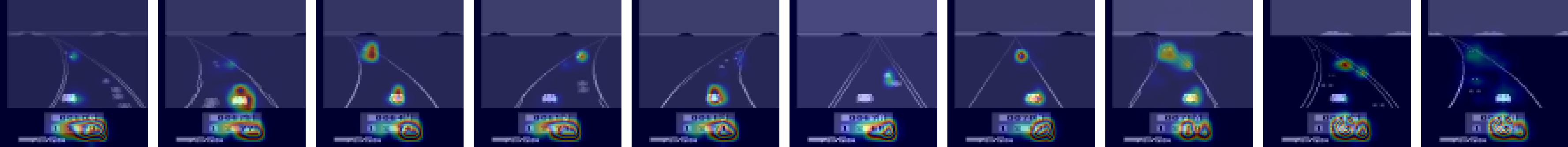}
\end{subfigure}
\par\medskip
\begin{subfigure}{1.0\textwidth}
    \centering
    \includegraphics[width=1.0\textwidth]{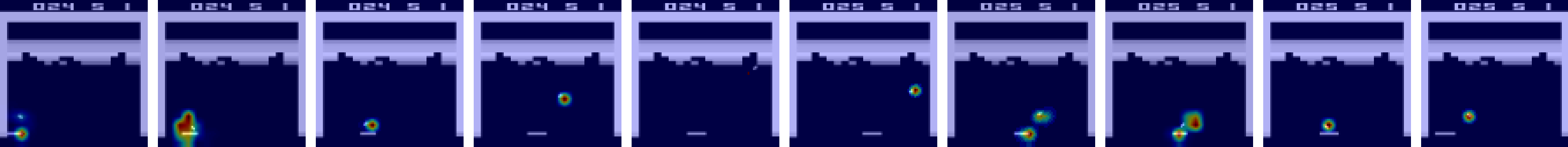}
\end{subfigure}
\caption{Saliency maps for Joint space time model on Pong, Enduro and Breakout(from top to bottom).}
\label{fig:joint_spt_actor_salmap}
\end{figure}

Unlike saliency map of the previous models, joint Space-Time model's saliency map(Figure \ref{fig:joint_spt_actor_salmap}) for Pong and Breakout, shows agent paying attention to upper half of frame too. Also, last five frames of Pong environment(Figure \ref{fig:joint_spt_actor_salmap} Top row) shows agent making a `kill shot'. Kill shot usually refers to a return shot from agent, in which it hits the ball using its corners and opponent is unable to return the fast advancing ball. Joint Space-Time model did not learn well on Pacman environment (Figure \ref{fig:pacman_result}) and hence saliency map analysis was not done for Pacman environment. Movie of saliency maps visualization for Joint Space-Time model trained on Pong, Enduro and Breakout can be found at \url{https://imgur.com/a/EovpEWE}.
    
    \chapter{Conclusion and Future work}
    To use Neural networks in safety critical real-world use cases, it is beneficial to have models which have good accuracy along with improved interpretability. We believe that a model being interpretable is a big value addition and enjoys the trust of the user. In our work, we presented various temporal architectures in RL domain, based on Attention mechanism, which performed well on Atari-2600 game suite. We were able to generate satisfactory spatial and spatio-temporal segmentations of an agent's environment for some architectures, to better interpret the agent's actions. For most of the models presented, we visualized the agent's attention matrix using alpha-blending techniques and saliency maps \cite{DBLP:journals/corr/abs-1711-00138}. To an extent, we were able to establish spatio-temporal similarity between video classification tasks and temporally extended RL tasks. This could open the possibility of trying similar attention-related video-classification architectures in RL as well. Also, ours was one of the first works using Vision Transformers \cite{dosovitskiy2021an} in RL domain, to the extent of our knowledge.  \\[0.1in]
In order to improve the quality of visualization, one potential option is to use more computational power to train models for longer duration. Similar to multilayer models \cite{dosovitskiy2021an}, \cite{DBLP:journals/corr/abs-2102-05095}, we plan to replace our current single layer config with a multilayer Transformer encoder, for improving visualization quality. As explained in Section \ref{sec:joint_spt_image_visualization}, the current spatio-temporal attention heads(Fig \ref{fig:joint_sp_temp_pong_head01} - Fig \ref{fig:joint_sp_temp_pong_head23}) are not probably attending to key-events in past. One possible option to improve spatio-temporal attention visualization is to debug the model using a toy environment like MDP Playground (MDPP)\cite{rajan2021mdp} which offers simple game dynamics compared to Atari. MDP Playground provides toy environments like polygon-based discrete image environment and grid world. For instance, in polygon-based discrete environment, an agent has to figure out the optimal state(an image polygon) which could be a triangle, square, pentagon, hexagon, etc and perform the transition to the optimal state in minimum number of steps. We expect our attention architectures to generate better interpretable visuals like more attention along polygon edges, or polygon difference area between consecutive states, with MDPP discrete image environments.

\begin{figure}[hbt!]
\centering
\begin{subfigure}{0.20\textwidth}
    \centering
    \includegraphics[width=1.0\linewidth]{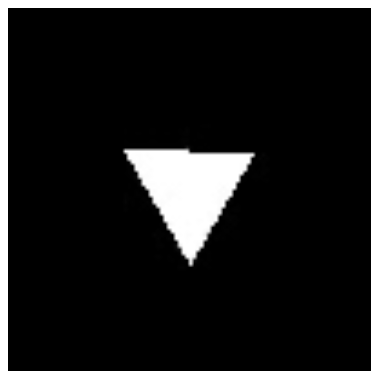}
\end{subfigure}
\hspace{0.1\textwidth}%
\begin{subfigure}{0.20\textwidth}
    \centering
    \includegraphics[width=1.0\linewidth]{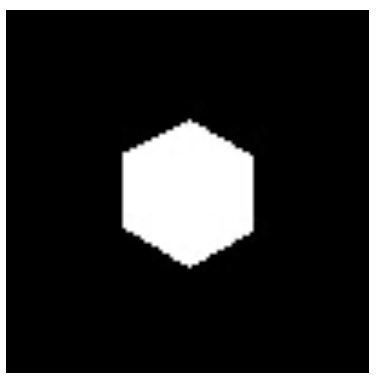}
\end{subfigure}%
\caption{Two sample states in MDPP discrete image environment \cite{rajan2021mdp}. States are $n$-sided polygons with $n \geq 3$. (Image source: \cite{rajan2021mdp})}
\label{fig: breakout_result}
\end{figure}
    
    \clearpage
    \printbibliography
    
    \appendix
    \titleformat{\chapter}[display]
    	{\normalfont\bfseries}{\chaptertitlename\ \thechapter}{20pt}{\huge}
    	
    
    \chapter{Hyperparameters used}
    In this chapter, we consolidate the optimal hyperparameter settings used in all our architectures.

\begin{table}[hbt!]
    \caption{Hyperparameter setting for arch-i:Mott (Section \ref{sec:motts_architecture_section}), arch-ii:Adaptive architecture (Section \ref{sec:adaptive_transformer_xl}), arch-iii:Spatio-Temporal sequential (Figure \ref{fig:spatio_temporal_architectures} Left) and arch-iv:Spatio-Temporal one-shot architecture using actor query(Figure \ref{fig:spatio_temporal_architectures} Right)}
    \centering
        \begin{tabular}{p{3cm}|p{1.5cm}|p{1.5cm}|p{1.5cm}|p{1.5cm}}
         parameter & arch-i & arch-ii & arch-iii & arch-iv\\
         \hline \hline
         unroll\_length & 160 & 239 & 239 & 239\\
         chunk\_size & NA & 80 & 80 & 80\\
         num\_buffers & 60 & 60 & 60 & 60\\
         num\_actors & 32 & 32 & 50 & 50\\
         batch\_size & 12 & 16 & 12 & 12\\
        \end{tabular}
    \label{tab:mott_sp_temp_hyperparams}
\end{table}

\begin{table}[hbt!]
    \caption{Hyperparameter setting for Divided Space-Time (Figure \ref{fig:vit_attention_scheme} Right) and Joint Space-Time (Figure \ref{fig:vit_attention_scheme} Left)}
    \centering
        \begin{tabular}{p{4cm}|p{2.5cm}|p{2.5cm}}
         parameter & Divided & Joint\\
         \hline \hline
         unroll\_length & 239 & 239\\
         chunk\_size & 80 & 10\\
         num\_buffers & 40 & 40\\
         num\_actors & 32 & 32\\
         batch\_size & 24 & 4\\
         embed\_dim & 16 & 16\\
         patch\_size & 7 & 7\\
         height & 84 & 42\\
         width & 84 & 42\\
         n\_layer & 1 & 1\\
        \end{tabular}
    \label{tab:vit_timesformer_hyperparams}
\end{table}

    \end{flushleft}
\end{document}